\documentclass[sigconf,nonacm]{aamas}

\usepackage{balance} 

\title[Efficient Planning in Reinforcement Learning via Model Introspection]{Efficient Planning in Reinforcement Learning \\ via Model Introspection}

\author{Gabriel Stella}
\affiliation{
  \institution{Texasm A\&M University}
  \city{College Station, TX}
  \country{USA}}
\email{gabrielrstella@tamu.edu}

\begin{abstract}
Reinforcement learning and classical planning are typically seen as two distinct problems, with differing formulations necessitating different solutions.
Yet, when humans are given a task, regardless of the way it is specified, they can often derive the additional information needed to solve the problem efficiently.
The key to this ability is \emph{introspection}: by reasoning about their internal models of the problem, humans directly synthesize additional task-relevant information.
In this paper, we propose that this introspection can be thought of as \emph{program analysis}.
We discuss examples of how this approach can be applied to various kinds of models used in reinforcement learning. We then describe an algorithm that enables efficient goal-oriented planning over the class of models used in relational reinforcement learning, demonstrating a novel link between reinforcement learning and classical planning.

\end{abstract}


\usepackage{amsmath, mathtools, amsthm}

\usepackage{caption}
\usepackage{subcaption}

\usepackage[table]{xcolor} 
\definecolor{verylightgray}{rgb}{0.9, 0.9, 0.9}

\usepackage[ruled,noend,linesnumbered,algo2e]{algorithm2e}     
\SetAlCapSkip{1em}      
\SetAlFnt{\sf\footnotesize}    
\SetAlCapNameFnt{\footnotesize}   
\SetAlCapFnt{\footnotesize}       
\SetArgSty{textsf}      
\SetCommentSty{textsf}  
\SetKwComment{myc}{\quad// }{} 

\SetNlSty{footnotesize}{}{}        
\DontPrintSemicolon

\usepackage{xcolor}

\usepackage{courier}
\newcommand{\code}[1]{\texttt{#1}}

\usepackage[normalem]{ulem}

\DeclareMathOperator*{\argmax}{argmax}

\begin{document}


\pagestyle{fancy}
\fancyhead{}


\maketitle

\section{Introduction}

One of the most amazing facets of human intelligence is the ability to generate new knowledge via reasoning.
A particularly important instantiation of this is that when given a description of a task, humans can devise efficient methods to solve that task, even making use of information that is not explicitly stated.
For example, consider a situation in which a person plays a maze-navigation video game. They learn the rules, including how to gain points (and what actions lead to penalties), and are instructed to play the game in an attempt to achieve maximal reward.

Unlike existing methods in artificial intelligence, most people -- even children -- could then apply goal-directed, spatially-guided search methods to efficiently solve any new maze, even though no goal or heuristic information was included in the problem specification.

Thus, even when a task is described using information typical of reinforcement learning, i.e., a transition and reward model \citep{sutton18}, humans naturally apply techniques reminiscent of classical task planning \citep{garrett20}. How can this be the case?

Although one may think that this human capability arises solely from the transfer of outside knowledge, we propose that all of the necessary information is actually contained within the person's learned models of the environment.
Thus, after learning from experience, humans can gain additional information by reasoning over these models, i.e., \emph{introspecting}.
Furthermore, we propose that this \emph{model introspection} can be viewed as \emph{program analysis}, allowing us to apply this concept to artificial agents.

\begin{figure}[b!] 
\begin{center}

     \begin{subfigure}[t]{0.49\columnwidth}
     \centering
         \includegraphics[width=\textwidth]{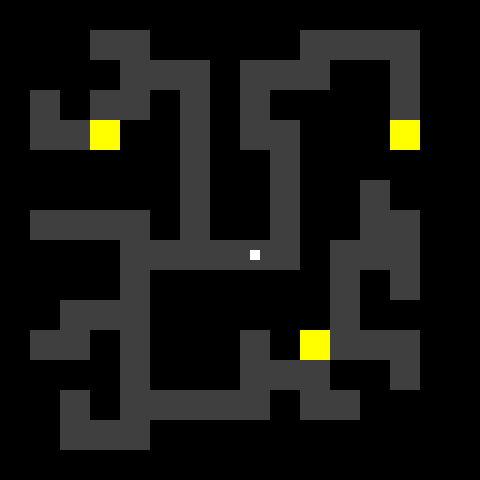}
         \caption{example maze state, with the player represented as a white square and three possible destinations marked as yellow squares}
         \label{subfig:goal-img}
     \end{subfigure}
     \hfill
     \begin{subfigure}[t]{0.49\columnwidth}
     \centering
         \includegraphics[width=\textwidth]{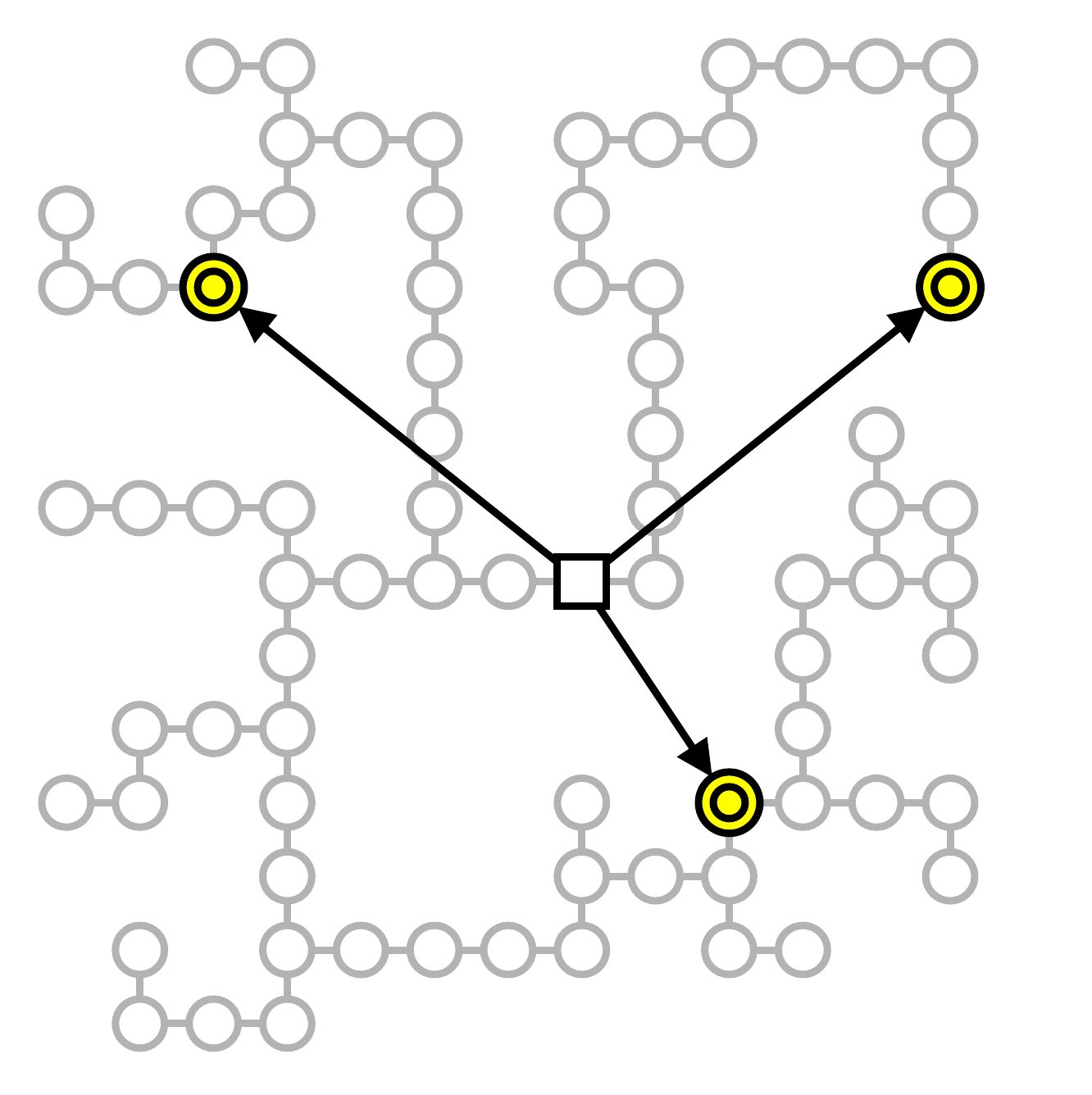}
         \caption{corresponding state space, with the current state denoted by a square and milestones marked using double-outlined circles}
         \label{subfig:goal-space}
     \end{subfigure}
     
\end{center}
\caption{Example maze with milestone states marked in the corresponding state space graph.
While existing approaches to model-based reinforcement learning search the state space without long-term guidance,
our reward-based milestones allow the agent to plan more efficiently using informed search algorithms.
Note that these milestones are not given as domain knowledge; instead, they are computed by the agent based on information contained within its learned models.
}
\Description{An image of a maze-like environment accompanied by a visualization of the corresponding state space, highlighting the utility of planning to reach specific states called milestones.}
\label{fig:goal-milestones}
\end{figure}

In the current paper, we explore one particularly promising application of this concept: the development of highly efficient planning methods for reinforcement learning.
To do so, we introduce the concept of \emph{milestones}: hypothetical future states which, if reached, will grant the agent a reward. An example is shown in Figure~\ref{fig:goal-milestones}.
These milestones can be constructed by analyzing the agent's reward model, then treated as goals for input into an efficient classical planning algorithm.
This represents an important new bridge between the fields of reinforcement learning and classical planning.

To demonstrate the significance of this connection, we provide an algorithm that generates milestones for models in the setting of relational reinforcement learning \citep{dzeroski01}. We then describe a bilevel planning algorithm that uses milestones, in addition to a domain-agnostic heuristic, to solve several challenging tasks. Our empirical evaluation reveals a substantial runtime improvement: while prior methods take exponentially longer
as the task complexity increases, our planning algorithm, which we call \emph{introspector}, solves these reinforcement learning problems in sub-exponential time.

Although we focus much of our discussion on this application to reward-maximization planning, this is just one example of the utility of model introspection.
The broader purpose of this paper is to introduce and motivate the use of program analysis over interpretable models \emph{in general}.
There are many exciting avenues for future research; our discussion at the end of this paper outlines several.
We also touch on the relationship between our work and the important problem of representation learning.
We hope that the promising results from our investigation into planning will spur interest in researching further aspects and applications of model introspection.

\section{Background}

We begin by covering relevant background in reinforcement learning and planning, including a discussion contrasting the existing methods in both areas. We then motivate the use of programs to implement transition and reward models.

\subsection{Reinforcement Learning}

Reinforcement learning is a subfield of machine learning that studies the interactions between an agent and its environment as the agent learns to obtain rewards.
We use the discrete, deterministic Markov Decision Process (MDP) formulation \cite{kaelbling96, sutton18}, where an environment consists of a discrete set of states $S$, a discrete set of actions $A$, a transition function $T(s, a)$, and a reward function $R(s, a, s')$.
Although MDPs in general may have stochastic transitions, we focus on deterministic environments, noting that our ideas can be extended to the general case.
The reward encodes some task objective, such as moving items into particular positions.

Problem-solving occurs in \emph{episodes}, each beginning in a new initial state $s_0$.
This state is sampled from the environment's \emph{initial state distribution}, $s_0 \sim B_\theta$, which may be parameterized by some $\theta$ that controls factors such as the number of each type of object.
Then, the agent and environment interact in a loop for some number of steps.
First, the agent receives an observation of the current state $s_i$ and chooses an action $a_i$.
The environment then transitions to a new state $s_{i+1} = T(s_i, a_i)$ and generates a reward $r_{i+1} = R(s_{i}, a_{i}, s_{i+1})$.
The agent learns by observing the \emph{transition} $(s_{i}, a_{i}, s_{i+1}, r_{i+1})$.
This process repeats until the episode ends after some number of steps (e.g., after reaching a particular state or after a specific number of actions), which we call the \emph{horizon} $H$.
The agent's \emph{score} for the episode, also called its \emph{return}, is $\sum_{i=1}^{H} r_i$.
The agent's objective is to collect information and devise a policy function $a = \pi(s)$ that will maximize its score.

Reinforcement learning algorithms can be broadly divided into two categories: \emph{model-free} and \emph{model-based}.
Methods in the former category do not attempt to make (or use) a model of the environment's dynamics functions, $T$ and $R$.
Two popular model-free approaches are \emph{policy learning} and \emph{value estimation}.
Policy learning methods aim to directly construct an effective policy, e.g., via the policy gradient method \citep{schulman17}.
Value estimation algorithms, such as $Q$-learning \citep{watkins92, mnih15}, use a learned approximation of the \emph{action-value function} $Q(s, a)$ to make decisions.
Although model-free techniques can solve challenging control tasks, they often perform poorly in complex domains that require planning far ahead, such as mazes and other puzzles. 

To tackle such tasks, model-based methods use their own models of the environment, which we denote by $\hat{T}$ and $\hat{R}$ for the transition model and reward model, respectively.
Once an agent learns accurate models of the environment's transition and reward dynamics, it 
can conduct simulated state-space search to find sequences of actions that obtain high scores.
We refer to this general process as \emph{planning}.
Unfortunately, planning in reinforcement learning is computationally expensive.
\emph{Value iteration} finds optimal plans, but it requires expanding all states reachable from $s_0$ and iterating a large number of times to converge \citep{sutton18}.
In order to reduce memory and runtime cost, methods such as \emph{beam search} and \emph{Monte Carlo Tree Search} (MCTS) explore only part of the reachable state space, weighted towards trajectories that give the agent some reward \citep{coulom07, rosin10}.
To depend less on intermediate rewards, recent work has introduced a version of MCTS that is guided by learned value and policy functions \citep{schrittwieser20}.
However, these algorithms are still drastically outperformed by methods from the field of \emph{classical planning}.

\subsection{Classical Planning}

The field of \emph{classical planning} takes a different approach to decision-making. While reinforcement learning studies agents that learn to obtain reward, planning focuses on reaching specified goal states, which are explicitly given to the agent along with a white-box model of the environment's dynamics \cite{garrett20}.
Environments are typically described using high-level, domain-specific information about objects, with states comprising sets of predicates (e.g., ``is object A on the floor?'') and actions represented using parameterized templates called \emph{action schemas} (e.g., ``move object A onto object C'') \citep{mcdermott98}.
Classical planning algorithms are responsible for finding sequences of actions that will take the environment from its initial state to a state that satisfies some goal conditions (e.g., ``object A is on object C and object B is on the floor''). While there are many approaches \citep{correa24},
one of the most popular is \emph{heuristic state-space search} \cite{bonet01},
which uses informed search algorithms such as A* to compute plans \cite{haslum00, nguyen00}.

These methods are highly efficient, able to find successful -- or even optimal -- sequences of actions in long-horizon tasks.
This stems from two key factors, which together enable the use of search methods that intelligently guide their exploration of the space towards promising trajectories \emph{without relying on intermediate feedback such as rewards}.
First, the presence of explicit goals, i.e., states (or conditions) for the planner to try to reach.
Second, the use of heuristics, which estimate the distance between two states (or from a state to the goal conditions).
While the goals for a classical planning task are given in the problem statement, heuristics are computed \emph{by the planner} using domain-agnostic methods that process the environment's transition model \citep{haslum00}.
Although they are not the focus of the current paper,
the concepts behind heuristic generation give an interesting hint as to how we can improve the efficiency of planning in reinforcement learning as well.

\subsection{Planning in Reinforcement Learning}


Learning accurate models presents a serious challenge, 
but even with a perfect model, planning algorithms for reinforcement learning are still inefficient.
The models are typically treated as \emph{black boxes}, which cannot be analyzed to get heuristics or other information.
Even if heuristics could be derived from the transition model, goals are not given to the agent. The task is instead framed as reward maximization. To solve this problem, the planner's search algorithm must seek high-scoring trajectories without any long-term direction or guidance. In other words, planners in reinforcement learning \emph{do not know where they are trying to go}.

One of the biggest impediments to this approach is the \emph{sparse reward problem}.
Methods like MCTS rely on rewards being common enough to stumble upon them by chance; the planner can then guide the search further along this same trajectory, hoping to find more. The intermediate rewards are like breadcrumbs for the search algorithm.
However, if very few transitions yield a reward, and especially if those transitions are far from the agent's current state, the planner may get ``unlucky'' and never encounter them at all.
Although this situation can potentially be improved by using learned value estimates and a prior policy, 
the current paper shows that this is not necessary: the agent's models $\hat{T}$ and $\hat{R}$ contain enough information by themselves to do efficient planning.

\paragraph{Goal-Conditioned Reinforcement Learning}

The great interest in planning for reinforcement learning has led to the development of new sub-fields, such as \emph{goal-conditioned reinforcement learning}.
This framework adds explicit goal states to MDPs.
The agent's objective is to take actions that achieve maximal reward \emph{and} lead to the specified goal.
Like in typical reinforcement learning, the environment has to be learned. Most methods struggle to construct models that maintain accuracy far into the future, so goal-conditioned reinforcement learning approaches often generate \emph{subgoals} to guide search iteratively through a reduced state space \citep{nasiriany19}.
Unlike the milestones we describe, subgoals are typically not related to the environment's reward structure \citep{zhao24};
they serve only to help the agent reach its goal, i.e., improve task completion.
In contrast, our discussion in the current paper focuses on increasing the \emph{computational efficiency} of planning. We assume that the agent is capable of constructing a model that is accurate enough to facilitate long-horizon planning. Thus, if efficiency were not a concern, then correctness (and optimality) could be guaranteed trivially.

\subsection{Reward Machines}


The idea of using the structure of the reward signal has been investigated in the form of \emph{reward machines} \cite{toroicarte22}.
A reward machine is a finite state automaton whose state determines the environment's reward function. Transitions of the automaton are based on propositional formulas evaluated over facts about the agent's environment state (e.g., ``the agent is holding coffee and is not at location A'').
This process requires an additional \emph{labeling function} $L(s, a, s') \to \{0, 1\}^n$ where $n$ is the number of distinct facts.
Algorithms can make use of their knowledge of the reward machine structure in order to learn effective policies more quickly.

The premise of reward machines is that regardless of the environment, an agent's reward signal has to be programmed by a human (even if the transition function is not). Thus, it is reasonable to give the agent white-box access to this information.
However, although reward machines expose a finite state automaton to the agent for analysis, existing methods in this line of work typically exploit only the information contained in the separation between environment state $s$ and reward machine state $u$, e.g., by learning distinct policies for each $u$, rather than inspecting the automaton directly. In addition, the reward machine is not fully white-box; details of the reward function for each $u$ are still hidden.
Thus, the agent's ability to extract useful information from the reward machine is dependent on the details of its construction.
This limitation arises due to the assumption that the reward machine is hand-crafted by a human; by instead constructing (i.e., learning) a model of its own, the agent could transform any hand-crafted reward function into a representation that is more conducive to analysis.

Another noteworthy aspect of this line of work is its application: reward machines are not used for model-based planning. Instead, prior work studied the development of model-free techniqus (based on, e.g., $Q$-learning) that exploit the reward machine to improve the agent's policy learning rate. These algorithms operate over traditional MDPs with asemantic representations, i.e., where $S$ and $A$ are arbitrary sets of unstructured elements; thus, generalization and transfer in the sense we discuss below are not possible.

\subsection{Generalization}

When humans learn a task, we are able to apply our knowledge and skills to solve novel instances of that problem. For example, after learning to solve mazes, a human can solve \emph{any} new maze. We are interested in reproducing this capability of perfect generalization across all examples of a task, i.e., solving environments with infinitely large state spaces.
This makes it impossible to represent the MDP explicitly, i.e., by building a graph or table of all transitions.
Instead, the agent's models must be implemented as \emph{programs} that operate on the environment's state and action representations.
While the problem of constructing such models is challenging, we assume that the agent's learning process has already converged such that $\hat{T}$ and $\hat{R}$ reproduce the behavior of $T$ and $R$ respectively.
Thus, rather than focusing on learning, our objective in the current paper is to investigate ways that learned models can be used more effectively.

The magnitude of the state spaces we consider imposes many challenges, most notably the fact that when a new episode starts in some initial state $s_0$, it is likely that the agent has never observed any part of the region of the state space containing $s_0$. Additionally, we do not allow the agent to explore the episode's transition graph before attempting to make a plan. Instead, it is required to solve the task instance ``on its first try'', just as would be expected of a human presented with, e.g., a novel maze; the person typically would not be given the opportunity to explore the maze ``before solving it''.

This means that the planner cannot improve its performance by pre-computing information for that specific episode; any computation that depends on the initial state is counted as part of its planning time.
Thus, there is no benefit to maintaining any finite amount of state-specific information (e.g., solutions for mazes up to a certain size),
 as this will not help the agent when averaged over the entire range of possible task instances.
For example, computing the optimal value function for every sub-maze of size $100 \times 100$ -- an expensive endeavor -- will have little to no effect on an agent's performance on instances of size $1,000 \times 1,000$.
On the other hand, the agent can spend time \emph{within} an episode in order to prepare a sequence of actions. For example, all planning can be done as soon as $s_0$ is observed, then the rest of the actions for that episode can be emitted without delay.

\section{Milestones}

There are many interesting parts of the state space that an agent may want to visit for both exploration (learning) and exploitation (reward) purposes. In the current paper, we focus on the latter problem: achieving rewards -- efficiently. We specifically tackle domains with \emph{sparse} rewards, on which existing search methods perform poorly. 
For the current exposition, we assume that there is a single maximum reward value $r_{max} = \max_{s, a, s'}R(s, a, s')$ and that all other values are, essentially, penalties for the agent to avoid.
We define a \emph{milestone} to be a state at which the agent can earn this maximal reward, i.e., a milestone $g$ is a state such that
\begin{equation}\exists s \in S, a \in A \colon [T(s, a) = g] \land [R(s, a, g) = r_{max}]\text{.}\end{equation}
In many environments, it is also convenient (and equivalent) to define milestones as states \emph{from which} the agent can earn maximal reward by taking a particular action, i.e.,
\begin{equation}\exists a \in A, s' \in S \colon [T(g, a) = s'] \land [R(g, a, s') = r_{max}]\text{.}\end{equation}
For example, a milestone in a maze environment could be defined as either ``arrive at the goal'' or ``stand left of the goal and press right''.
Our method can be extended in a straightforward manner to more-complex scenarios, e.g., domains with a finite set of desirable reward values.

When defining the set of candidate milestones for a particular state, it is helpful to consider the notion of \emph{reachable future states}. We define $F(s)$ to be the set of states reachable from a state $s$ by any sequence of actions.
We then denote by $G^*(s)$ the set of all milestones reachable from $s$:
\begin{equation}G^*(s) = \{g \in F(s) \mid g \text{ is a milestone}\}\text{.}\end{equation}
This set defines an abstracted \emph{milestone space} for the agent to search over. However, many of these milestones may be impossible to reach without first passing through another milestone, e.g., in an environment that involves obtaining a sequence of rewards. Thus, it is also helpful to define the \emph{immediate milestones} set $G(s)$, which is the subset of $G^*(s)$ containing states that can be reached through \emph{some} trajectory that does not include another milestone. This can be thought of as the \emph{node expansion function} for the milestone space.

In the current paper, we consider agents that model $G(s)$ with a \emph{milestone enumeration function} $\hat{G}(s)$, which produces hypothetical milestone states for the agent to pursue.
Iterating on this function (i.e., feeding its output back in) yields an approximation of $G^*(s)$.
In order to enable more-efficient search, $\hat{G}(s)$ should
include as many of the elements of $G(s)$ as possible
while containing few extra states that are not in $G(s)$ (i.e., states that are unreachable or at which no reward can be gained).

\paragraph{State and Model Representations}
A key point to consider is that the form of an agent's models is determined by that agent's learning algorithm.
This means that the agent's milestone enumeration procedure need only be able to operate over the class of models learnable by that agent.
While this makes the choice of representation -- for both state and model -- highly consequential, we defer extended discussion of this topic to Section~\ref{sec:discussion}.
The important result of this fact for now is that we can design specialized milestone-enumeration algorithms for different kinds of models.
Thus, in the next sections, we assume that a suitable state representation has been chosen (or learned) then discuss introspection in the context of an appropriate class of models.

\section{Example: Search on a Grid}

\begin{figure*}[tb!] 
\begin{center}

     \begin{subfigure}[b]{0.32\textwidth}
     \centering
         \includegraphics[width=\textwidth]{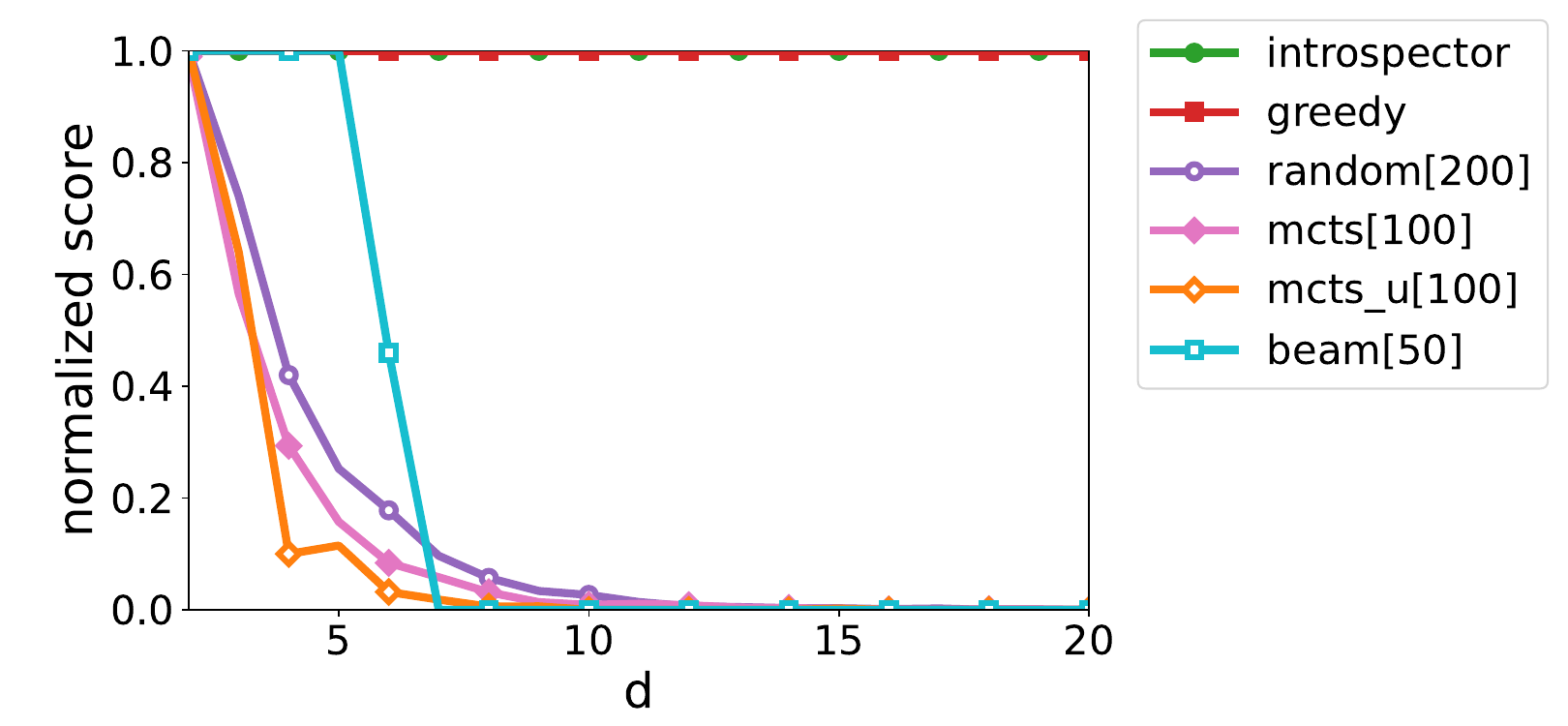}
         \caption{score}
         \label{subfig:grid-score}
     \end{subfigure}
     \hfill
     \begin{subfigure}[b]{0.32\textwidth}
     \centering
         \includegraphics[width=\textwidth]{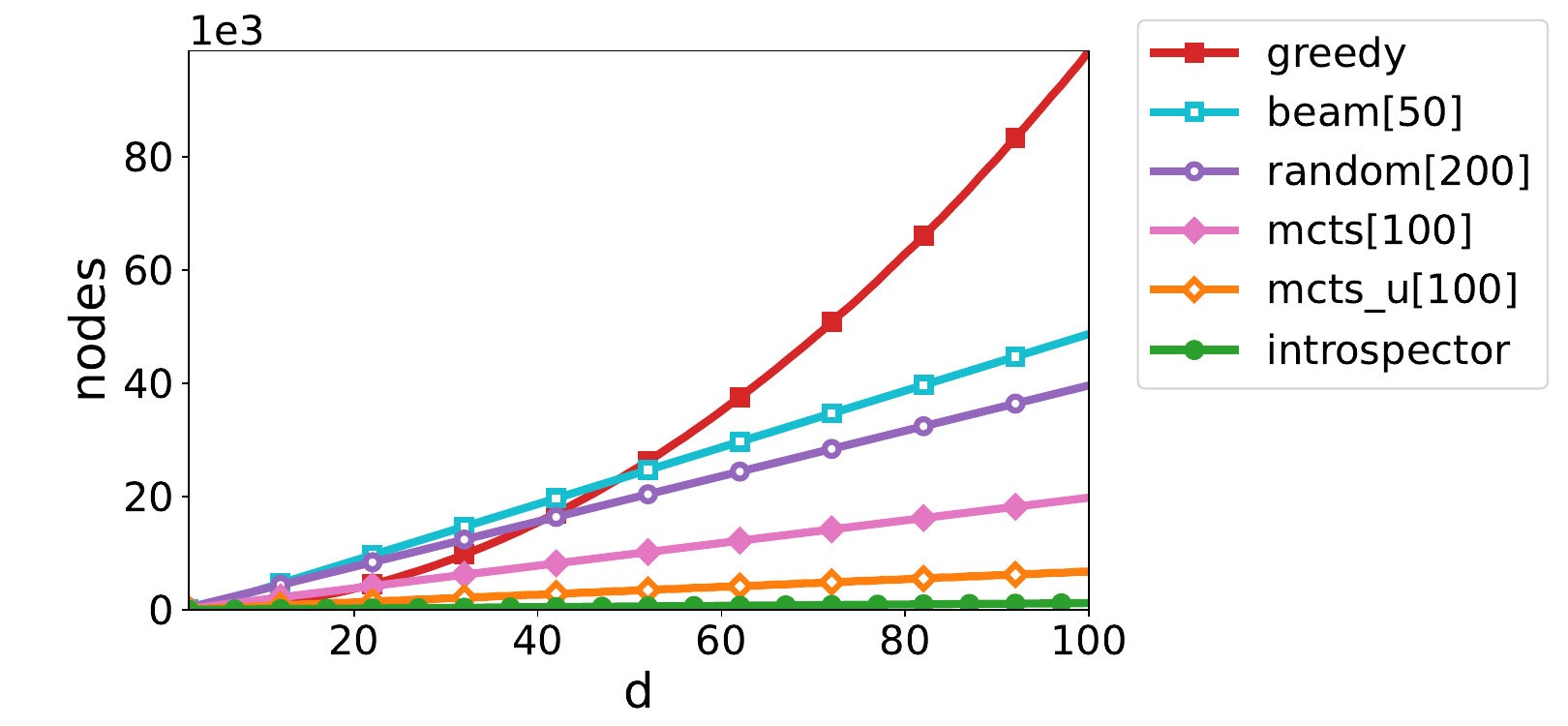}
         \caption{nodes expanded}
         \label{subfig:grid-nodes}
     \end{subfigure}
     \hfill
     \begin{subfigure}[b]{0.32\textwidth}
     \centering
         \includegraphics[width=\textwidth]{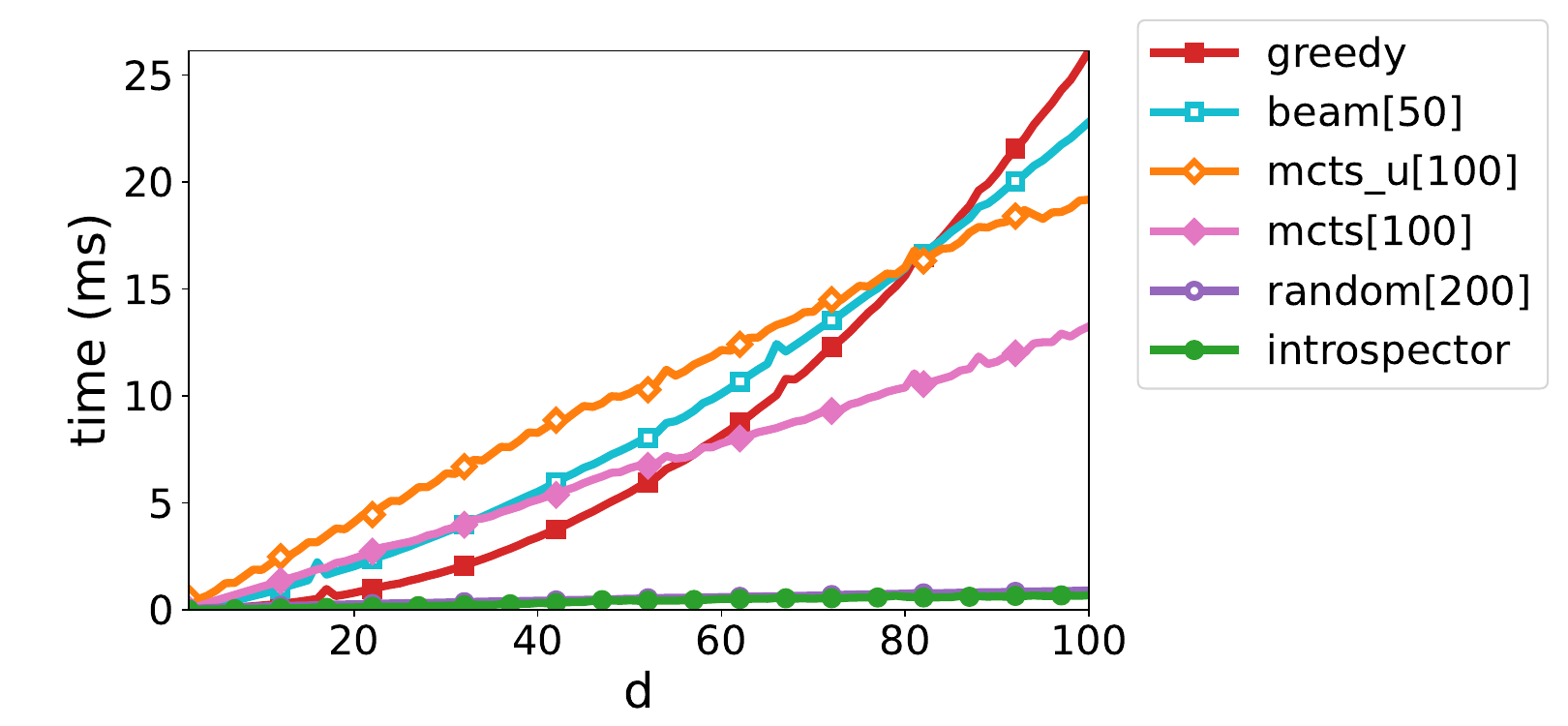}
         \caption{time}
         \label{subfig:grid-time}
     \end{subfigure}

\end{center}
\caption{Planning results in the grid search domain}
\Description{Depiction of results using three metrics for planning in the grid search domain: normalized score, number of nodes expanded, and time taken to plan.}
\label{fig:grid}
\end{figure*}

As an introduction to model introspection, 
consider an environment in which the agent exists on an infinite, empty grid. The state is represented using the player's integer coordinates, $s = (x, y) \in \mathbb{Z} \times \mathbb{Z}$, with initial states sampled from some arbitrary distribution. There are four actions, corresponding to movement by one unit in each cardinal direction. The actions are represented by vectors, e.g., \code{UP}~$=(0, 1)$, which are added to the state during transitions. Every action results in a penalty for the player, except that when the player arrives at the origin $(0, 0)$, they receive a positive reward. Episodes terminate after a fixed number of steps.

Suitable models for this environment can be expressed as programs consisting of a series of if-else statements, which we show using case notation:
\begin{align}
\hat{T}(s, a) &= s + a\label{eq:grid-T}\\
\hat{R}(s, a, s') &=
\begin{cases}
+1 & \text{if $s' = (0, 0)$}\\
-1 & \text{otherwise.}
\end{cases}\label{eq:grid-R}
\end{align}

Before applying model introspection, consider how a ``naive'' model-based planner must solve this problem.
Given black-box forward models $\hat{T}$ and $\hat{R}$, a typical planner would conduct \emph{uninformed} search over the state space, e.g., exhaustive greedy search,
simulating trajectories up to the episode's horizon.
If the initial state is $d$ steps away from the origin, then the planner necessarily expands $\Omega(d^2)$ states to guarantee optimality (i.e., the number of nodes up to $d$ steps away from the initial position), since it has no way to prune any branches of the search.
On the other hand, model-free agents can solve this environment both optimally and efficiently. For example, a policy that selects actions that move each coordinate towards zero can be computed in constant time per action (not dependent on $d$), yet it still yields an optimal $d$-step solution. Can we get similar results from a model-based planner?

The answer is \emph{yes}, though we will not provide an algorithm for computing milestones over this class of model; we reserve this level of analysis for the relational models in Section~\ref{sec:relational}. For now, let us \emph{informally} consider the information contained in \eqref{eq:grid-T} and \eqref{eq:grid-R}.
First, it is easy to see that there is only one state at which the agent receives maximal reward; thus, there is a unique milestone $g = (0, 0)$.
Second, the changes shown in the transition function clearly correspond to unit movement in each direction on the plane; thus, the Manhattan distance ($L_1$ norm) is a perfect estimate of the distance between any two states.
We can use this information to implement a \emph{milestone-directed planning algorithm}.

We implement our algorithm, \emph{introspector}, and compare against the following reward-maximization planning algorithms:
\texttt{GREEDY}, which implements exhaustive best-first search;
\texttt{RANDOM[K]}, a policy which chooses the best out of $K$ random plans;
\texttt{BEAM[K]}, width-$K$ beam search;
\texttt{MCTS[K]}, pure Monte Carlo tree search \cite{coulom07};
and
\texttt{MCTS-U[K]}, Monte Carlo tree search with the pUCT rule \cite{rosin10}.
In this environment, best-first search finds optimal plans, so the greedy planner provides a baseline for the typical cost of ensuring maximum score. On the other hand, the random planner serves as a reference point for the runtime efficiency of our introspector. \mbox{MCTS-U}, which is used in modern reinforcement-learning algorithms such as MuZero \cite{schrittwieser20}, represents a strong baseline planner that is typically more efficient than exhaustive best-first search.
We vary the agent's initial distance from the origin, $|x| + |y| = d$, to measure how each method fares as the search difficulty increases.
The length of each episode is $H = 2(d-1)$.
We report \emph{normalized scores} in $[0, 1]$, where zero is the worst possible result for an episode and one is the performance of an optimal plan. Averaged results (from 100 episodes per planner per $d$) are shown in Figure~\ref{fig:grid},
with semilog-$y$ versions of the plots included in Appendix~\ref{appendix:grid-search}.

The score plots cut off at $d=20$, as all planners other than introspector and greedy drop to zero; see Appendix~\ref{appendix:planner-scaling} for more results with varying computation budgets $K$.
While greedy search displays runtime growth that is super-linear in $d$, introspector maintains optimal plan quality with only linear runtime growth; this cannot be reduced further, since every planner must return a plan of length $H = \Theta(d)$.
This increase in efficiency -- both in actual runtime and in empirically-measured asymptotic growth -- demonstrates the potential of model introspection.
However, the milestone and heuristic in this example came from \emph{human} analysis.
In the next section, we resolve this issue by providing an \emph{automatic, domain-agnostic algorithm} for milestone enumeration over an important class of models. We then integrate this procedure, along with domain-agnostic heuristic calculation methods, into a planner that demonstrates an even more significant runtime improvement than shown here.

\section{Planning in Relational Domains}\label{sec:relational}

While the prior example served to motivate the utility of model introspection, the form of the model may seem somewhat simplistic. Although we note again that the specific representation of the model is actually dependent on the agent's learning algorithm, and simple models are an appropriate choice when they suffice, it is interesting to consider tasks with more-complex structure. Here, we discuss planning in environments that use first-order logical (i.e., relational) representations \citep{dzeroski01}, which are similar to those used in classical planning.

First-order logic (FOL), also called \emph{predicate logic}, allows us to express formulas that use boolean-valued funtions called \emph{predicates} to represent information about objects (and sets of objects).
A \emph{term} is either a constant (a, b, c, ...) or a variable ($X_0$, $X_1$, ...).
A \emph{literal} is a predicate applied to some term(s), e.g., $On(a, b)$, $OnTable(X_0)$.
A literal is called \emph{ground} if it contains no variable terms.
A \emph{formula} is a predicate constructed by combining literals using \emph{connectives} (and~$\land$, or~$\lor$, not~$\neg$) and \emph{quantifiers} (existential~$\exists$ and universal~$\forall$). Rather than restricting to a normal form, we allow formulas of arbitrary construction.
These formulas can be learned from examples using Inductive Logic Programming methods \cite{driessens01, cropper22}.

In relational reinforcement learning (RRL), states are represented using a set of objects (i.e., constants) and the set of ground literals (with terms ranging over the state's objects) that are true in that state.
Like in classical planning, \emph{action schemas} taking zero or more object parameters are used. Each schema has three components: the \emph{precondition}, an arbitrary formula that determines whether the action can be applied in a given state; the \emph{add list}, a set of literals that become true after the action is taken; and the \emph{delete list}, a set of literals that become false after the action is taken.
We represent the environment's reward signal using a \emph{FOL decision list}, which encodes a series of if-else statements whose conditions are first-order logic formulas.
We also include a \emph{termination signal} $C(s, a, s')\to \{-1, 0, 1\}$ (meaning failure, continue, and success, respectively), also implemented as a FOL decision list, which determines when the episode ends.
Although a fixed horizon would work as well, this plan-dependent termination mechanism makes it easier to study the time complexity of each planning algorithm.

\begin{algorithm2e}[tb!]
\SetInd{0.7em}{0.7em}

\SetKwProg{Fn}{Func}{}{}

\SetKwProg{Action}{Action}{}{}
\SetKw{Predicates}{predicates:}
\SetKwInOut{Pre}{pre}
\SetKwInOut{Add}{add}
\SetKwInOut{Del}{del}

\tcp{The notation below for predicate declaration gives name/arity}
\Predicates{OnTable/1, On/2, Clear/1, HandEmpty/0, Holding/1}

\BlankLine
\BlankLine
\BlankLine

\tcp{Pick a block off the table into the hand}
\Action{Pick(X)}{

	\Pre{OnTable(X) $\land$ Clear(X) $\land$ HandEmpty()}
	
	\Add{Holding(X)}
	
	\Del{OnTable(X), Clear(X), HandEmpty()}
	
}

\BlankLine

\tcp{Place a block from the hand onto the table}
\Action{Place(X)}{

	\Pre{Holding(X)}
	
	\Add{OnTable(X), Clear(X), HandEmpty()}
	
	\Del{Holding(X)}
	
}

\BlankLine

\tcp{Place a block from the hand onto another block $Y$}
\Action{Stack(X, Y)}{

	\Pre{Holding(X) $\land$ Clear(Y)}
	
	\Add{On(X, Y), Clear(X), HandEmpty()}
	
	\Del{Holding(X), Clear(Y)}
	
}

\BlankLine

\tcp{Pick a block off another block $Y$ into the hand}
\Action{Unstack(X, Y)}{

	\Pre{On(X, Y) $\land$ Clear(X) $\land$ HandEmpty()}
	
	\Add{Holding(X), Clear(Y)}
	
	\Del{On(X, Y), Clear(X), HandEmpty()}
	
}

\BlankLine
\BlankLine
\BlankLine

\tcp{The agent gets a reward once all blocks are on the table}
\tcp{(not stacked on another block)}
\Fn{R($s$, $a$, $s'$)}{
	\BlankLine
	\If{$\forall X \in s'\text{.constants}\colon OnTable(X)$}{
		\Return{$+1$}
	}
	\BlankLine
	\Return{$0$}
}

\BlankLine
\BlankLine
\BlankLine

\tcp{Similarly to the reward}
\tcp{the episode terminates successfully once all blocks are on the table}
\Fn{C($s$, $a$, $s'$)}{
	\BlankLine
	\If{$\forall X \in s'\text{.constants}\colon OnTable(X)$}{
		\Return{\code{SUCCESS}} \myc{The episode is over}
	}
	\BlankLine
	\Return{\code{CONTINUE}} \myc{The agent continues to take actions}
}

\BlankLine

\caption{Definition of the Blocks-World domain for relational reinforcement learning}
\label{alg:blocks-world}
\end{algorithm2e}

\paragraph{Example Environment}
Algorithm~\ref{alg:blocks-world} includes definitions of the action shemas, reward function, and termination function for the Blocks-World domain \cite{slaney01}. Episodes begin with blocks in random stacks and the hand empty.
Actions either pick up a block (from the top of a stack or the table) or place it (on top of a stack or on the table).
The agent receives a single reward (and the episode terminates successfully) once all blocks are on the table.

\subsection{Milestone Computation}

When formalizing the concept of milestones, we introduced the milestone enumeration function $\hat{G}(s)$, which produces hypothetical future states.
However, it is often more efficient to just describe the necessary \emph{changes} to the agent's current state.
We thus formulate a procedure called \emph{state mutation} of the form
\begin{equation}
(s, \hat{R}) \to \{\delta_1, \delta_2, ...\}\text{,}
\end{equation}
where $s$ is the agent's current state, $\hat{R}$ is the reward program, and each $\delta$ in the output set specifies changes to $s$ which, if achieved, will result in the agent obtaining reward $r_{max}$.

For RRL, state mutation takes a state (consisting of a set of true literals) and a formula (which, if satisfied, will lead the FOL decision list $\hat{R}$ to output $r_{max}$) and produces a set of \emph{mutations}, each consisting of a set of literals to make true and a set of literals to make false.
Each mutation represents a distinct way of satisfying the formula. 
In particular, mutations comprise a minimal set of necessary changes, constraining only some of the state's literals. Because of this, a single mutation may correspond to one or more milestone states, which can be found through state-space search.
Note that not all literals can be modified; e.g., any literal that does not show up in the add (or delete) list of any action cannot be changed from its initial value in~$s$. Thus, the formula may be impossible to satisfy, or it may be \emph{immutably valid} -- meaning that it is true in~$s$ and it cannot be made false through any action.
To account for this, we also include a special mutation object that denotes that a formula is immutably valid (a mutation comprising two empty sets is taken to mean that the formula is unsatisfiable).

Our formulas are constructed recursively out of six possible building blocks (literals, and, or, not, existential quantifier, universal quantifier). Similarly, we implement state mutation recursively, with a separate specialization for each type of formula element. The function signature is
\begin{equation}
\text{mutateT(State, Formula) $\to$ set[Mutation],}
\end{equation}
which returns a set of possible mutations that will make the formula true.
To support negation, we also implement \code{mutateF}, which computes mutations that will make a formula false. However, since this can be derived from the definitions we give for \code{mutateT} by inversion (and using De Morgan's laws), we omit the description here. The following paragraphs outline the mutation procedure at a high level; we have included pseudocode in Appendix~\ref{appendix:pseudocode}.

\paragraph{Literal} If a literal $l$ is mutable (can be modified by some action), then a single Mutation consisting of $+l$ (ensure that $l$ true) is returned. Otherwise, if $l$ is already true (but not mutable), the special value \code{IMMUTABLY VALID} is returned. If neither holds, then an empty set is returned (unsatisfiable).

\paragraph{And} A formula $f = f_1 \land f_2 \land ... \land f_n$ is satisfied when all of its children are satisfied. Thus, we calculate the set of Mutations for each child formula, then construct a new set of Mutations by picking every way of satisfying each child (i.e., a Cartesian Product).

\paragraph{Or} A formula $f = f_1 \lor f_2 \lor ... \lor f_n$ is satisfied when any of its children are satisfied. We calculate the set of Mutations for each child formula, then return the union of these sets.

\paragraph{Not} The formula $f = \neg f'$ is true when $f'$ is false. We simply call \code{mutateF} on the child formula $f'$.

\paragraph{Existential} The formula $f = \exists X\colon f'(X)$ is satisfied when $f'$ is satisfied for \emph{at least one} possible binding of the variable $X$. We perform each substitution of $X$ with an object in the state $s$, calculating the set of Mutations for each resulting formula. We then combine the sets using a union, similarly to how \code{or} formulas are mutated.

\paragraph{Universal} The formula $f = \forall X\colon f'(X)$ is satisfied when $f'$ is satisfied for \emph{all possible} bindings of the variable $X$. We perform each substitution of $X$ with an object in the state $s$, calculating the set of Mutations for each resulting formula. We then combine the sets using a Cartesian Product, similarly to how \code{and} formulas are mutated.

\subsection{Relational State Mutation Example}

\begin{figure}[tb!] 
\begin{center}

     \begin{subfigure}[b]{\columnwidth}
     \centering
         \includegraphics[width=0.8\textwidth]{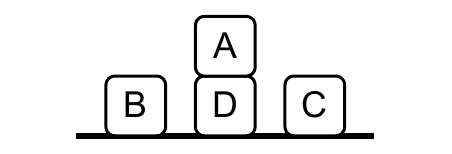}
         
         On(A, D), OnTable(B), OnTable(C), OnTable(D),
         
         Clear(A), Clear(B), Clear(C), HandEmpty()
         \caption{example state}
         \label{subfig:blocks-world-state}
     \end{subfigure}

     \begin{subfigure}[b]{\columnwidth}
     \centering
         \includegraphics[width=0.8\textwidth]{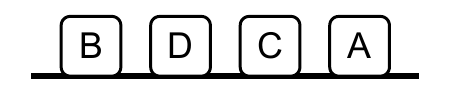}
         
         OnTable(A), OnTable(B), OnTable(C), OnTable(D),
         
         Clear(A), Clear(B), Clear(C), Clear(D), HandEmpty()
         \caption{corresponding milestone}
         \label{subfig:blocks-world-milestone}
     \end{subfigure}
     
\end{center}
\caption{Example Blocks-World state and milestone}
\Description{Example state visualization for the Blocks-World domain, showing blocks in various arrangements on a surface. The states are also described in plain text using the relational representation.}
\label{fig:blocks-world-milestone}
\end{figure}

Consider the example Blocks-World state shown in Figure~\ref{subfig:blocks-world-state}.
In this domain, the agent receives a reward once the condition
\begin{equation}
\forall X \in s'\text{.constants}\colon OnTable(X)
\end{equation}
holds, i.e., when all blocks are on the table (rather than stacked on another block). For the example state, this condition is satisfied by the milestone shown in Figure~\ref{subfig:blocks-world-milestone}, which corresponds to the state mutation $\{$\textbf{+OnTable(A), +OnTable(B), +OnTable(C), +OnTable(D)}$\}$ since this is the minimal information necessary to satisfy the reward predicate. The truth values of the other facts in the state, such as \textbf{Clear(A)}, are determined by the environment (and can be ignored during state mutation and heuristic computation).
A longer example in another domain is included in Appendix~\ref{appendix:rrl-example}.

\subsection{Planning Algorithm}

We integrate the state mutation procedure into a bilevel planning algorithm. The outer level of the planner conducts greedy search through milestone space by computing mutations and passing them to an informed-search subroutine;
pseudocode for this procedure is given in Appendix~\ref{appendix:rrl-planner}.
The inner planner uses heuristic search with the STRIPS literal-counting heuristic \citep{fikes71}
to efficiently find trajectories that satisfy one of the specified mutations. Note that the state mutation procedure is included in the planner's runtime since, as mentioned earlier, we do not allow pre-computation. In addition, our algorithm requires no manual configuration; unlike with the grid-search example, our RRL \emph{introspector} computes milestones (and heuristics) in a fully automatic, domain-agnostic manner.

\subsection{Experiments}\label{subsec:rrl-experiments}

We conduct experiments in three RRL domains. To focus on the runtime efficiency of the planners, the reward structure of these domains requires all rewards to be obtained before the episode can successfully terminate. There are also no penalties (other than termination with failure). Thus, we do not report scores, as they are the same for any valid plan.
As other methods do not always find these plans, we compare our \emph{introspector} algorithm against exhaustive greedy search; both solve every task instance successfully.

\begin{figure}[tb!] 
\begin{center}
     \begin{subfigure}[b]{0.49\columnwidth}
     \centering
         \includegraphics[width=\textwidth]{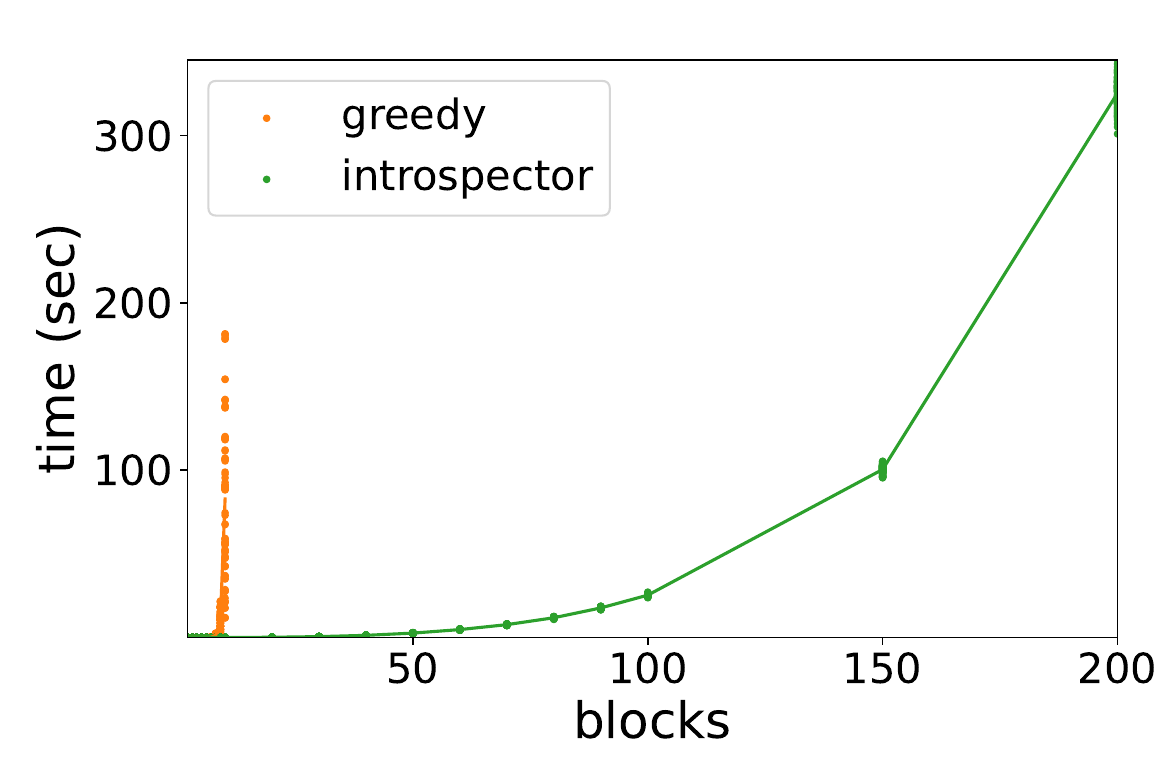}
         \caption{blocks-world}
         \label{subfig:rrl-blocks-world}
     \end{subfigure}
     \hfill
     \begin{subfigure}[b]{0.49\columnwidth}
     \centering
         \includegraphics[width=\textwidth]{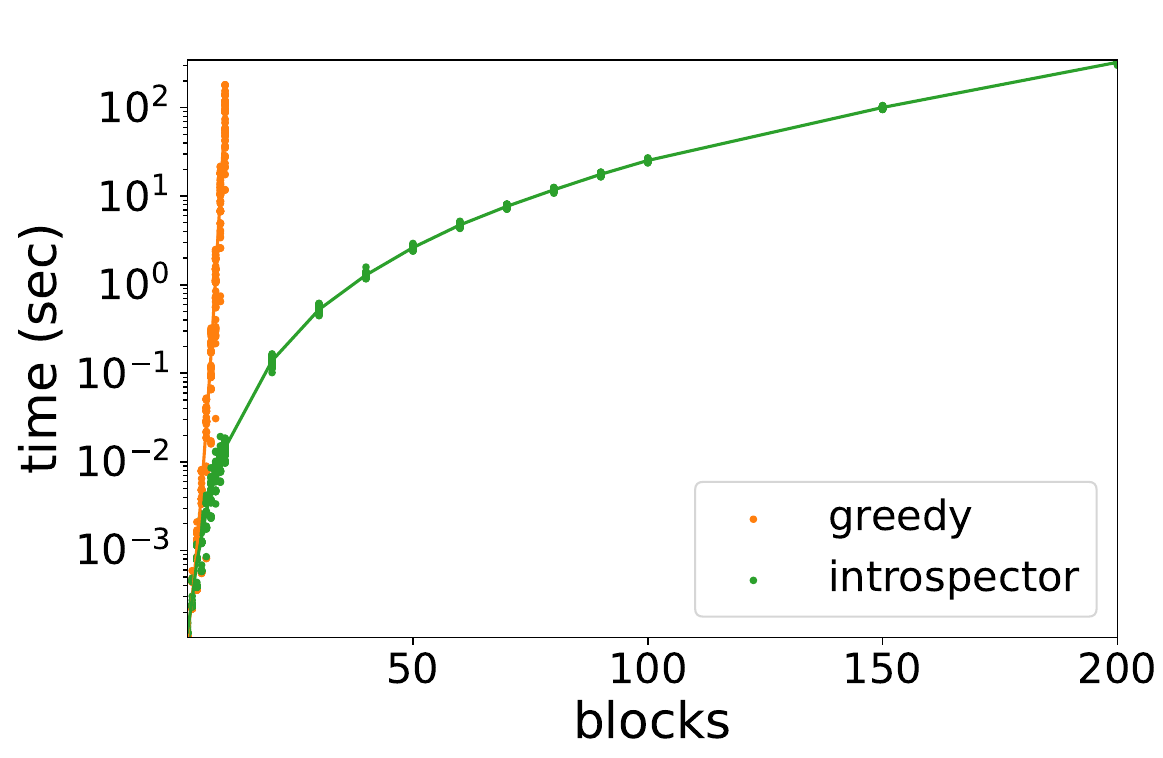}
         \caption{blocks-world (log-$y$)}
         \label{subfig:rrl-blocks-world-logy}
     \end{subfigure}
     
     \begin{subfigure}[b]{0.49\columnwidth}
     \centering
         \includegraphics[width=\textwidth]{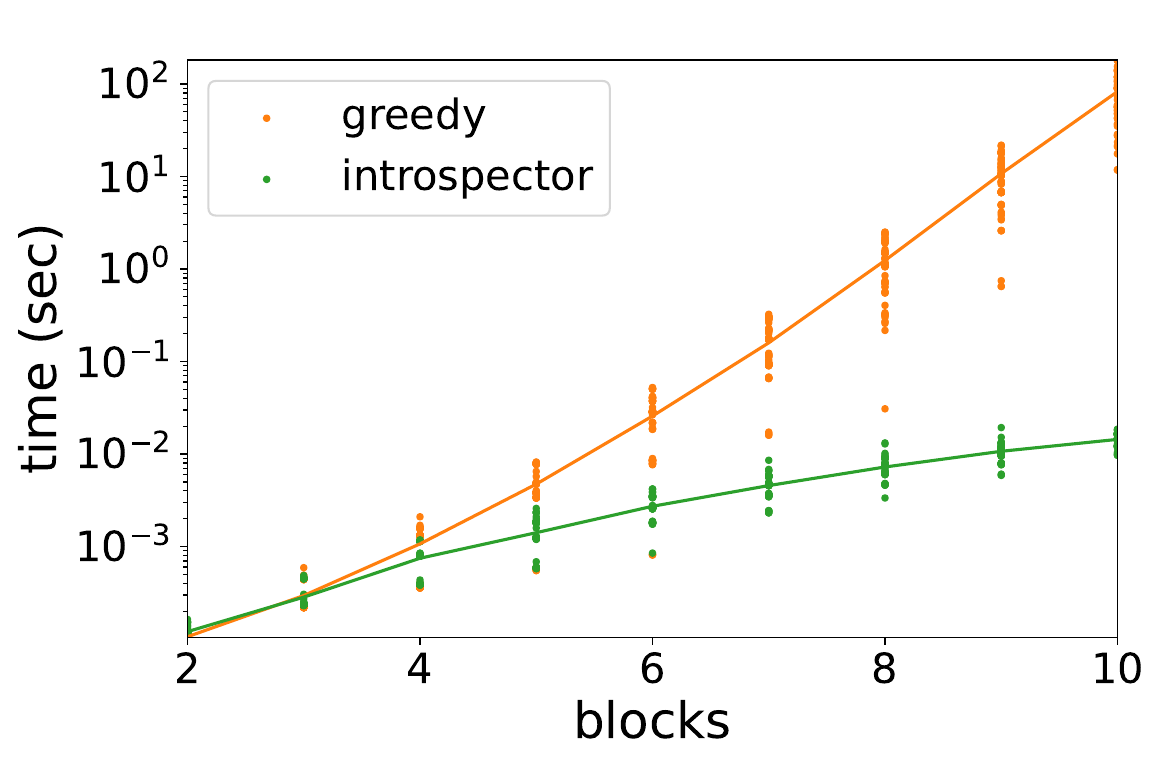}
         \caption{blocks-world planner runtime, zoomed in (logy-$y$)}
         \label{subfig:rrl-blocks-world-zoom}
     \end{subfigure}
     \hfill
     \begin{subfigure}[b]{0.49\columnwidth}
     \centering
         \includegraphics[width=\textwidth]{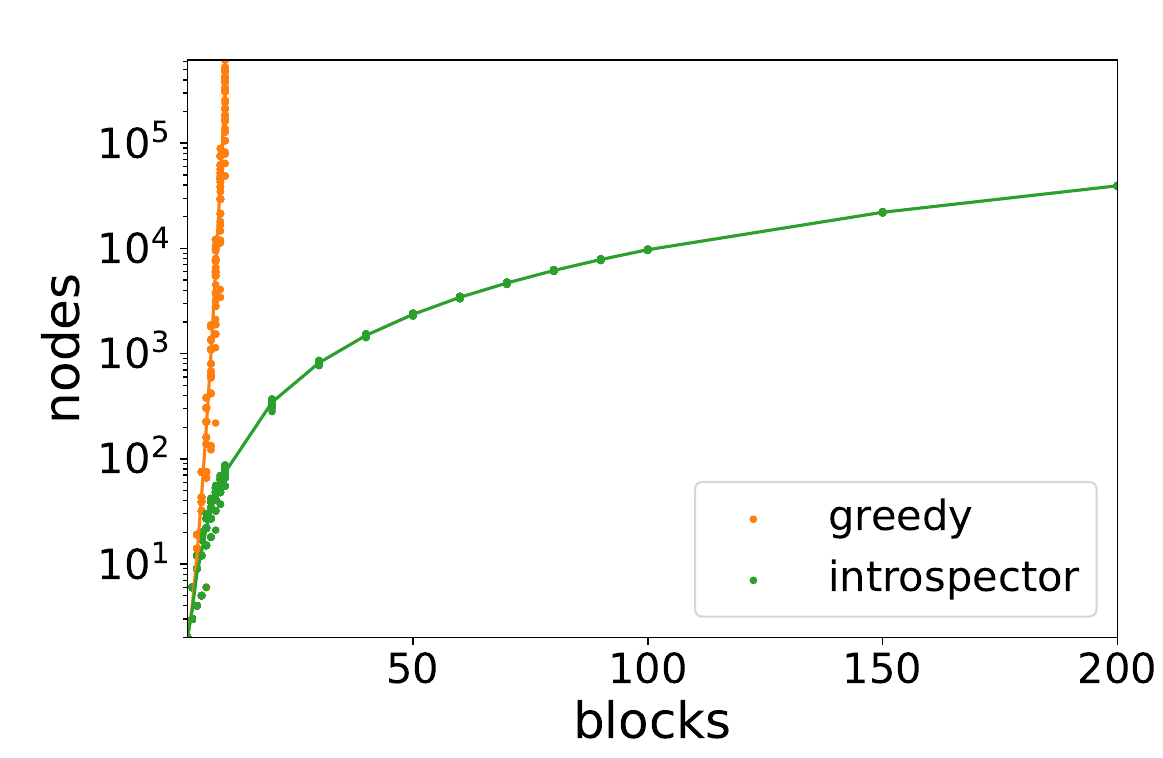}
         \caption{blocks-world, number of nodes expanded (log-$y$)}
         \label{subfig:rrl-blocks-world-nodes}
     \end{subfigure}
     
     \begin{subfigure}[b]{0.49\columnwidth}
     \centering
         \includegraphics[width=\textwidth]{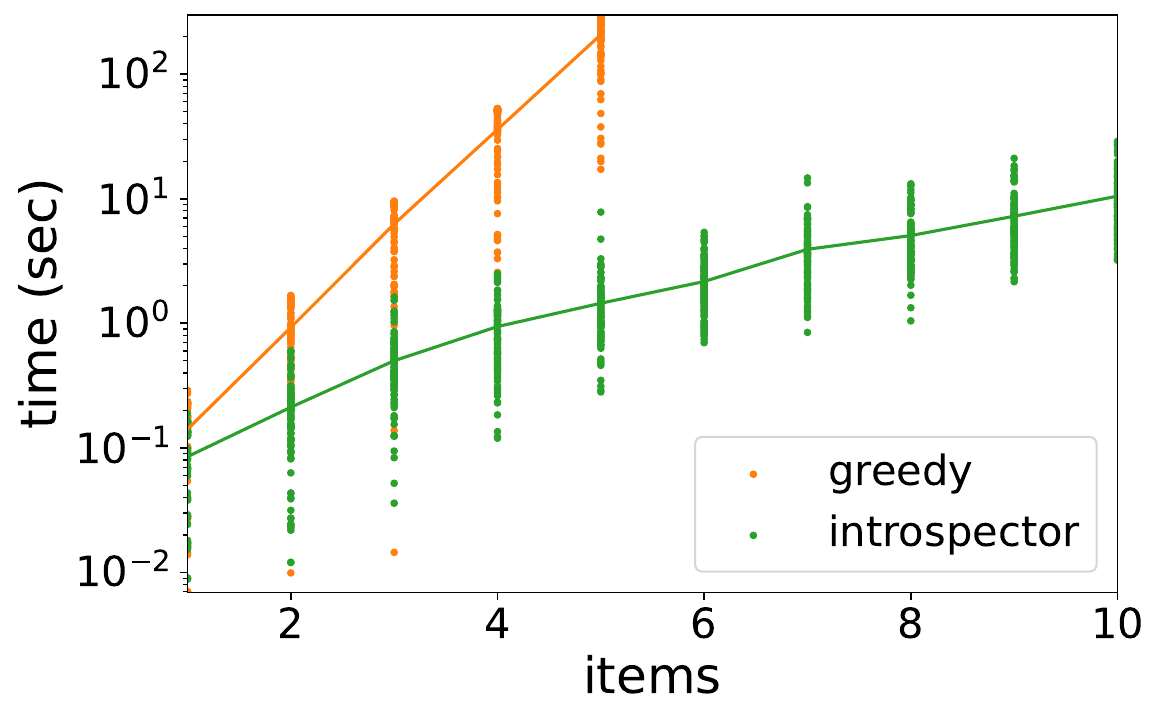}
         \caption{drawers, $n = 3$ (log-$y$)}
         \label{subfig:rrl-drawers-3-logy}
     \end{subfigure}
     \hfill
     \begin{subfigure}[b]{0.49\columnwidth}
     \centering
         \includegraphics[width=\textwidth]{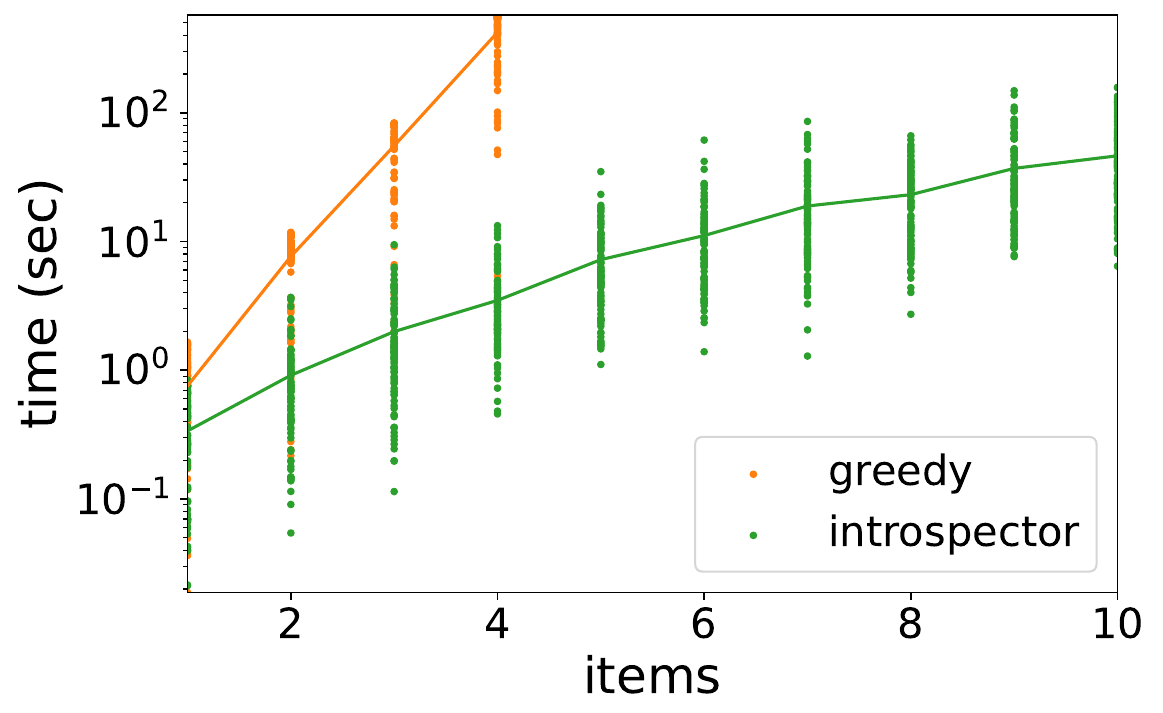}
         \caption{drawers, $n = 4$ (log-$y$)}
         \label{subfig:rrl-drawers-4-logy}
     \end{subfigure}
     
     \begin{subfigure}[b]{0.49\columnwidth}
     \centering
         \includegraphics[width=\textwidth]{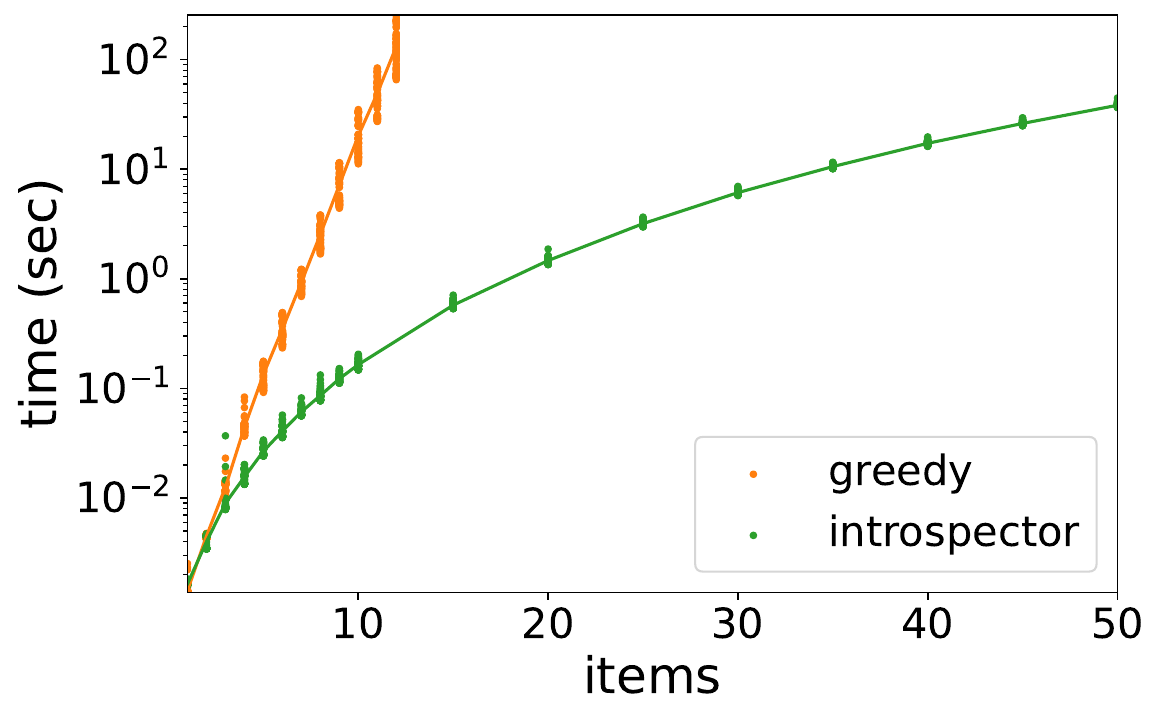}
         \caption{bins, $n = 2$ (log-$y$)}
         \label{subfig:rrl-bins-2-logy}
     \end{subfigure}
     \hfill
     \begin{subfigure}[b]{0.49\columnwidth}
     \centering
         \includegraphics[width=\textwidth]{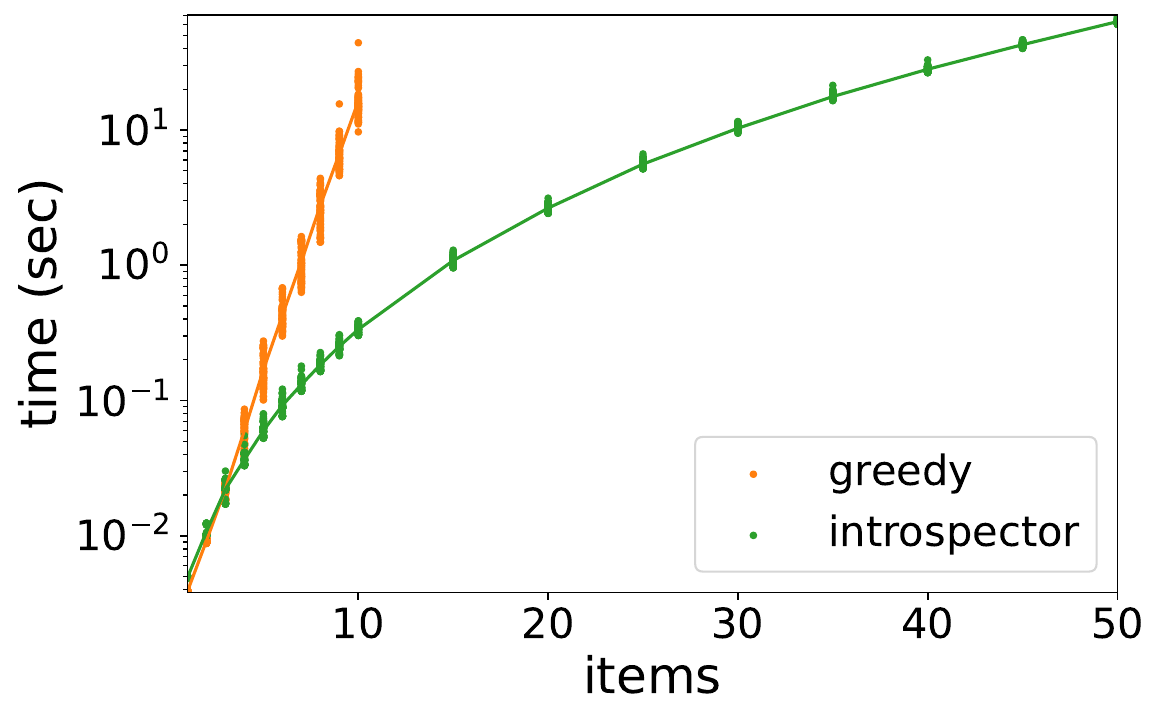}
         \caption{bins, $n = 3$ (log-$y$)}
         \label{subfig:rrl-bins-3-logy}
     \end{subfigure}
     
\end{center}
\caption{Planning results in relational domains}
\Description{Planning results in our three relational domains (blocks-world, drawers, bins) showing that our introspection-based planner performs significantly better than greedy search.}
\label{fig:rrl}
\end{figure}

The first task is unstacking blocks in the Blocks-World domain. 
When creating an episode with $n$ blocks, our initial state generator makes approximately $O(\sqrt{n}$) stacks on average,
meaning that the shortest solution plan has $\Omega(n-\sqrt{n})=\Omega(n)$ actions.
However, the reachable state space is \emph{massive} \citep{buligiu20},
and the number of available actions is proportional to the number of stacks -- which gets \emph{larger} as the agent solves the task.
Thus, as shown in Figures~\ref{subfig:rrl-blocks-world-logy} and~\ref{subfig:rrl-blocks-world-zoom}, the runtime of exhaustive search grows \emph{super-exponentially} in the number of blocks.
In contrast, our milestone-based planner demonstrates \emph{sub-exponential runtime growth}. Figure~\ref{subfig:rrl-blocks-world-nodes} shows that these trends also hold for the number of nodes expanded by each planner.

The second task is placing items in drawers. Every item and each of the $n$ drawers is given a label in $\{1, ..., n\}$. Items begin in random locations (any drawer or the ``shelf'') and must be moved into the drawer with the same label. The singular reward state (and termination state) is when all items are in the correct drawer \emph{and} all drawers are closed. Due to the large state space and the number of actions available at each step, uninformed search takes time exponential in the number of items, as shown in Figures~\ref{subfig:rrl-drawers-3-logy} and~\ref{subfig:rrl-drawers-4-logy}. However, we again observe that \emph{introspector} is able to construct successful plans in sub-exponential time.

While the prior domains only admit a single reward per episode, milestones can be generated and used multiple times within a single task instance. In our third domain, $n$ bins begin with items randomly distributed among them. The agent must take all items out of every bin, placing them on a shelf. Each time a bin is emptied \emph{and closed}, the agent gets a reward; the episode terminates successfully when all bins are empty and closed. Figures~\ref{subfig:rrl-bins-2-logy} and~\ref{subfig:rrl-bins-3-logy} show, again, that our planning algorithm is able to achieve sub-exponential runtime, this time by guiding its search along several milestones in succession.
For an extended discussion of this domain and an example task instance, see Appendix~\ref{appendix:rrl-example}.
Altogether, these results demonstrate that use of model introspection can yield a massive improvement over existing methods for planning in reinforcement learning.
Additional results and metrics are included in Appendix~\ref{appendix:rrl-results}.

\section{Discussion and Future Work}\label{sec:discussion}

The purpose of this paper is to motivate \emph{model introspection}, i.e., program analysis over an agent's learned models, to enable human-like reasoning and knowledge synthesis.
We investigated the application of this idea to planning in reinforcement learning, yielding algorithms that are more efficient and effective than prior work. In particular, our \emph{introspector} algorithm achieves a speedup over existing methods \emph{from (super-)exponential to sub-exponential runtime} in several relational reinforcement learning environments.
However, the potential utility of introspection reaches far beyond the applications shown in the current paper.
In addition, the introduction of program analysis to reinforcement learning brings a new perspective on the importance of \emph{representation}.
These ideas set the stage for a variety of future work, several aspects of which we now discuss in more detail.

\subsection{Representations and Interpretability}

One key insight to note is that this type of analysis \emph{requires} that the models be interpretable, though this does not mean that the agent must use one of the representations we discussed. The structure of the models is dependent on the agent's learning algorithm, so analysis of the model's interpretation can be specific to the type of models the agent learns. We thus encourage future work in two related directions.
The first is the development of new types of models, along with the corresponding algorithms to conduct automated analysis and produce milestones.
This may include the integration of general program analysis methods such as \emph{symbolic execution} \cite{godefroid05, cadar08, ma11}.
The second is further study of existing models to determine whether this kind of introspection can be applied. For example, reinforcement learning is often tackled with ``black-box'' deep neural networks, which are notoriously difficult to interpret; the development of introspection algorithms for this class of models would have a profound impact on the field.

Another important point is that the state and action representations described in this paper are typically considered \emph{high-level}, i.e., not able to be provided directly by a sensing system. We propose that one goal of an agent's perceptual processing module should be to extract these kinds of representations. Specifically, an ideal representation should allow the agent to, among other things: (1) describe the environment's dynamics compactly, (2) estimate the distance between states, and (3) construct plausible future states.
Beyond this, our paper demonstrates an important fact that has often been overlooked in prior work:
the importance of \emph{representation learning} applies not just to the features of the state representation, but also to \emph{the structure of the agent's models}. In other words, the form of the programs used to represent the environment's dynamics is just as important as the form of the data used to represent its states and actions.

\subsection{Other Applications of Introspection}

In the current paper, we defined milestones as \emph{states where the agent will receive a (maximal) reward}. However, there are many other kinds of future states that could be computed.
For instance, while an agent is still learning about an environment, it may have several hypotheses about the nature of $T$ in the form of a set of models.
The agent would like to gather information to distinguish between these different models, but random actions -- which are often used for exploration -- may lead to transitions on which many of the hypotheses agree, meaning that the agent gains little to no information. However, if the agent could aim towards transitions for which the models produce distinct outputs -- i.e., run \emph{intentional experiments} -- then learning could potentially be accelerated.

More broadly, the introspection process can be thought of as a kind of \emph{program synthesis}, which takes as input the agent's model(s) and produces another program as output. Here, we described the synthesis of the milestone enumeration function. The automatic derivation of other kinds of programs is a promising direction for future work. In particular, novel methods for heuristic function synthesis (based on $\hat{T}$) could enable yet-more-efficient search. Beyond this, it may be possible to derive value functions (or even entire policy programs) by analyzing the transition and reward models together.
This leads to a very general view of what introspection may enable: 
while our \emph{introspector} planning algorithm uses milestones within the framework of heuristic search,
future research may enable the synthesis of \emph{entire solution algorithms}, whether search-based or not.

\subsection{Deriving and Verifying Optimal Solvers}

The possibility of deriving efficient solvers makes \emph{verification} an interesting avenue of study.
This includes verification of \emph{correctness}, i.e., that the agent's policy always leads to successful termination, and \emph{optimality}, i.e., that the agent always finds the best solution.
Perhaps more importantly, the results of our experiments indicate that a solver's \emph{asymptotic time complexity} can be improved \emph{while maintaining solution quality} by using model introspection.
We conjecture that, with further research, it will be possible to create methods that can \emph{efficiently and automatically derive runtime-optimal solvers} for certain classes of problems.

\balance

\bibliographystyle{ACM-Reference-Format} 
\bibliography{references}


\newpage
\appendix
\onecolumn

\clearpage
\section{Grid Search Results}\label{appendix:grid-search}

\begin{figure}[h!] 
\begin{center}

     \begin{subfigure}[b]{0.49\textwidth}
     \centering
         \includegraphics[width=\textwidth]{plots/plot_grid_score}
         \caption{score}
     \end{subfigure}
     \hfill
     \begin{subfigure}[b]{0.49\textwidth}
     \centering
         \includegraphics[width=\textwidth]{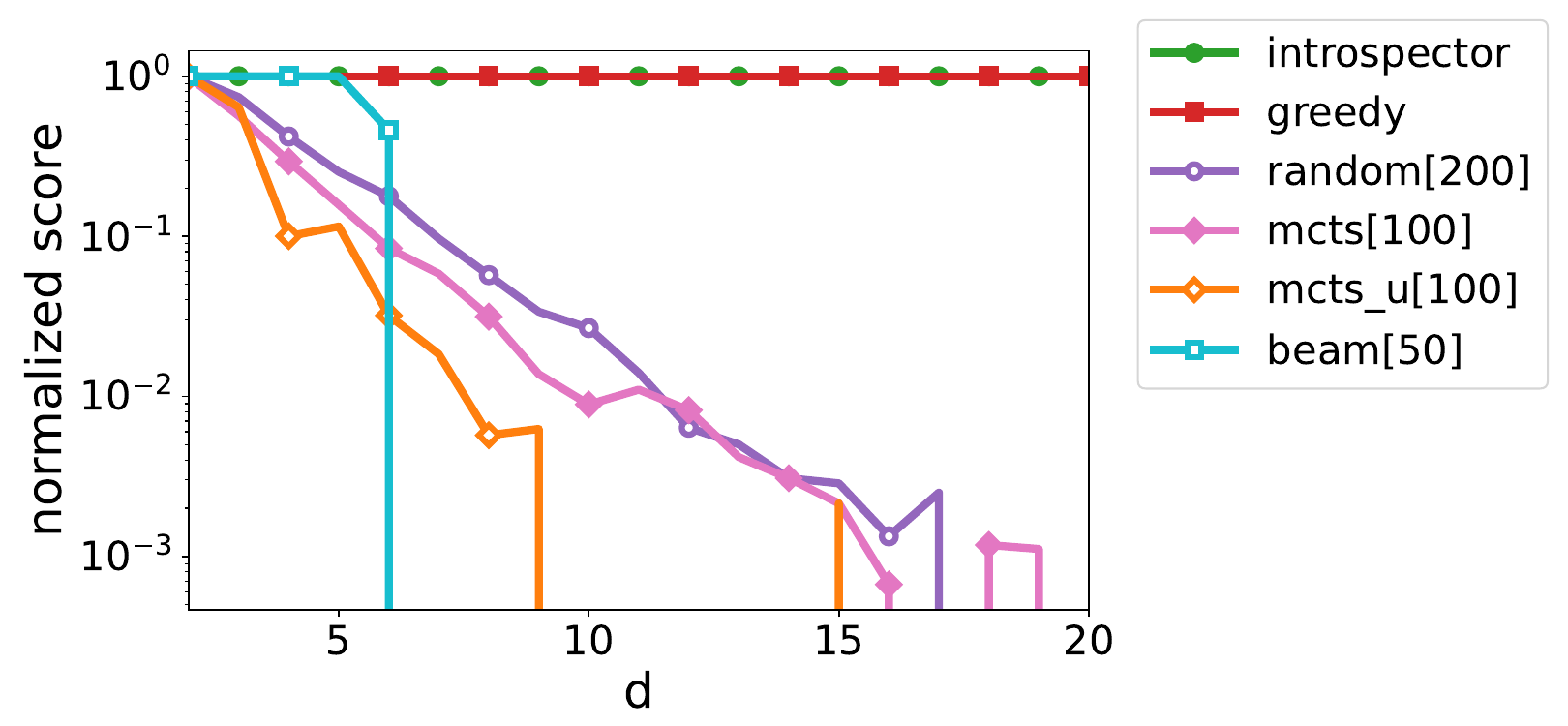}
         \caption{score (log-$y$)}
     \end{subfigure}
     
     \begin{subfigure}[b]{0.49\textwidth}
     \centering
         \includegraphics[width=\textwidth]{plots/plot_grid_nodes}
         \caption{nodes expanded}
     \end{subfigure}
     \hfill
     \begin{subfigure}[b]{0.49\textwidth}
     \centering
         \includegraphics[width=\textwidth]{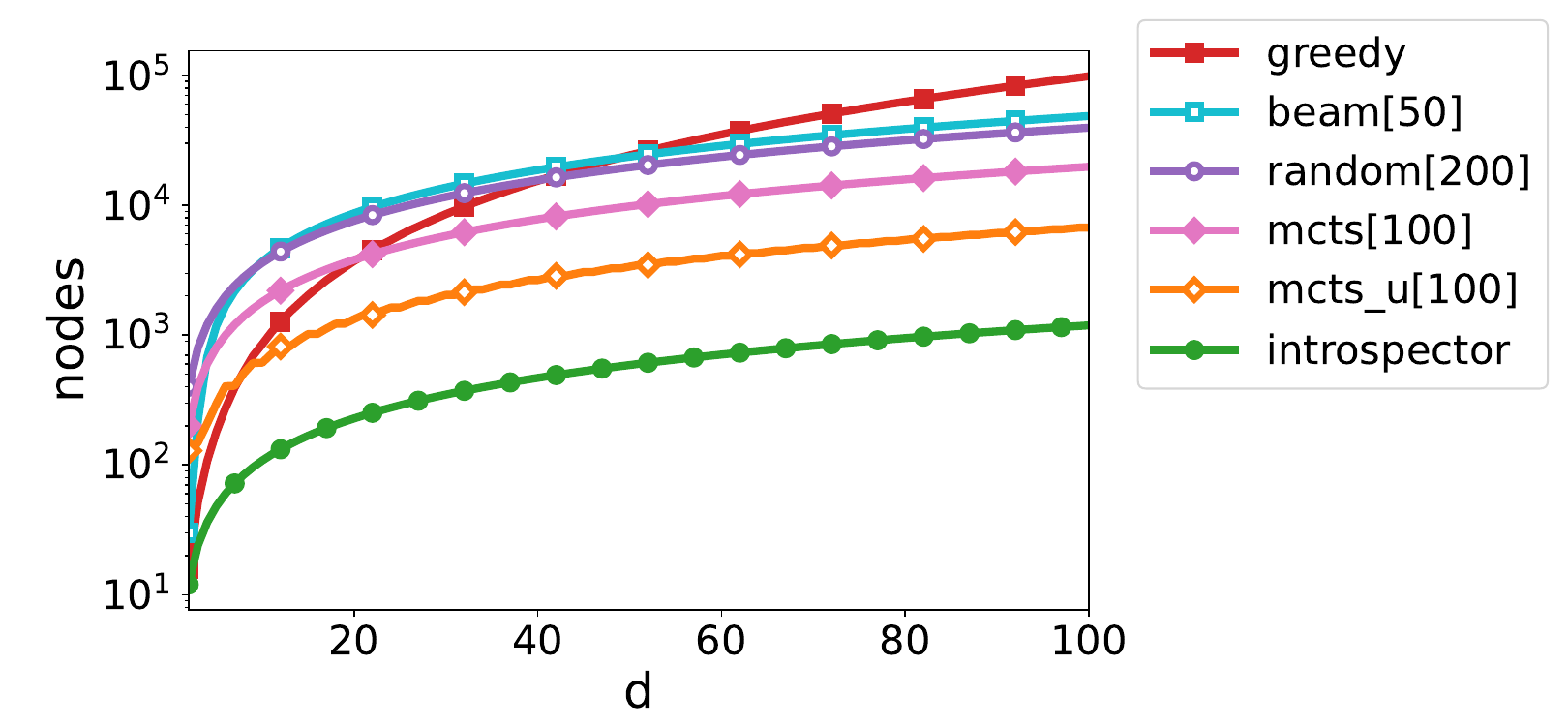}
         \caption{nodes expanded (log-$y$)}
     \end{subfigure}
     
     \begin{subfigure}[b]{0.49\textwidth}
     \centering
         \includegraphics[width=\textwidth]{plots/plot_grid_time}
         \caption{time}
     \end{subfigure}
     \hfill
     \begin{subfigure}[b]{0.49\textwidth}
     \centering
         \includegraphics[width=\textwidth]{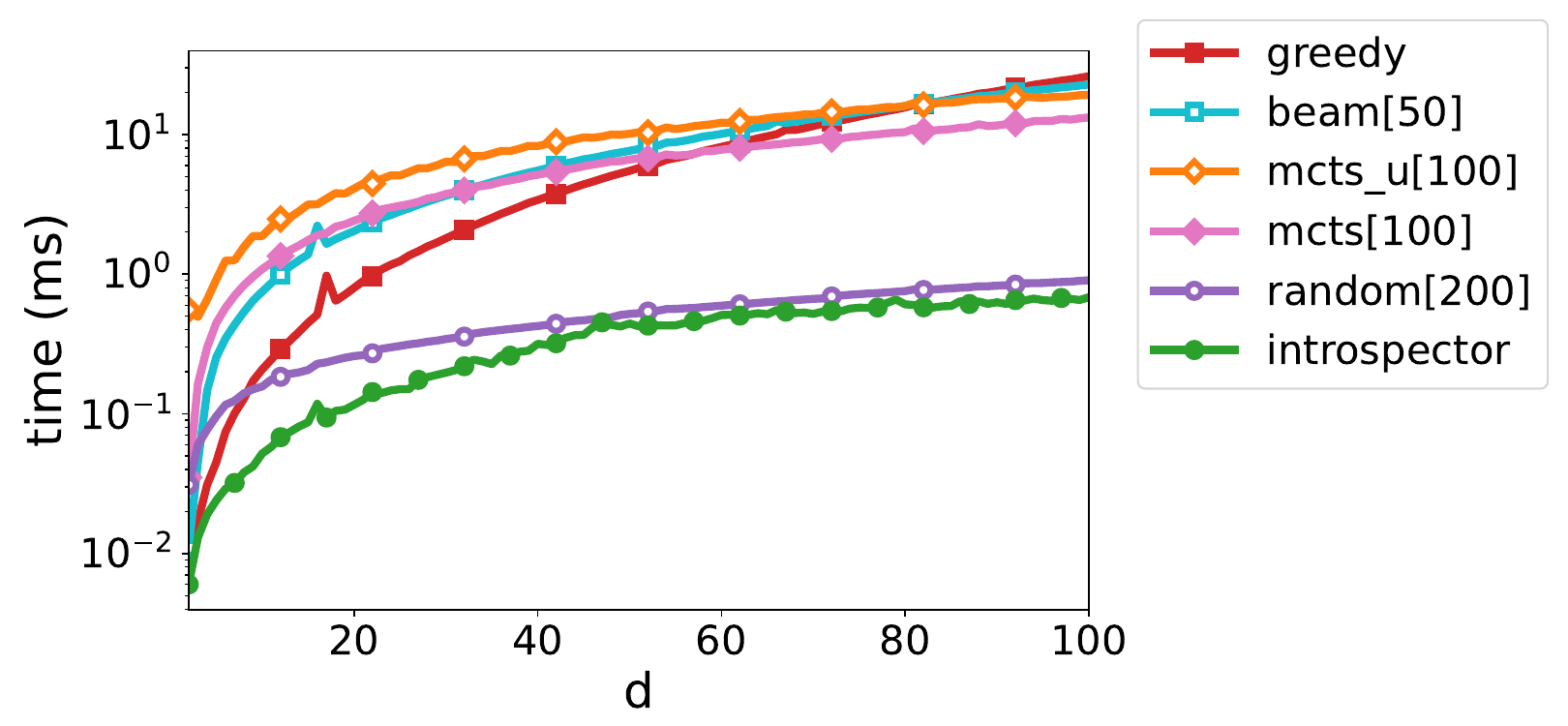}
         \caption{time (log-$y$)}
     \end{subfigure}
     
     \begin{subfigure}[b]{0.49\textwidth}
     \centering
         \includegraphics[width=\textwidth]{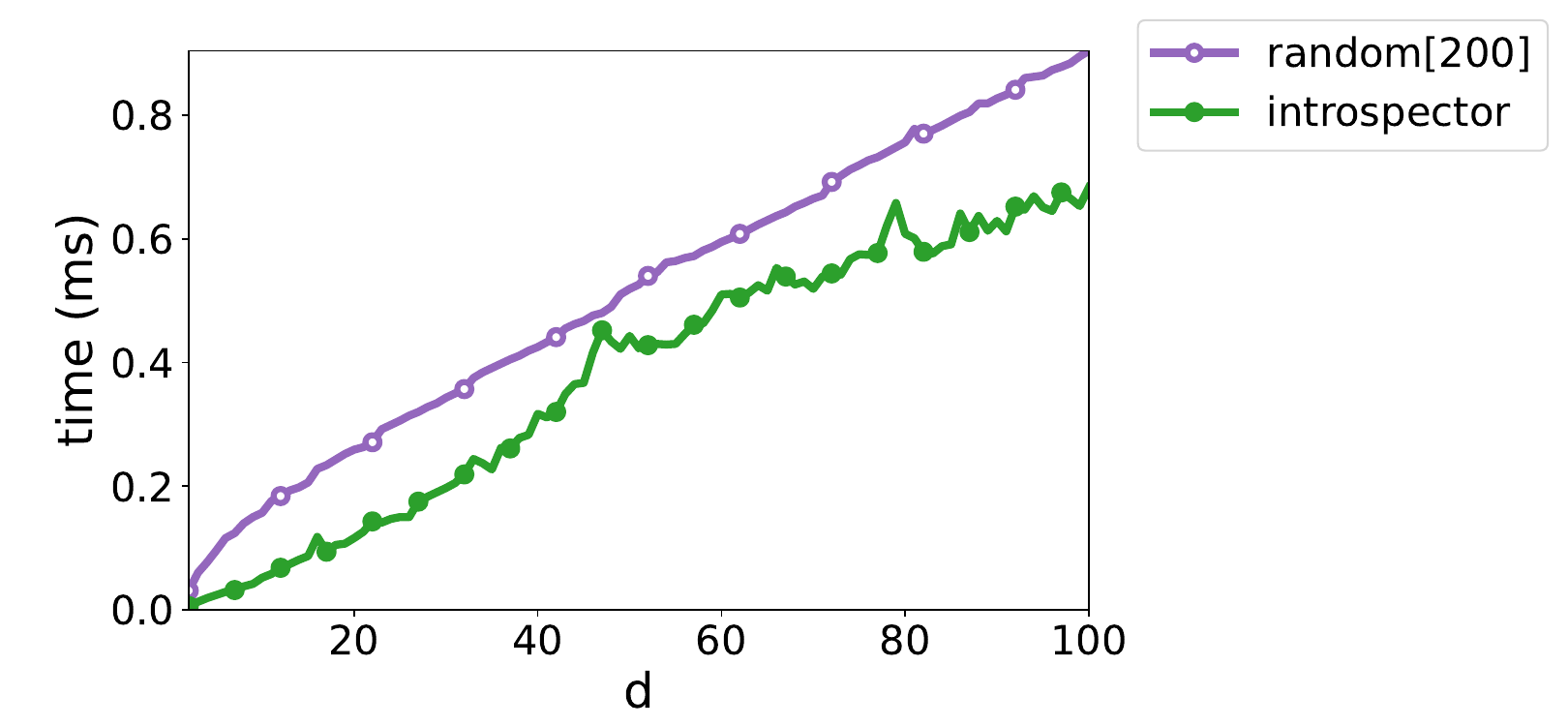}
         \caption{time (introspector and random)}
         \label{subfig:grid-time-2}
     \end{subfigure}
     \hfill
     \begin{subfigure}[b]{0.49\textwidth}
     \centering
         \includegraphics[width=\textwidth]{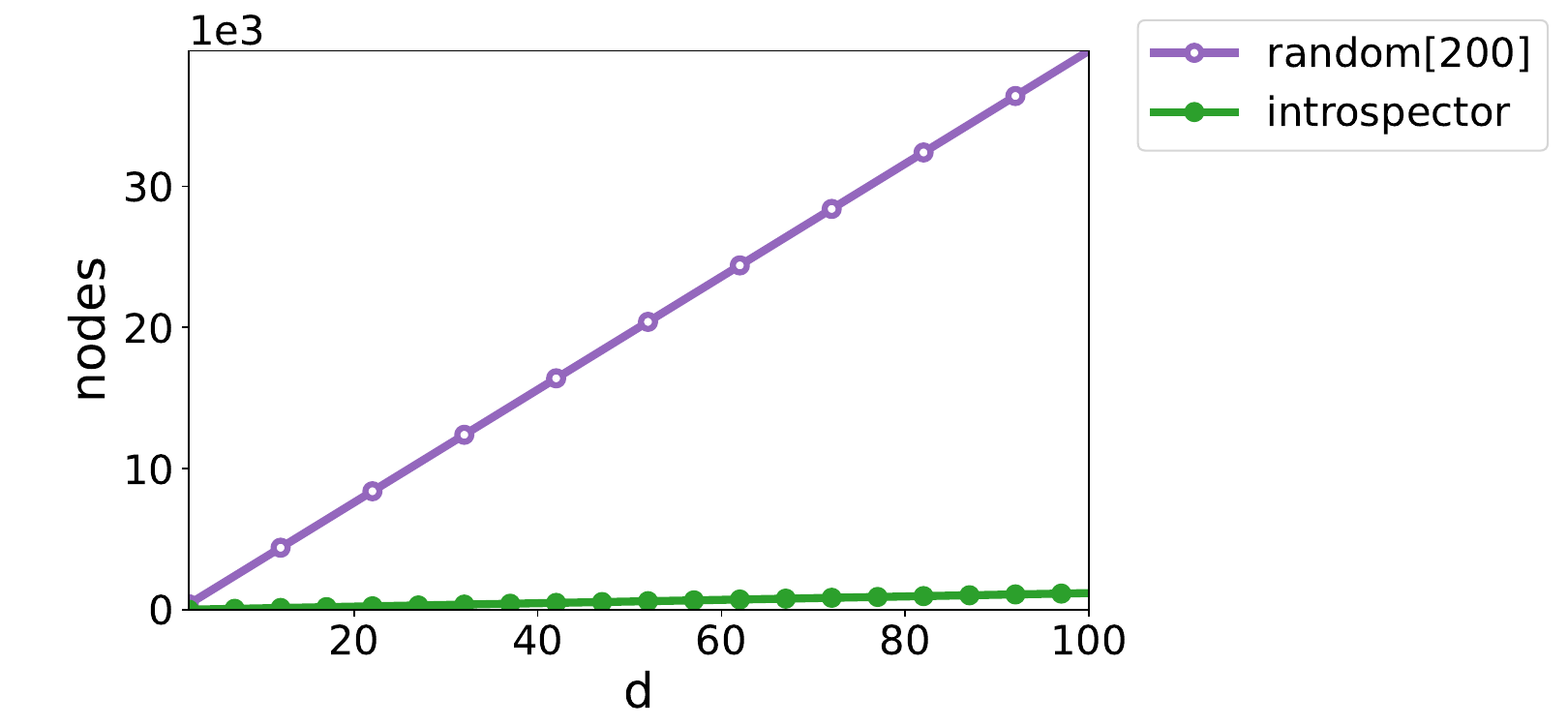}
         \caption{nodes expanded (introspector and random)}
         \label{subfig:grid-nodes-2}
     \end{subfigure}
     
\end{center}
\caption{Full planning results in the grid search domain}
\Description{Additional planning results in our grid search domain, highlighting the fact that our introspecting planner runs faster than a random sampling planner}
\label{fig:grid-full}
\end{figure}

Linear and semilog-$y$ plots from our grid search domain are shown in Figure~\ref{fig:grid-full}, enlarged relative to the original text for easier inspection. Figures~\ref{subfig:grid-time-2} and~\ref{subfig:grid-nodes-2} isolate our \emph{introspector} and the random planner to more clearly compare their runtime cost, showing that our method maintains efficiency as task difficulty (i.e., search space size) increases.

\clearpage
\section{Computational Growth of Baseline Planners in Grid Search Domain}\label{appendix:planner-scaling}

\begin{figure}[h!] 
\begin{center}

     \begin{subfigure}[b]{0.48\textwidth}
     \centering
         \includegraphics[width=\textwidth]{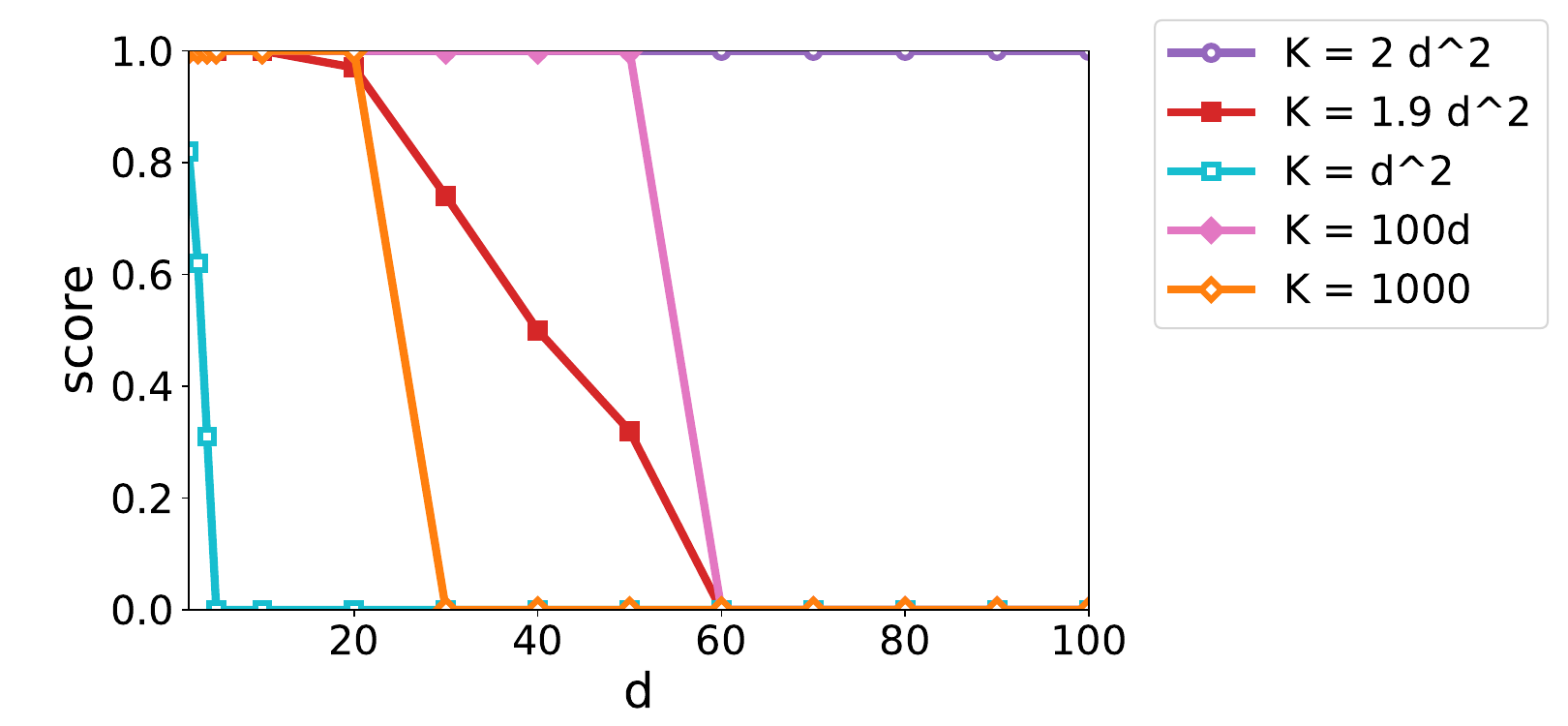}
         \caption{\code{BEAM[K]} score}
     \end{subfigure}
     \hfill
     \begin{subfigure}[b]{0.48\textwidth}
     \centering
         \includegraphics[width=\textwidth]{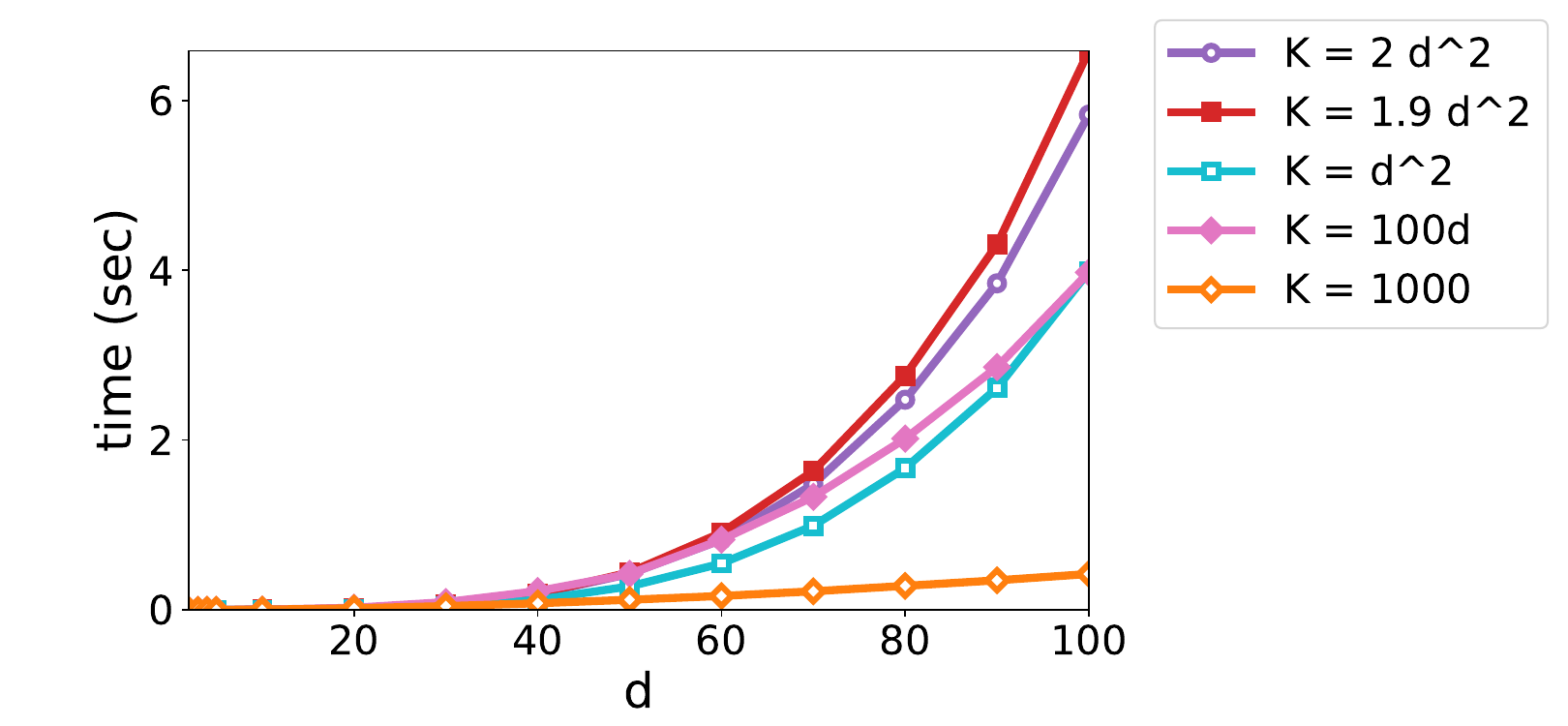}
         \caption{\code{BEAM[K]} time}
     \end{subfigure}

     \begin{subfigure}[b]{0.48\textwidth}
     \centering
         \includegraphics[width=\textwidth]{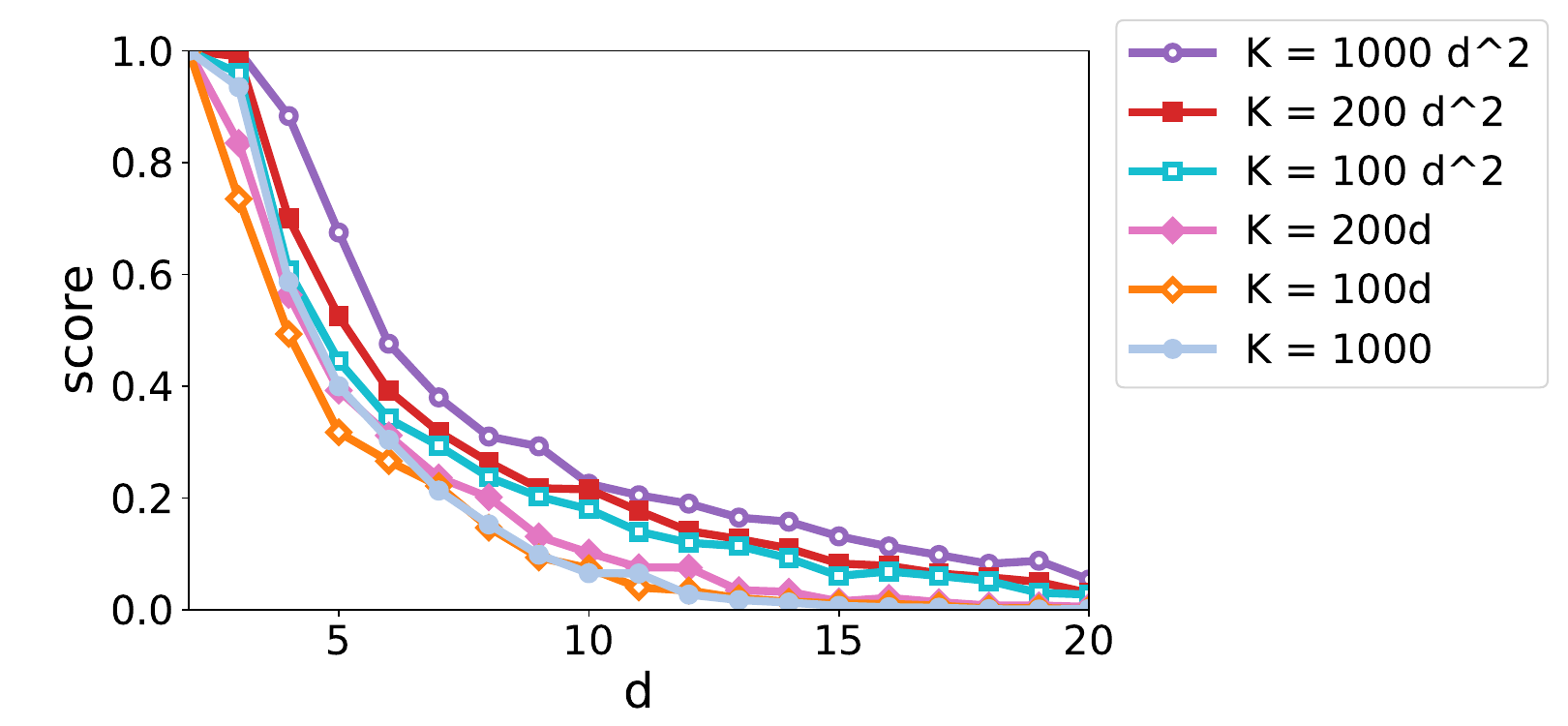}
         \caption{\code{MCTS[K]} score}
     \end{subfigure}
     \hfill
     \begin{subfigure}[b]{0.48\textwidth}
     \centering
         \includegraphics[width=\textwidth]{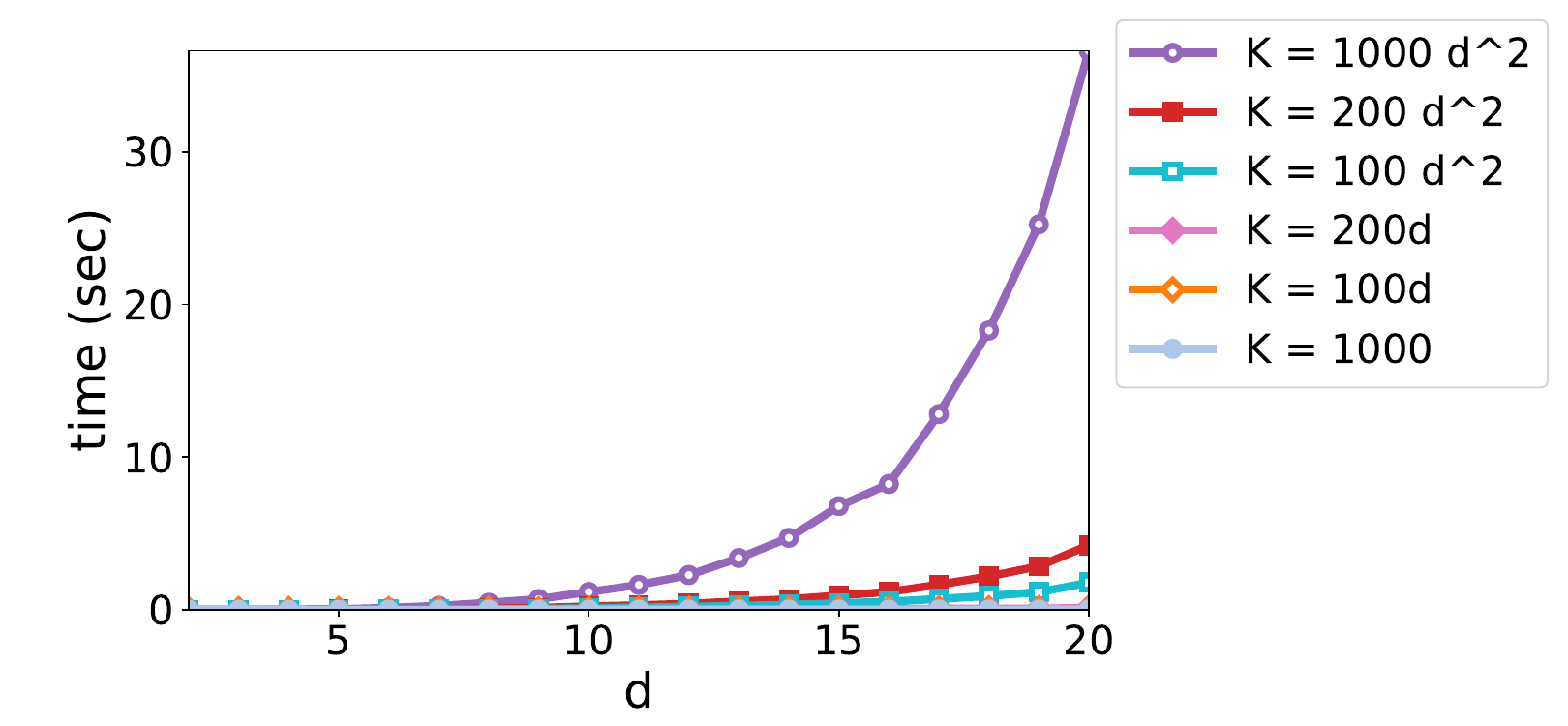}
         \caption{\code{MCTS[K]} time}
     \end{subfigure}

     \begin{subfigure}[b]{0.48\textwidth}
     \centering
         \includegraphics[width=\textwidth]{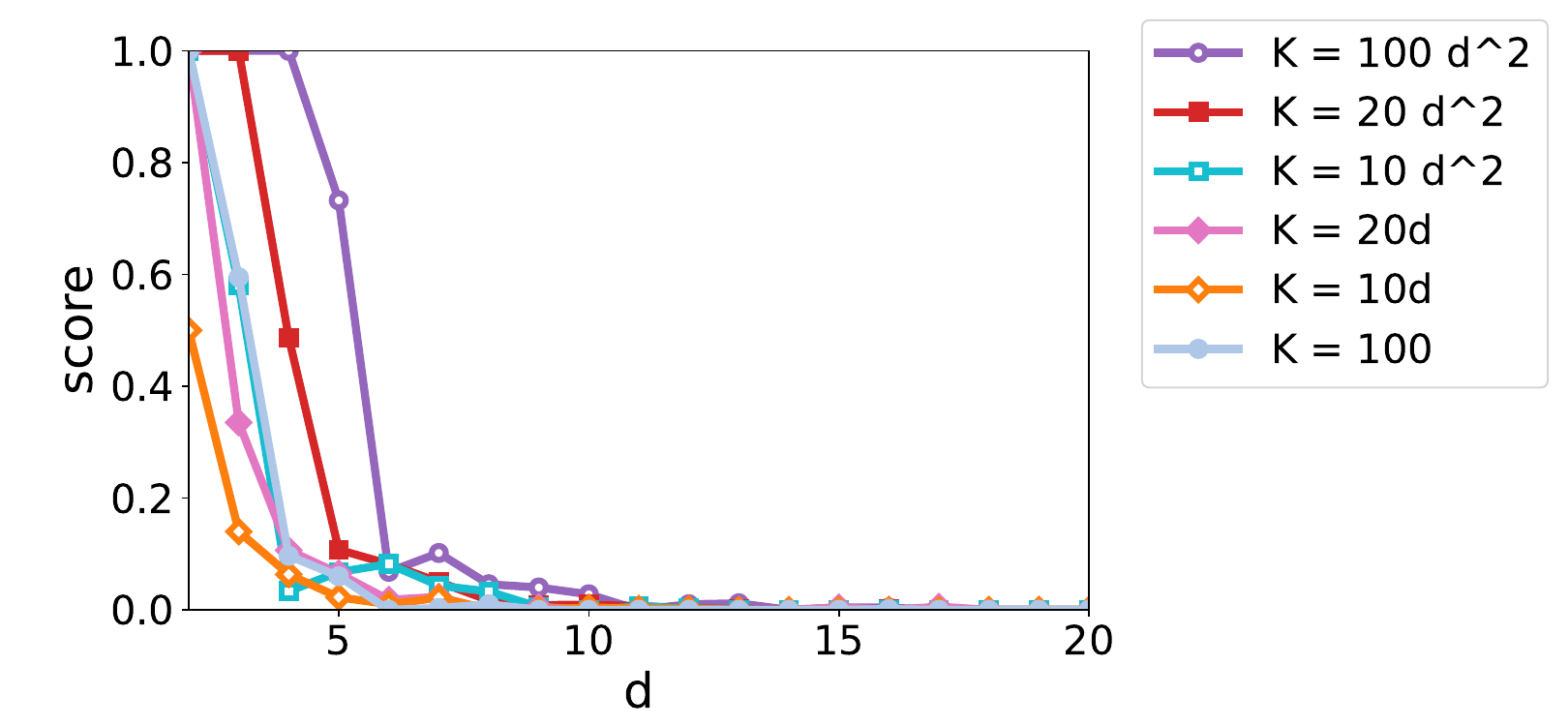}
         \caption{\code{MCTS-U[K]} score}
     \end{subfigure}
     \hfill
     \begin{subfigure}[b]{0.48\textwidth}
     \centering
         \includegraphics[width=\textwidth]{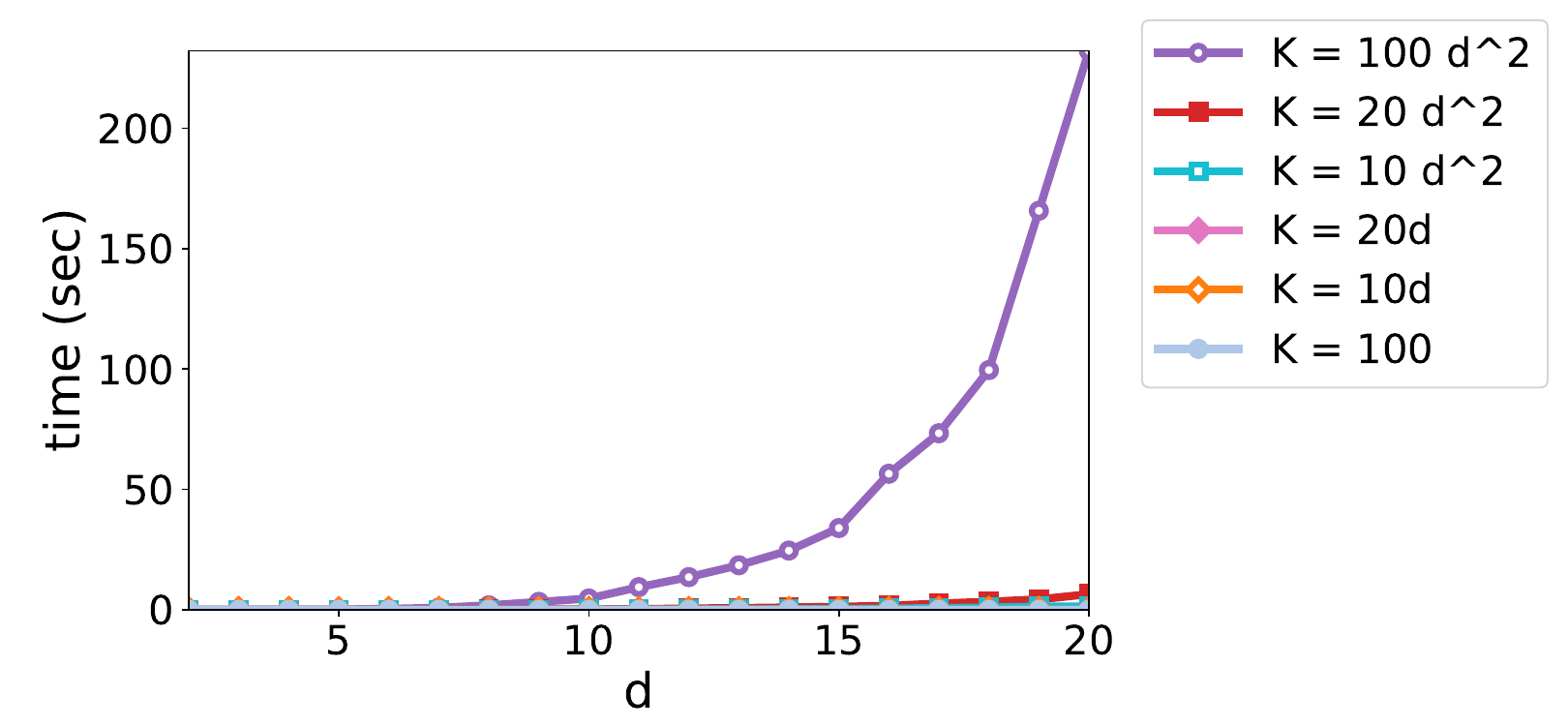}
         \caption{\code{MCTS-U[K]} time}
     \end{subfigure}
     
\end{center}
\caption{Planner scaling results in the grid search domain}
\Description{Results showing the scaling properties of baseline planners in the grid search domain (as their computation budgets increase). While planning time rises significantly, performance still drops off rapidly, except for beam search with $K = 2d^2$.}
\label{fig:grid-scale}
\end{figure}

It is well-known that no uninformed search method
can beat breadth-first search \emph{in general}.
Unfortunately, this fact applies to essentially all existing methods for planning in reinforcement learning: in the worst case, they regress to some version of uninformed search. However, given the recent successes in applying techniques such as MCTS \cite{schrittwieser20}, one may suspect that these algorithms can, through some cleverness, perform search more efficiently. While this may be the case in environments with plentiful rewards, or when learned approximations $\hat{Q}$ and $\hat{P}$ are available, we show in Figure~\ref{fig:grid-scale} that it is \emph{not} the case in our grid search example domain. Specifically, we observe:
\begin{itemize}
\item \code{BEAM[K]} deterministically achieves optimal scores only with a computational budget of $K=2 d^2$; any smaller budget eventually suffers from a sudden drop in performance at some $d$ value. Since the value of $K$ determines beam search's width, the planner cannot ensure that the reward will be found unless the budget is sufficient to retain \emph{all} nodes found up to the $d^{th}$ step.
\item \code{MCTS[K]} displays a gradual drop-off in score as $d$ increases, \emph{regardless of the computation budget}, since it degenerates to a simple random-sampling search in this environment. The additional budget merely increases the odds that it finds a reward.
\item \code{MCTS-U[K]} spends a great deal of computation time attempting to compute a good path based on the reward signals it has encountered. However, since rewards in this domain are extremely sparse, this time is essentially wasted. Thus, like \code{MCTS}, this planner degenerates to a random sampler in this environment. Notably, the performance of \code{MCTS-U} is even worse than \code{MCTS} given the same amount of computation time due to the effort spent by this algorithm on clever sampling.
\end{itemize}
On the other hand, as highlighted in Appendix~\ref{appendix:grid-search}, our \emph{introspector} (which does not require a computation budget hyperparameter) demonstrates roughly linear runtime growth in this domain while achieving optimal scores.

\clearpage
\section{Relational State Mutation Pseudocode}\label{appendix:pseudocode}

Algorithm~\ref{alg:mutation} shows pseudocode for each specialization of the \code{mutateT} function. Algorithm~\ref{alg:mutation-helpers} includes definitions of three helper functions for processing sets of (sets of) Mutations.

\newcommand{\IV}{ \code{IMMUTABLY VALID} }

\begin{algorithm2e}[h!]
\SetInd{0.7em}{0.7em}
\SetKwProg{Fn}{Func}{}{}
\SetKw{Is}{is}
\SetKw{In}{in}
\SetKw{For}{for}

\BlankLine

\tcp{Mutation for a literal formula, $f = l$}
\Fn{mutateT(State $s$, LiteralFormula $f$) $\to$ set[Mutation]}{
	
	\BlankLine
	$l$ = $f$.literal
	
	\BlankLine
	\tcp{If $l$ is mutable, return a Mutation that ensures that $l$ true}
	\If{$l$ \Is mutable}{
		\Return{$\{$ Mutation($\{+l\}$) $\}$}
	}
	
	\BlankLine
	\tcp{If $l$ is not mutable, but is true, then this formula is always valid}
	\If{$l \in s$.facts}{
		\Return{$\{$\IV$\}$}
	}
	
	\BlankLine
	\tcp{Failure; this literal is not satisfiable}
	\Return{$\{\}$}
}

\BlankLine
\BlankLine
\BlankLine

\tcp{Mutation for a conjunctive formula, $f = f_1 \land f_2 \land ... \land f_n$}
\Fn{mutateT(State $s$, AndFormula $f$) $\to$ set[Mutation]}{

	\BlankLine
	
	set[set[Mutation]] child\_mutations = $\{$mutateT($s$, $f_i$) for $f_i$ in $f$.children$\}$
	
	\Return{satisfy\_each(child\_mutations)} \myc{Every way of satisfying at least one Mutation \emph{for each child formula}}
}

\BlankLine
\BlankLine
\BlankLine

\tcp{Mutation for a disjunctive formula, $f = f_1 \lor f_2 \lor ... \lor f_n$}
\Fn{mutateT(State $s$, OrFormula $f$) $\to$ set[Mutation]}{

	\BlankLine
	
	set[set[Mutation]] child\_mutations = $\{$mutateT($s$, $f_i$) for $f_i$ in $f$.children$\}$
	
	\Return{satisfy\_any(child\_mutations)} \myc{Every way of satisfying \emph{at least one child formula}}
}

\BlankLine
\BlankLine
\BlankLine

\tcp{Mutation for a negation formula, $f = \neg f'$}
\Fn{mutateT(State $s$, NotFormula $f$) $\to$ set[Mutation]}{

	\BlankLine
	
	$f'$ = $f$.child \myc{Negation formulas have just one child formula}

	\BlankLine

	\Return{mutateF($s$, $f'$)} \myc{Return all the Mutations that make the child formula $f'$ false}
	
}

\BlankLine
\BlankLine
\BlankLine

\tcp{Mutation for an existentially-quantified formula, $f = \exists X\colon f'(x)$}
\Fn{mutateT(State $s$, ExistentialFormula $f$) $\to$ set[Mutation]}{
	
	\BlankLine
	
	$f'$ = $f$.child \myc{Quantified formulas have just one child formula}
	
	$v$ = $f$.bound\_variable \myc{The variable bound by this quantifier; we will try substituting it with every constant in $s$}

	\BlankLine
	
	\tcp{Each way of satisfying the inner formula, bound using every choice of variable substitution for this state}
	set[set[Mutation]] resulting\_mutations = $\{$mutateT($s$, substitute[$f'$, $\{v \to c\}$]) \For $c$ \In $s$.constants$\}$
	
	\BlankLine
	
	\tcp{This quantifier is existential, so we only need it to be satisfied for a single binding}
	\Return{satisfy\_any(resulting\_mutations)}
	
}

\BlankLine
\BlankLine
\BlankLine

\tcp{Mutation for a universally-quantified formula, $f = \forall X\colon f'(x)$}
\Fn{mutateT(State $s$, UniversalFormula $f$) $\to$ set[Mutation]}{
	
	\BlankLine
	
	$f'$ = $f$.child \myc{Quantified formulas have just one child formula}
	
	$v$ = $f$.bound\_variable \myc{The variable bound by this quantifier; we will try substituting it with every constant in $s$}

	\BlankLine
	
	\tcp{Each way of satisfying the inner formula, bound using every choice of variable substitution for this state}
	set[set[Mutation]] resulting\_mutations = $\{$mutateT($s$, substitute[$f'$, $\{v \to c\}$]) \For $c$ \In $s$.constants$\}$
	
	\BlankLine
	
	\tcp{This quantifier is universal, so we need \emph{every} binding to be satisfied}
	\Return{satisfy\_each(resulting\_mutations)}
}

\BlankLine

\caption{State mutation pseudocode for each Formula type}
\label{alg:mutation}
\end{algorithm2e}

\begin{algorithm2e}[h!]
\SetInd{0.7em}{0.7em}
\SetKwProg{Fn}{Func}{}{}
\SetKw{Is}{is}
\SetKw{In}{in}
\SetKw{For}{for}

\BlankLine

\tcp{Create a new Mutation that includes (satisfies) every Mutation from a set}
\Fn{satisfy\_all(set[Mutation] mutations) $\to$ Mutation}{

	\BlankLine
	\If{all mutations are \IV}{
		\Return{\IV}
	}
	
	\BlankLine
	\tcp{Produce a result by flattening all the mutations into one}
	\tcp{I.e., by combining all the positive (and negative) literals they require}
	\tcp{(and skipping any \IV mutations in the set)}
	\BlankLine
	\Return{$\{l$ \For $m$ \In mutations, \For $l$ \In $m$ (if $m$ is not \IV)$\}$}
}

\BlankLine
\BlankLine
\BlankLine

\tcp{Create a new set of Mutations, each of which satisfies one Mutation from each of the input sets}
\tcp{This operation can be thought of as a Cartesian Product for Mutations}
\Fn{satisfy\_each(set[set[Mutation]] mutations) $\to$ set[Mutation]}{

	\BlankLine
	
	set[set[Mutation]] p = CartesianProduct(mutations) \myc{Every way of picking one Mutation from each set}
	
	\BlankLine
	
	\Return{$\{$satisfy\_all(ms) \For ms \In p$\}$}
}

\BlankLine
\BlankLine
\BlankLine

\tcp{Create a new set of Mutations consisting of each Mutation in each of the input sets}
\tcp{This operation can be thought of as ``flattening'' the input sets into a single set}
\Fn{satisfy\_any(set[set[Mutation]] mutations) $\to$ set[Mutation]}{
	
	\BlankLine
	ms = $\{m$ \For $s$ \In mutations, \For $m$ \In $s\}$ \myc{Flatten the sets}
	
	\BlankLine
	\If{\IV  \In ms}{
		\Return $\{$\IV$\}$ \myc{One of the Mutations is \IV, so we can ignore the others}
	}
	
	\BlankLine
	\Return{ms}
}

\BlankLine

\caption{Mutation helper functions}
\label{alg:mutation-helpers}
\end{algorithm2e}

\clearpage
\section{Relational State Mutation Example}\label{appendix:rrl-example}

The definition of our \code{bins} environment is given in Algorithm~\ref{alg:bins-env}. In this domain, episodes begin with items randomly distributed in several bins; the agent can take an item out of a bin, put an item in a bin, or close a bin. Every time an empty bin is closed, the agent gets a reward. The episode terminates successfully once all bins are closed and empty. Note that bins can be closed even when they are not empty, which prevents the episode from being completed successfully. In addition, putting items in bins is a "distractor" action that is never useful; it makes the search problem more difficult by increasing the size and connectedness of the reachable state space.

\begin{algorithm2e}[h!]
\SetInd{0.7em}{0.7em}

\SetKwProg{Fn}{Func}{}{}


\SetKwProg{Action}{Action}{}{}
\SetKw{Predicates}{predicates:}
\SetKwInOut{Pre}{pre}
\SetKwInOut{Add}{add}
\SetKwInOut{Del}{del}

\tcp{The notation below for predicate declaration gives name/arity}
\Predicates{IsItem/1, IsBin/1, OnShelf/1, InBin/2, Open/1}

\BlankLine
\BlankLine
\BlankLine

\tcp{Close a bin}
\Action{CloseBin(X)}{

	\Pre{IsBin(X) $\land$ Open(X)}
	
	\Add{}
	
	\Del{Open(X)}
	
}

\BlankLine

\tcp{Take an item from a bin and place it on the shelf}
\Action{Pick(X, Y)}{

	\Pre{IsItem(X) $\land$ IsBin(Y) $\land$ Open(Y) $\land$ InBin(X, Y)}
	
	\Add{OnShelf(X)}
	
	\Del{InBin(X, Y)}
	
}

\BlankLine

\tcp{Take an item from the shelf and place it in a bin}
\Action{Put(X, Y)}{

	\Pre{IsItem(X) $\land$ IsBin(Y) $\land$ Open(Y) $\land$ OnShelf(X)}
	
	\Add{InBin(X, Y)}
	
	\Del{OnShelf(X)}
	
}

\BlankLine
\BlankLine
\BlankLine

\tcp{The agent gets a reward every time it closes a bin that is empty}
\tcp{Note that unlike in the Blocks-World domain, predicates are evaluated over $s$ (rather than $s'$) for the reward decision list in the bins environment}
\tcp{(handling both cases is a straightforward implementation detail)}
\Fn{R($s$, $a$, $s'$)}{
	\BlankLine
	\If{$\exists X \in s\text{.constants}\colon IsBin(X) \land a = CloseBin(X) \land \neg [\exists Y \in s\text{.constants}\colon: IsItem(Y) \land InBin(Y, X)]$}{
		\Return{$+1$}
	}
	\BlankLine
	\Return{$0$}
}

\BlankLine
\BlankLine
\BlankLine

\tcp{The episode terminates successfully once all bins are closed and empty}
\tcp{(predicates are evaluated over $s'$ for the termination decision list)}
\Fn{C($s$, $a$, $s'$)}{
	\BlankLine
	\If{$\forall X \in s'\text{.constants}\colon IsBin(X) \to [\neg Open(X) \land \neg \exists Y \in s'\text{.constants}\colon [IsItem(Y) \land InBin(Y, X)]]$}{
		\Return{\code{SUCCESS}} \myc{The episode is over}
	}
	\BlankLine
	\Return{\code{CONTINUE}} \myc{The agent continues to take actions}
}

\BlankLine

\caption{Definition of the \code{bins} domain for relational reinforcement learning}
\label{alg:bins-env}
\end{algorithm2e}
 
\begin{figure}[h!] 
\begin{center}

     \begin{subfigure}[b]{\textwidth}
     \centering
         \includegraphics[width=\textwidth]{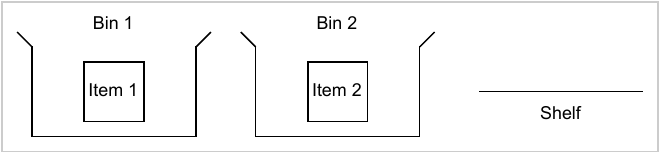}
         
         \caption{state 1 (initial state)}
         \label{subfig:bins-example-state-1}
     \end{subfigure}

     \begin{subfigure}[b]{0.49\textwidth}
     \centering
         \includegraphics[width=\textwidth]{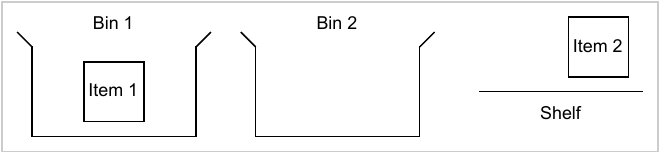}
         
         \caption{state 5}
         \label{subfig:bins-example-state-5}
     \end{subfigure}
     \hfill
     \begin{subfigure}[b]{0.49\textwidth}
     \centering
         \includegraphics[width=\textwidth]{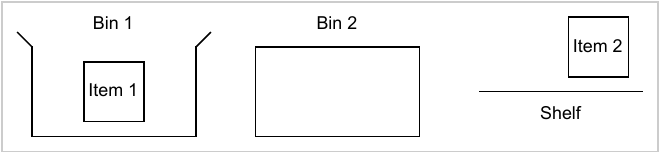}
         
         \caption{state 12 (first milestone)}
         \label{subfig:bins-example-state-12}
     \end{subfigure}

     \begin{subfigure}[b]{0.49\textwidth}
     \centering
         \includegraphics[width=\textwidth]{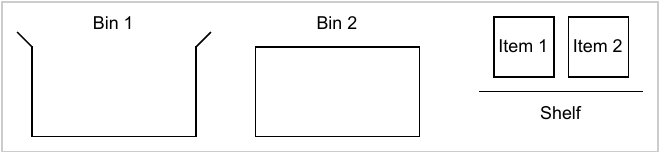}
         
         \caption{state 18}
         \label{subfig:bins-example-state-18}
     \end{subfigure}
     \hfill
     \begin{subfigure}[b]{0.49\textwidth}
     \centering
         \includegraphics[width=\textwidth]{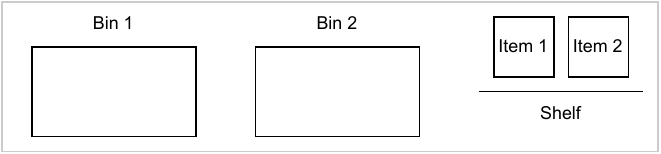}
         
         \caption{state 24 (successful terminal state)}
         \label{subfig:bins-example-state-24}
     \end{subfigure}
     
\end{center}
\caption{Example \code{bins} domain states, corresponding to the state-space diagram in Figure~\ref{fig:bins-state-space}.}
\Description{Example state visualizations for the bins domain.}
\label{fig:bins-example-states}
\end{figure}

\begin{figure}[h!] 
\begin{center}

\includegraphics[width=\textwidth]{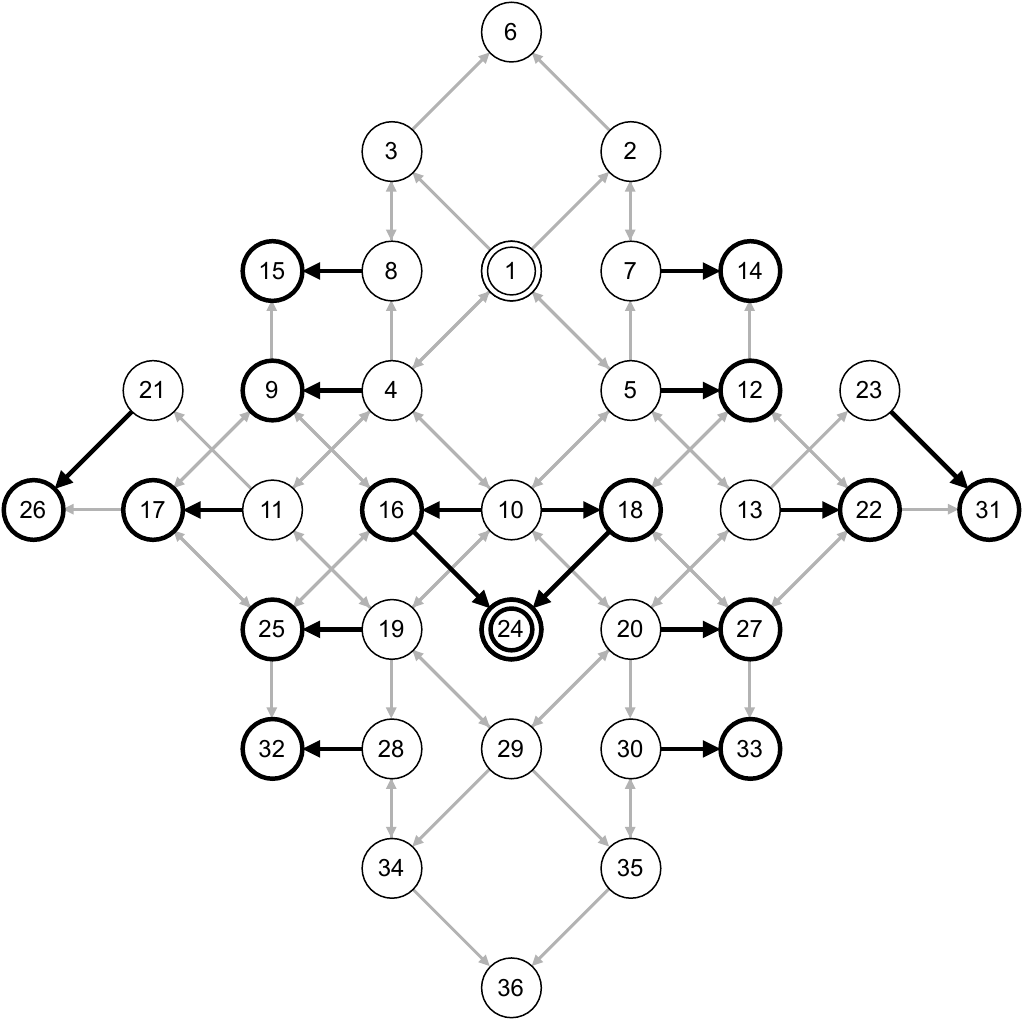}
     
\end{center}
\caption{Reachable state space of the bins example from Figure~\ref{fig:bins-example-states}. Milestone states are bolded, the initial state is labeled 1, and reaching state 24 leads to successful termination of the episode. Bolded edges indicate a transition yielding a reward, while gray edges denote a transition with no reward. States are numbered based on BFS visitation order.}
\Description{Reachable state space of the bins example, with milestones highlighted}
\label{fig:bins-state-space}
\end{figure}

An example initial state is shown in Figure~\ref{subfig:bins-example-state-1}, with two items and two bins. The reachable state space for this example state is shown in Figure~\ref{fig:bins-state-space}, which highlights the various milestone states. In total, there are thirty-six (36) reachable states in this example. In order to solve the episode successfully, the agent must find a plan that leads to state number 24. This involves getting exactly two rewards, though there are many distinct action sequences that lead to success. There are also many dead-end states from which the agent cannot recover.\footnote{The success state (24) is not reachable from states 2, 3, 6, 7, 8, 14, 15, 21, 23, 26, 28, 30, 31, 32, 33, 34, 35, and 36 (18 out of the 36 total states).}

The smallest number of actions to get from the initial state (1) to the terminal state (24) is four. This is the length of the plan found by \emph{introspector}, which takes the actions \code{Pick(i2, d2)} $\to$ \code{CloseBin(d2)} $\to$ \code{Pick(i1, d1)} $\to$ \code{CloseBin(d1)} resulting in the state trajectory $1 \to 5 \to 12 \to 18 \to 24$, all of which are shown in Figure~\ref{fig:bins-example-states}.
A sequence of actions like this can be found efficiently using the domain-agnostic predicate-counting heuristic, even in instances with many items and bins, where the search space grows rapidly.
In this environment, the heuristic provides a strong signal that guides the planner along trajectories that remove items from one bin at a time before closing the bin to reach a milestone. Without the guidance of milestones and heuristics, planning becomes intractable. 

In the initial state, with items in both bins, the \code{mutateT} function computes two possible mutations:
\begin{itemize}
\item Empty bin 1: $\{\neg InBin(I_1, B_1), \neg InBin(I_2, B_1), CloseBin(B_1)\}$
\item Empty bin 2: $\{\neg InBin(I_1, B_2), \neg InBin(I_2, B_2), CloseBin(B_2)\}$
\end{itemize}
Where an action (e.g., $CloseBin(B_1)$) in a mutation indicates that the specified action should be taken once the other conditions are met in order to cause a rewarding transition. Our planning algorithm runs the inner-level search procedure to find one trajectory per mutation, leading to the state-space path $1 \to 4 \to 9$ for the first mutation (where 9 is a milestone) and $1 \to 5 \to 12$ (where 12 is a milestone) for the second.
Because of the particular tie-breaking method in our implementation, milestone-space search continues from the second sub-plan. The second segment begins at state 12, with bin 2 empty and closed; thus, the \code{mutateT} function now returns only one mutation, corresponding to emptying bin 1. To satisfy this mutation from state 12, the inner state-space search find a path to state 24, which leads to successful termination of the episode.

As a final note on this example, consider the relationship between state mutations and milestones. In this example, each mutation has many corresponding milestones; the first mutation (emptying bin 1) is satisfied by the transitions
\begin{center}
$4 \to 9$, $8 \to 15$, $10 \to 16$, $11 \to 17$, $18 \to 24$, $19 \to 25$, $21 \to 26$, and $28 \to 32$,
\end{center}
while the second (emptying bin 2) is satisfied by transitions
\begin{center}
$5 \to 12$, $7 \to 14$, $10 \to 18$, $13 \to 22$, $16 \to 24$, $20 \to 27$, $23 \to 31$, and $30 \to 33$.
\end{center}
Put simply, there is a one-to-many relationship between mutations and milestones in this domain: each mutation can be satisfied by one of several transitions.
Many of these can be incorporated into a valid plan, though some (e.g., $8 \to 15$) will prevent the agent from completing the task if pursued.
Although one may suspect that the large number of potential milestone states would reduce the effectiveness of milestone-based search, it turns out to pose very little problem for the planning algorithm.
There are at least three factors to consider here.

\textbf{First}, the number of reachable milestone states is strictly non-increasing as the search progresses through the state space. For example, although 16 milestone transitions are reachable from the initial state in Figure~\ref{fig:bins-state-space}, the \emph{only} reward that is available from state 12 is the one that is obtained by transitioning from state $18 \to 24$, successfully completing the task.
\textbf{Second}, it is typically easy for an informed search algorithm to avoid the paths that lead to irrelevant milestones. In the \code{bins} environment, alternate routes such as the one leading to the dead-end state 14 involve taking actions that do nothing to further the agent's goal of satisfying a given mutation; thus, the planner will strongly prefer to avoid these regions of the state space.
\textbf{Third}, milestones are a \emph{partial} abstraction in the sense that the agent still has access to the entire state space. If the outer-level search reaches a milestone that turns out to be suboptimal (or even a dead-end), the planner can back-track and try to satisfy the same mutation while explicitly avoiding any specific milestone states that have already been discovered.

In general, the agent's method of generating candidate milestones may over- or under-estimate the number of useful future states at any given point; this includes producing states that are undesirable or even completely unreachable. However, when integrated into a robust planning algorithm, the use of milestones can lead to a significant improvement in runtime.

\clearpage
\section{RRL Planning Algorithm}\label{appendix:rrl-planner}

The \emph{introspector} algorithm for relational reinforcement learning consists of two levels: an outer milestone-space search algorithm and an inner state-space search algorithm. Pseudocode for our implementation of the outer level, which uses greedy best-first search, is given in Algorithm~\ref{alg:planning-1}. Note that other techniques, such as exhaustive search, could be used here as well. In addition, although our implementation incorporates an upper limit on plan length $H$ and terminates as soon as a successful plan has been found, these are not necessary parts of the algorithm.
The key novelty to focus on compared to prior planning algorithms for reinforcement learning is the use of the domain-agnostic state-mutation procedure.

The inner state-space search, which finds a path from a given state to some state that satisfies a specified mutation (i.e., a milestone), can be performed by any informed search algorithm; we use greedy best-first search with the predicate-counting heuristic, leaving experiments with other techniques (e.g., A* with an admissible heuristic \cite{correa24}) to future work. The combination of this goal-directed search with reward-oriented milestones allows our bi-level planning algorithm to bypass a large fraction of the state-space when searching, leading to significant efficiency gains in environments with sparse rewards. The key to this is that the state mutations, and the milestones they correspond to, are specifically constructed such that reaching them will ensure the agent obtains a reward; thus, the agent wastes less time expanding state-space trajectories that lead to fruitless future states.

In Algorithm~\ref{alg:planning-1}, the extract\_maximal\_reward\_condition($\hat{R}$) function is a procedure that constructs a single formula for a decision list which, if satisfied, will ensure that the decision list outputs a specific one of its possible values. Let $L = [(f_1, r_1), (f_2, r_2), ..., (f_n, r_n)]$ be a decision list, where each $f_i$ is a FOL formula and each $r_i$ is a reward. The decision list outputs the reward value for the first satisfied formula. Thus, for some reward $r_i$ to be returned, we need $f_i$ to be satisfied and all preceding $f_1, f_2, ..., f_{i-1}$ to evaluate to false. This corresponds to the formula $f = \neg f_1 \land \neg f_2 \land ... \land \neg f_{i-1} \land f_i$. If we let $i = \argmax(r_i)$, then a state that satisfies the resulting formula $f$ is a milestone.

\begin{algorithm2e}[h!]
\SetInd{0.7em}{0.7em}
\SetKwProg{Fn}{Func}{}{}
\SetKw{Is}{is}
\SetKw{Isnt}{is not}
\SetKw{Not}{not}
\SetKw{In}{in}
\SetKw{Let}{let}
\SetKw{And}{and}
\SetKw{Or}{or}

\BlankLine

\tcp{Plan using models $\hat{T}$, $\hat{R}$, and $\hat{C}$}
\Fn{plan(State $s$, int $H$) $\to$ list[Action]}{
	
	\BlankLine
	
	\tcp{The reward model $\hat{R}$ is a decision list, from which we can easily extract a FOL formula that ensures maximal reward}
	condition = extract\_maximal\_reward\_condition($\hat{R}$)
	
	\BlankLine
	
	\Let Plan = tuple[int, int, list[Action], State] \myc{candidate plan = (score, moves remaining, actions, resulting state)}
	
	\BlankLine
	
	max\_priority\_queue[Plan] plans = $[]$ \myc{Greedy search in milestone space; candidate plans are ordered lexicographically, based on order of tuple elements}
	
	map[State, int] visited = $\{\}$ \myc{Keep track of the best achieved score at each milestone}
	
	\BlankLine
	
	plans.push($(0, H, [], s)$)
	
	visited[$s$] = $0$
	
	\BlankLine
	
	\While{plans.size() $> 0$}{
		
		\BlankLine
		
		Plan $p$ = plans.pop() \myc{Get the current best plan off the top of the stack}
		
		(score, $H'$, actions, state) = $p$ \myc{Unpack elements for convenience}
		
		\BlankLine
		
		\If{$H' = 0$}{
			\tcp{We ran out of moves without finding a successful plan;}
			\tcp{Let's just return the one with the best score so far}
			\tcp{(note that the plan length limit can be removed without substantially changing the behavior of our algorithm)}
			\Return{actions}
		}
		
		\BlankLine
		
		\tcp{Enumerate goals using state mutation procedure}
		
		\For{Mutation $m$ \In mutateT(state, condition)}{
			
			
			\BlankLine
			
			\tcp{Use an informed search algorithm to find a path from the current state to a new state that satisfies the mutation $m$ (within $H'$ steps)}
			(reward, new\_actions, new\_state, status) = search(state, $H'$, $m$)
			
			\BlankLine
			
			new\_score = score + reward \myc{Add the reward found along the returned trajectory}
			
			new\_actions = concatenate(actions, new\_actions)
			
			\BlankLine
			
			\If{status $=$ SUCCESS}{
			
				\BlankLine
				
				\tcp{A successful trajectory has been found; let's return now}
				
				\Return{new\_actions}
			}
			\ElseIf{status $=$ CONTINUE \And (new\_state \Not \In visited \Or visited[new\_state] < new\_score)}{
			
				\BlankLine
				
				\tcp{Continue searching}
				
				visited[new\_state] = new\_score
				
				plans.push($($new\_score, $H'-$ new\_actions.size(), new\_actions, new\_state$)$)
			}
		}
	}
	
	\BlankLine
	\tcp{The algorithm should never reach this line; return empty plan as failure}
	\Return{$[]$}
}

\BlankLine

\caption{Outer milestone-space search function}
\label{alg:planning-1}
\end{algorithm2e}

\clearpage
\section{RRL Planning Results}\label{appendix:rrl-results}

\begin{figure}[h!] 
\begin{center}

     \begin{subfigure}[t]{0.33\textwidth}
     \centering
         \includegraphics[width=\textwidth]{plots/plot_rrl_blocks-world-time}
         \caption{planner runtime (linear scale)}
     \end{subfigure}
     \hfill
     \begin{subfigure}[t]{0.33\textwidth}
     \centering
         \includegraphics[width=\textwidth]{plots/plot_rrl_blocks-world-time_logy}
         \caption{planner runtime (log-$y$)}
     \end{subfigure}
     \hfill
     \begin{subfigure}[t]{0.33\textwidth}
     \centering
         \includegraphics[width=\textwidth]{plots/plot_rrl_blocks-world-partial-time_logy}
         \caption{planner runtime, zoomed in (log-$y$)}
     \end{subfigure}
     \begin{subfigure}[t]{0.33\textwidth}
     \centering
         \includegraphics[width=\textwidth]{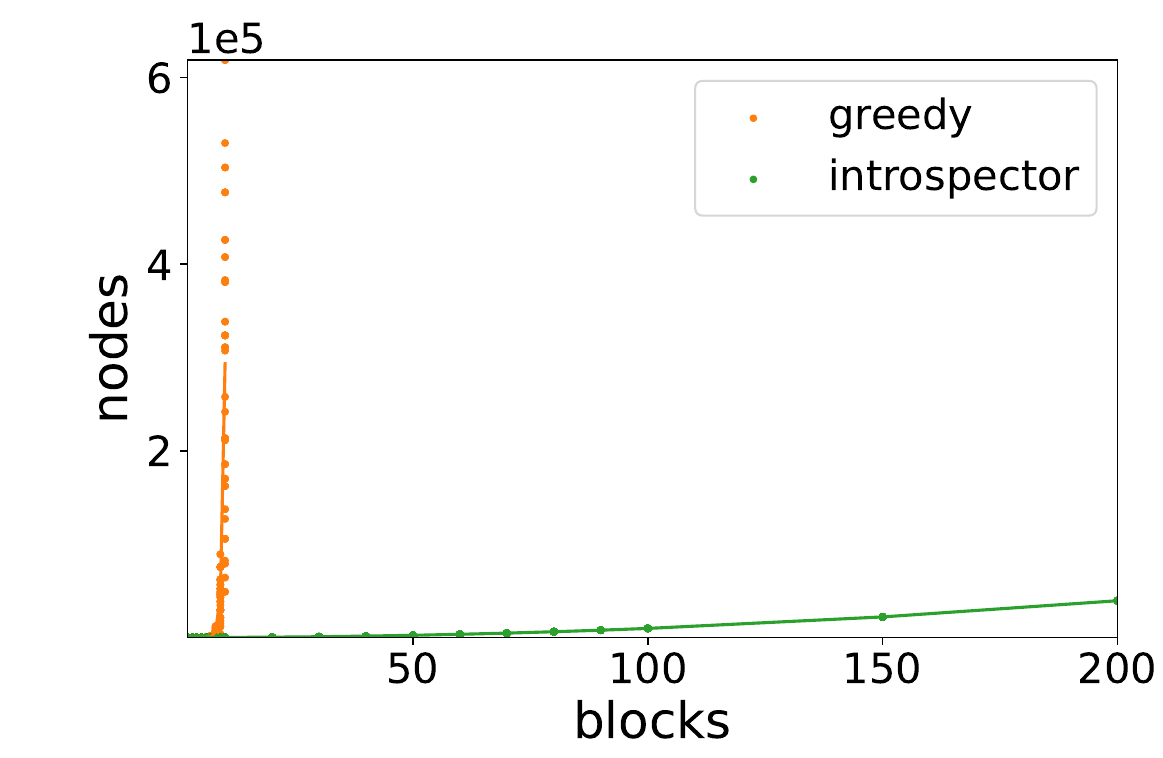}
         \caption{nodes expanded (linear scale)}
     \end{subfigure}
     \hfill
     \begin{subfigure}[t]{0.33\textwidth}
     \centering
         \includegraphics[width=\textwidth]{plots/plot_rrl_blocks-world-nodes_logy}
         \caption{nodes expanded (log-$y$)}
     \end{subfigure}
     \hfill
     \begin{subfigure}[t]{0.33\textwidth}
     \centering
         \includegraphics[width=\textwidth]{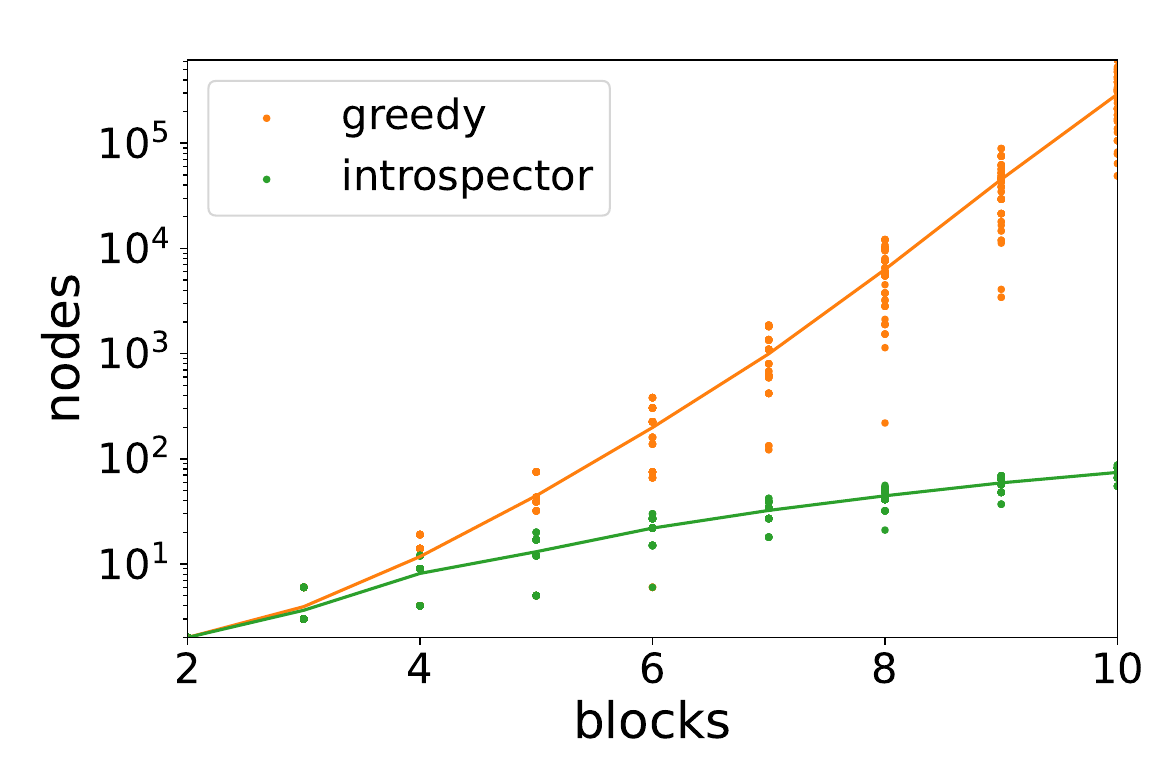}
         \caption{nodes expanded, zoomed in (log-$y$)}
     \end{subfigure}
     
\end{center}
\caption{Planning results in the Blocks-World domain}
\Description{Planning results in the Blocks-World relational domain, showing that our introspection-based planner performs significantly better than greedy search.}
\label{fig:rrl-complete-1}
\end{figure}

\begin{figure}[h!] 
\begin{center}

     \begin{subfigure}[t]{0.33\textwidth}
     \centering
         \includegraphics[width=\textwidth]{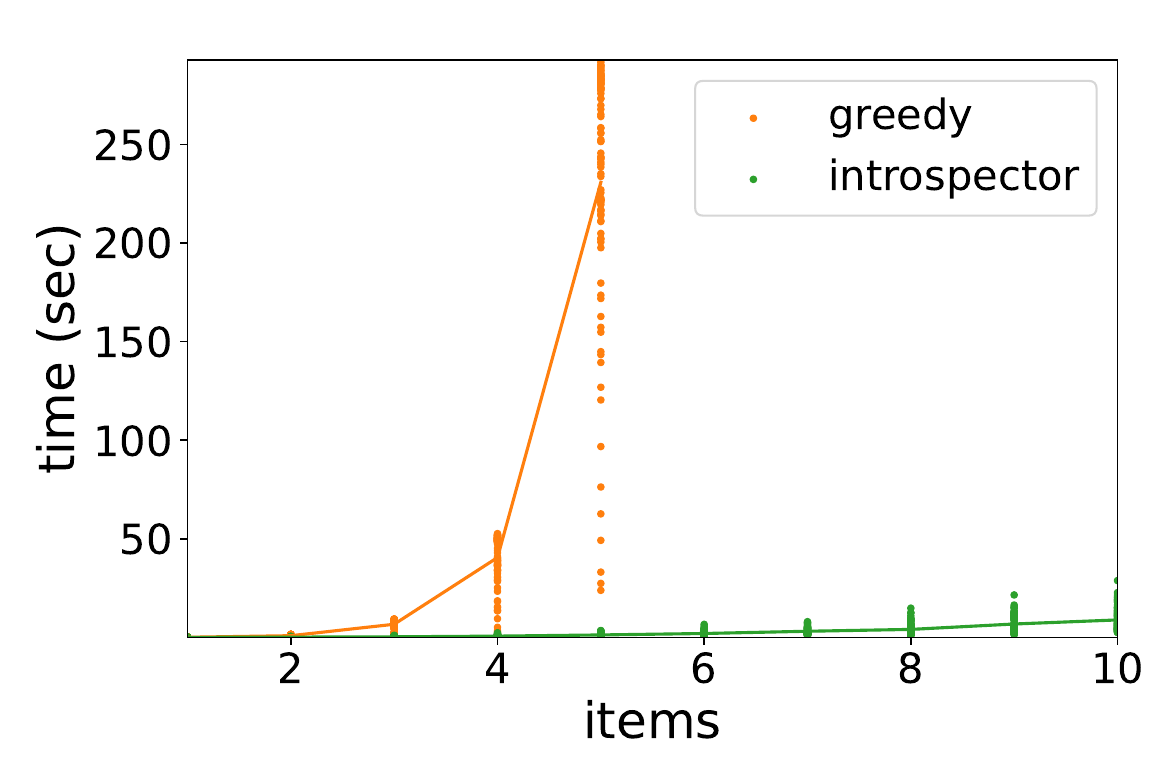}
         \caption{planner runtime (linear scale)}
     \end{subfigure}
     \hfill
     \begin{subfigure}[t]{0.33\textwidth}
     \centering
         \includegraphics[width=\textwidth]{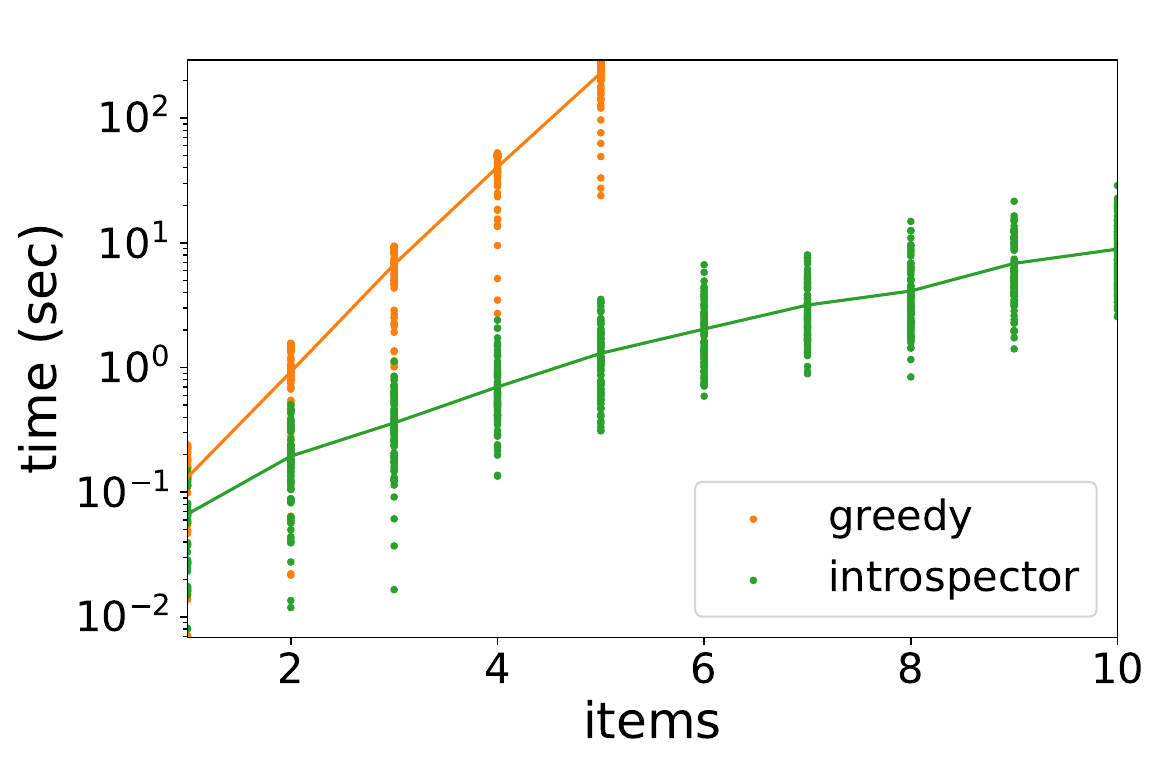}
         \caption{planner runtime (log-$y$)}
     \end{subfigure}
     \hfill
     \begin{subfigure}[t]{0.33\textwidth}
     \centering
         \includegraphics[width=\textwidth]{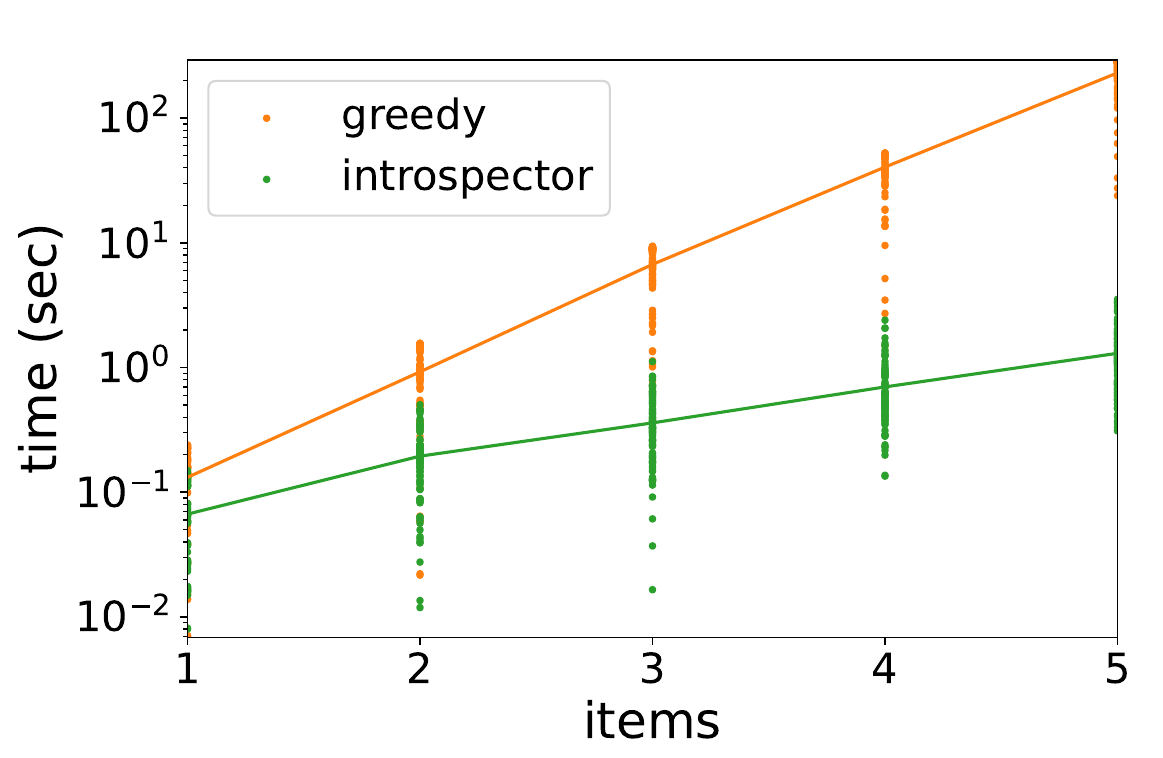}
         \caption{planner runtime, zoomed in (log-$y$)}
     \end{subfigure}
     \begin{subfigure}[t]{0.33\textwidth}
     \centering
         \includegraphics[width=\textwidth]{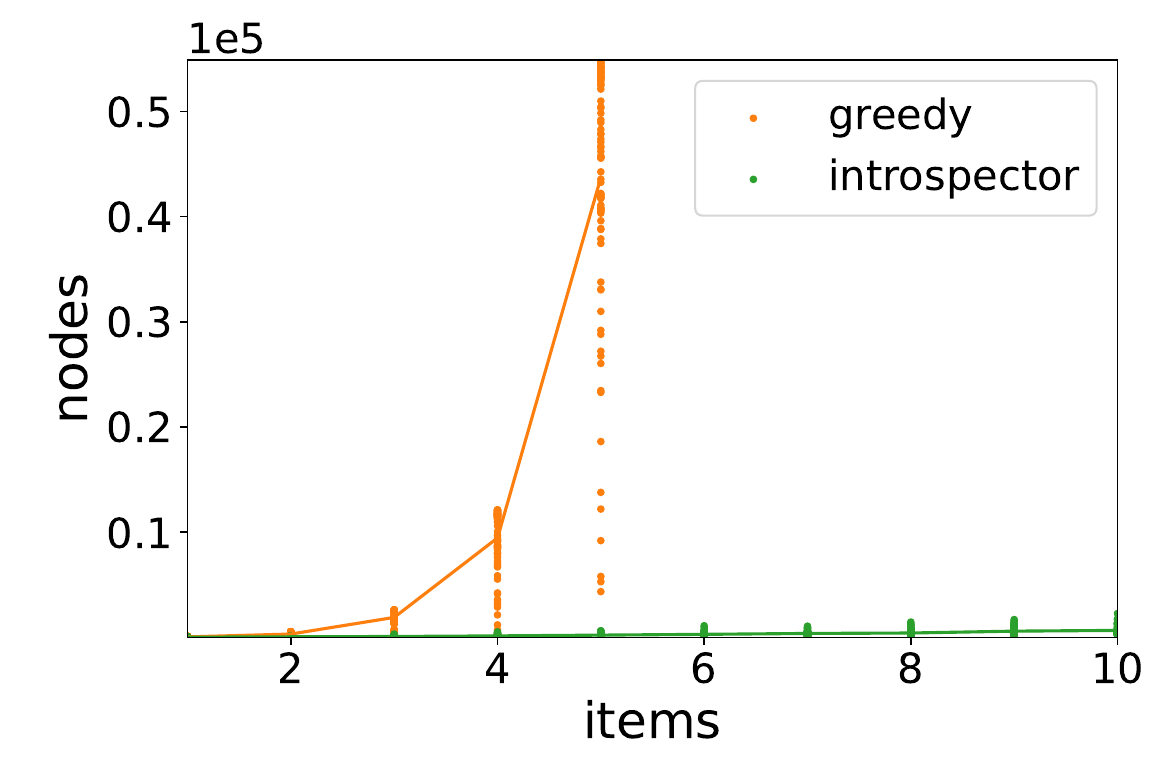}
         \caption{nodes expanded (linear scale)}
     \end{subfigure}
     \hfill
     \begin{subfigure}[t]{0.33\textwidth}
     \centering
         \includegraphics[width=\textwidth]{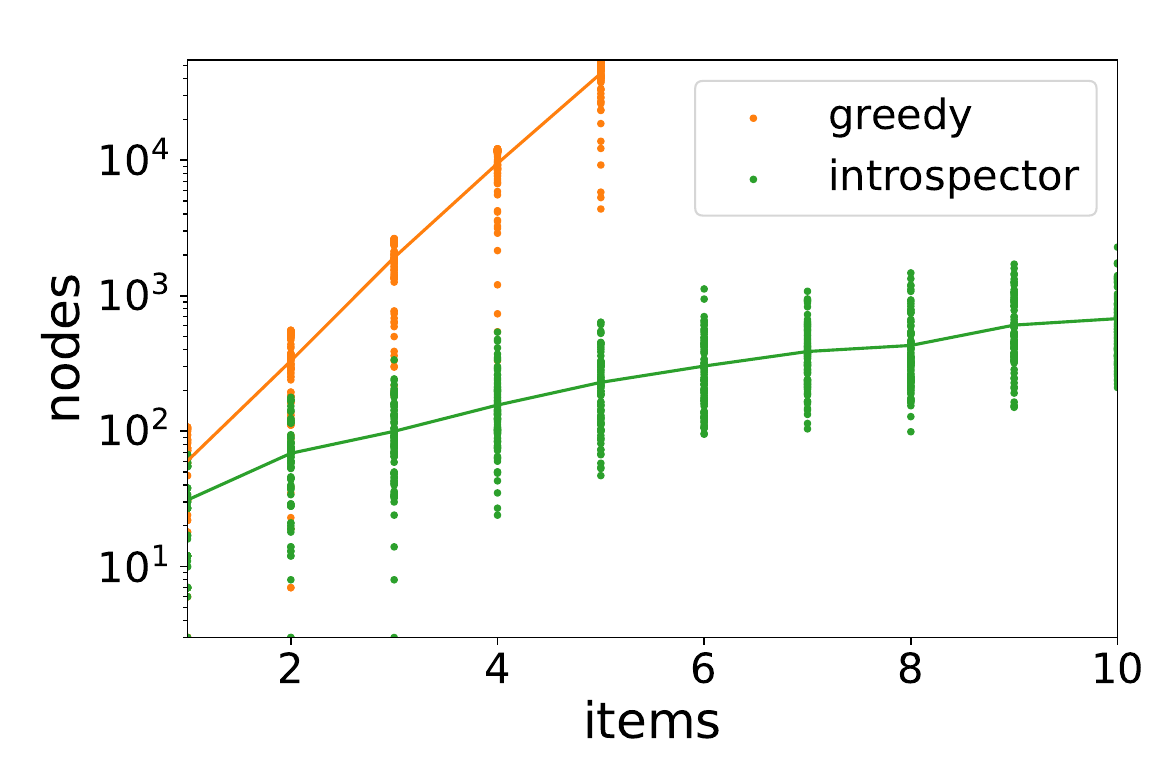}
         \caption{nodes expanded (log-$y$)}
     \end{subfigure}
     \hfill
     \begin{subfigure}[t]{0.33\textwidth}
     \centering
         \includegraphics[width=\textwidth]{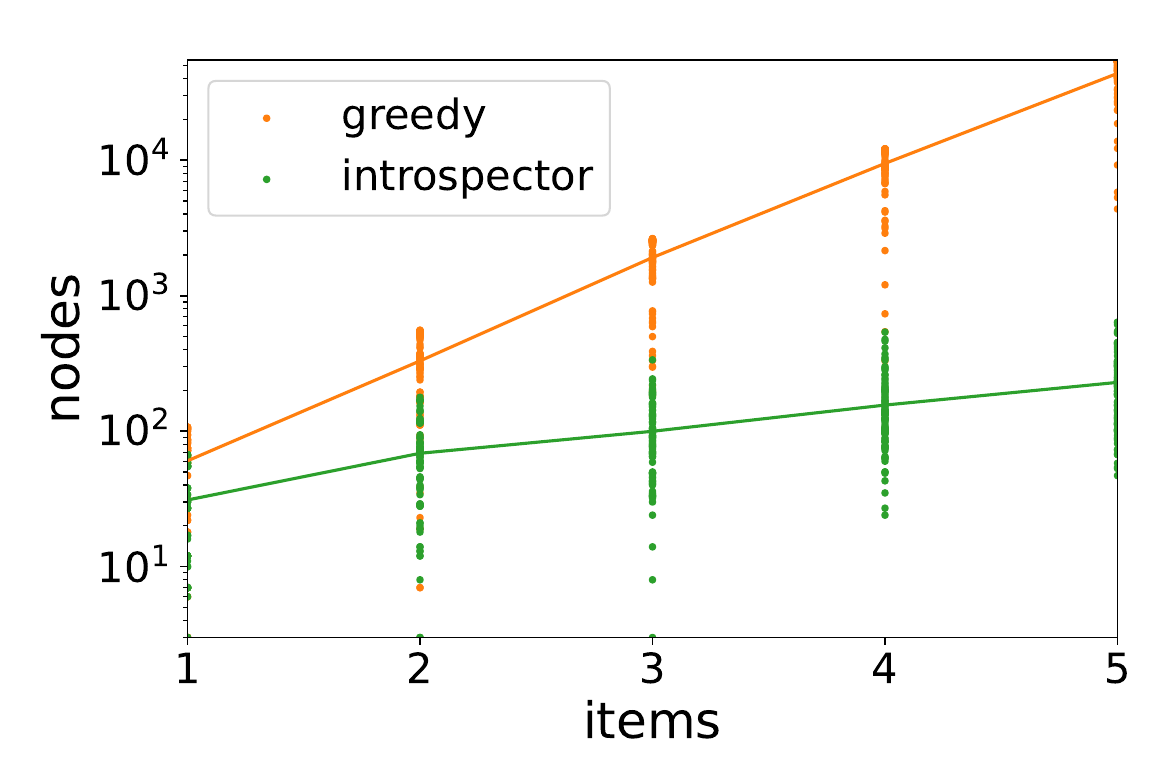}
         \caption{nodes expanded, zoomed in (log-$y$)}
     \end{subfigure}
     
\end{center}
\caption{Planning results in the drawers domain with $n = 3$}
\Description{Planning results in the drawers relational domain, showing that our introspection-based planner performs significantly better than greedy search.}
\label{fig:rrl-complete-2}
\end{figure}

\begin{figure}[h!] 
\begin{center}

     \begin{subfigure}[t]{0.33\textwidth}
     \centering
         \includegraphics[width=\textwidth]{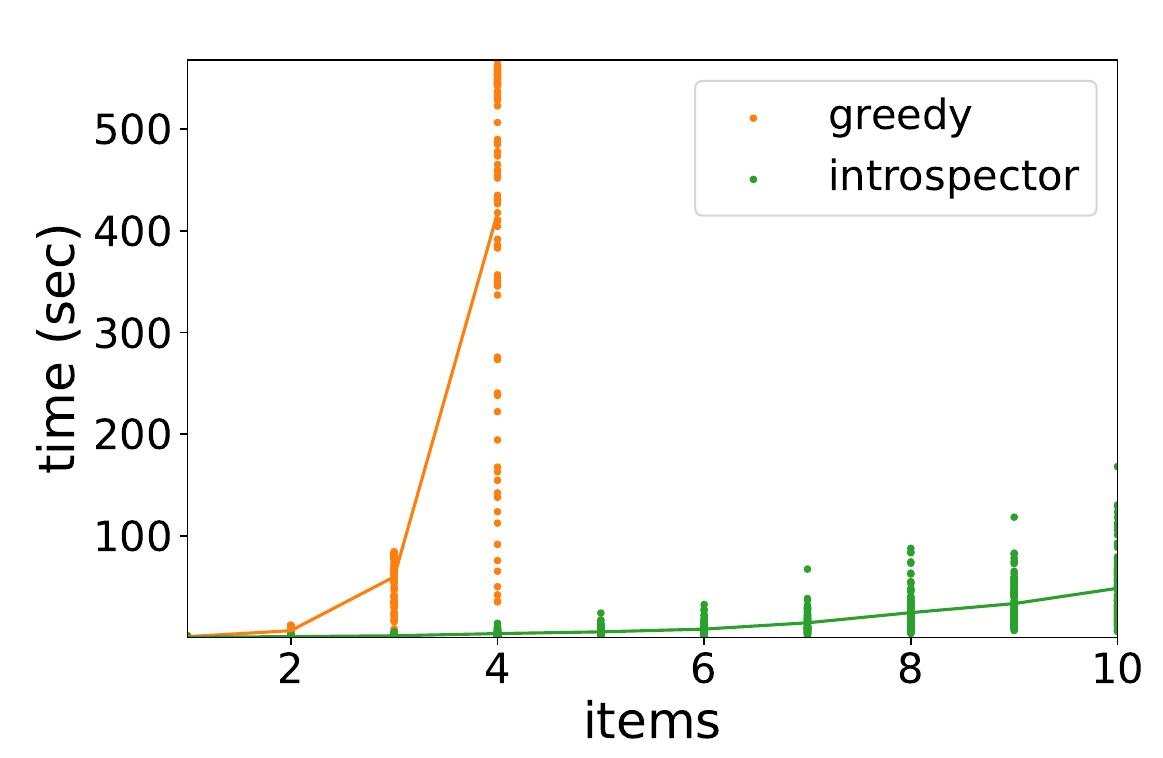}
         \caption{planner runtime (linear scale)}
     \end{subfigure}
     \hfill
     \begin{subfigure}[t]{0.33\textwidth}
     \centering
         \includegraphics[width=\textwidth]{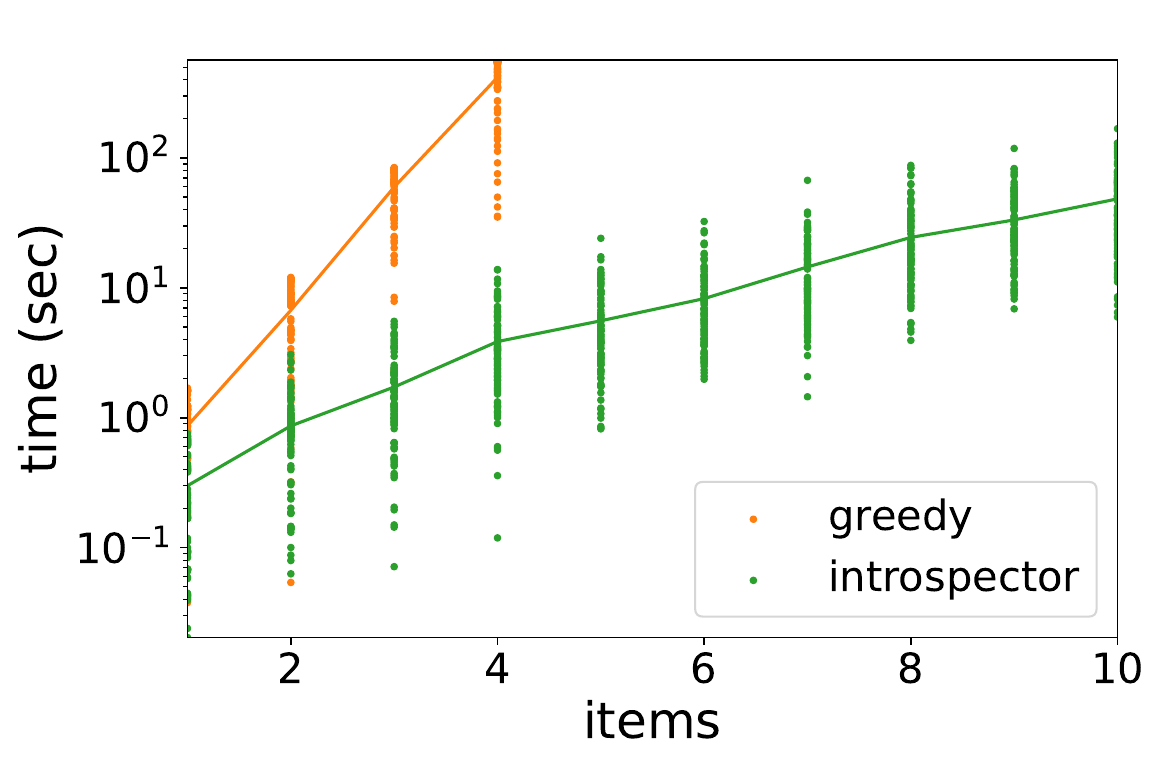}
         \caption{planner runtime (log-$y$)}
     \end{subfigure}
     \hfill
     \begin{subfigure}[t]{0.33\textwidth}
     \centering
         \includegraphics[width=\textwidth]{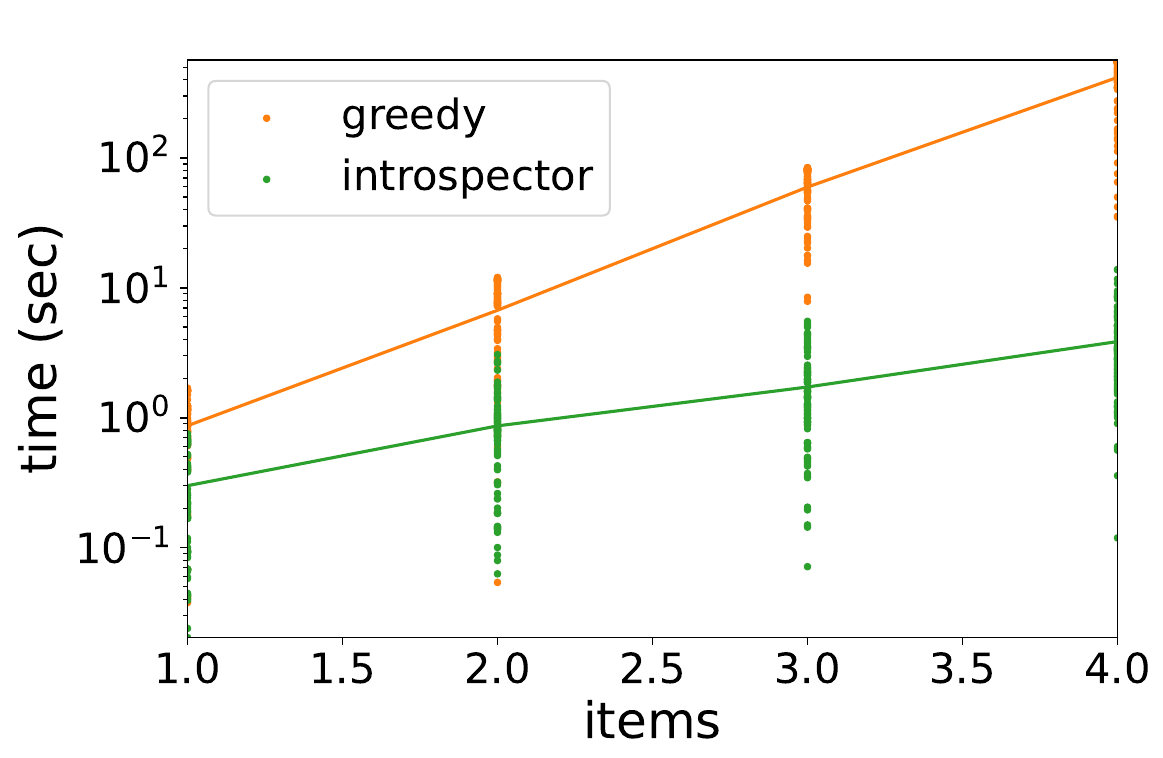}
         \caption{planner runtime, zoomed in (log-$y$)}
     \end{subfigure}
     \begin{subfigure}[t]{0.33\textwidth}
     \centering
         \includegraphics[width=\textwidth]{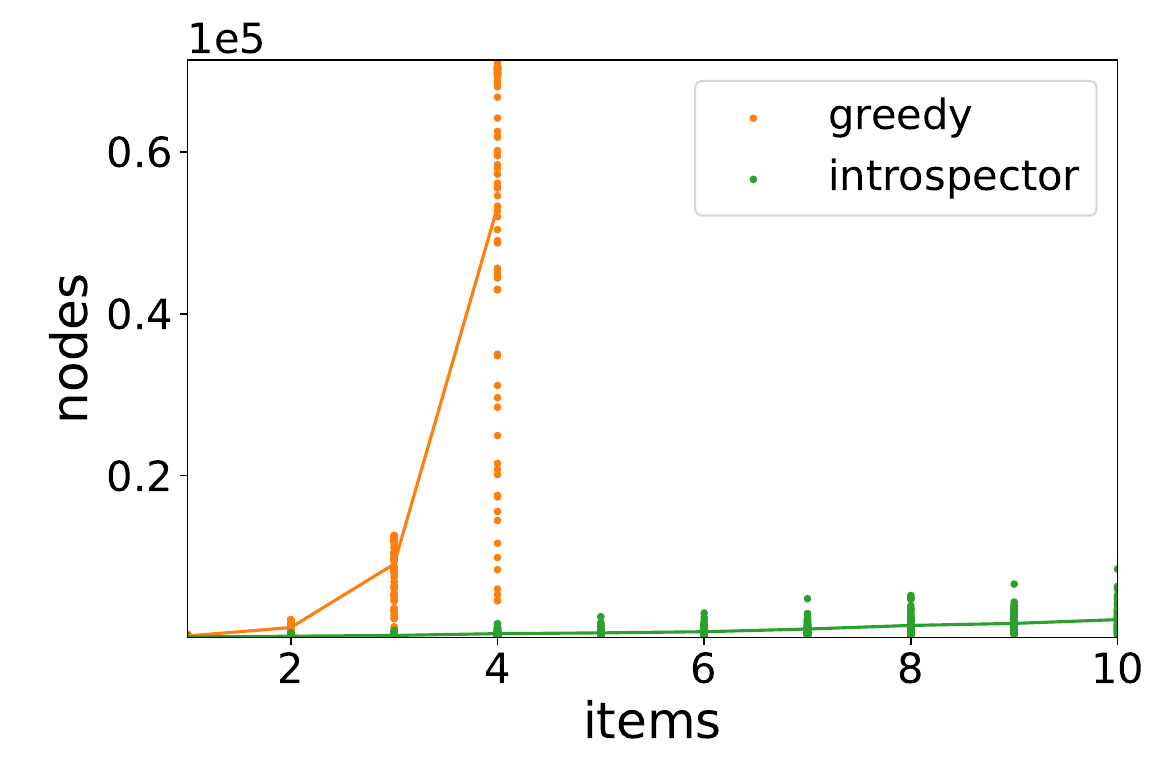}
         \caption{nodes expanded (linear scale)}
     \end{subfigure}
     \hfill
     \begin{subfigure}[t]{0.33\textwidth}
     \centering
         \includegraphics[width=\textwidth]{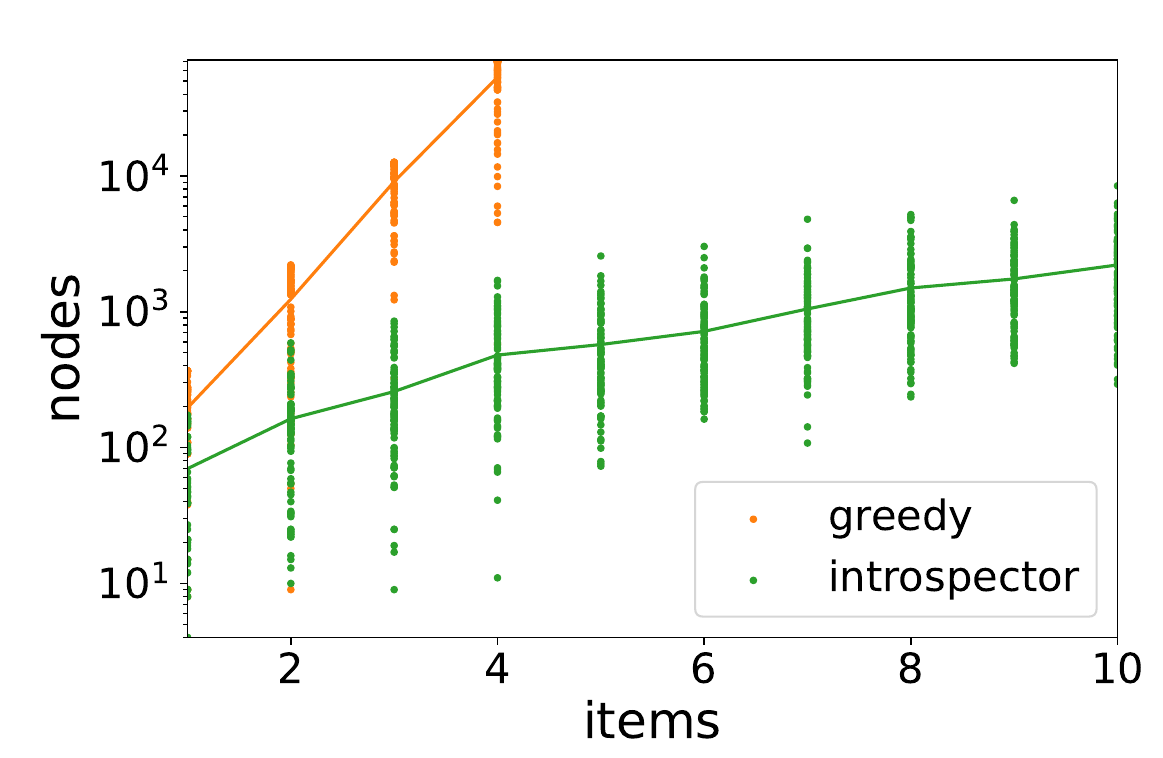}
         \caption{nodes expanded (log-$y$)}
     \end{subfigure}
     \hfill
     \begin{subfigure}[t]{0.33\textwidth}
     \centering
         \includegraphics[width=\textwidth]{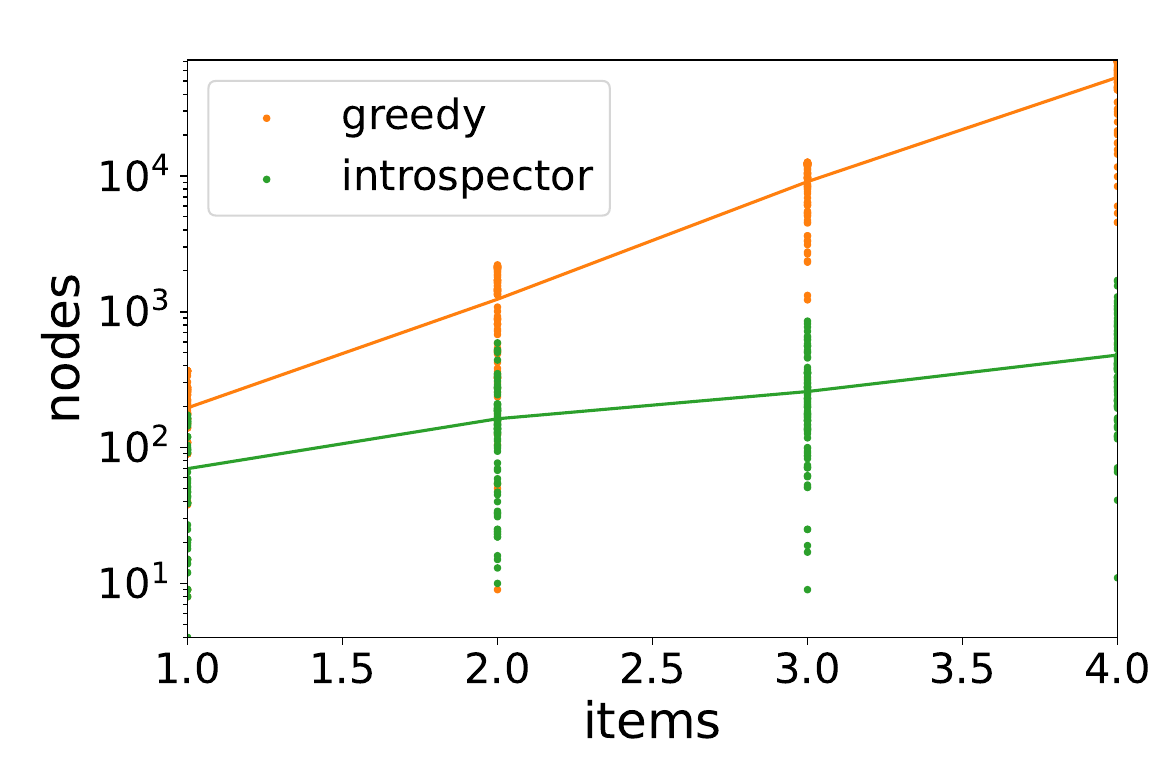}
         \caption{nodes expanded, zoomed in (log-$y$)}
     \end{subfigure}
     
\end{center}
\caption{Planning results in the drawers domain with $n = 4$}
\Description{Planning results in the drawers relational domain, showing that our introspection-based planner performs significantly better than greedy search.}
\label{fig:rrl-complete-3}
\end{figure}

\begin{figure}[h!] 
\begin{center}

     \begin{subfigure}[t]{0.33\textwidth}
     \centering
         \includegraphics[width=\textwidth]{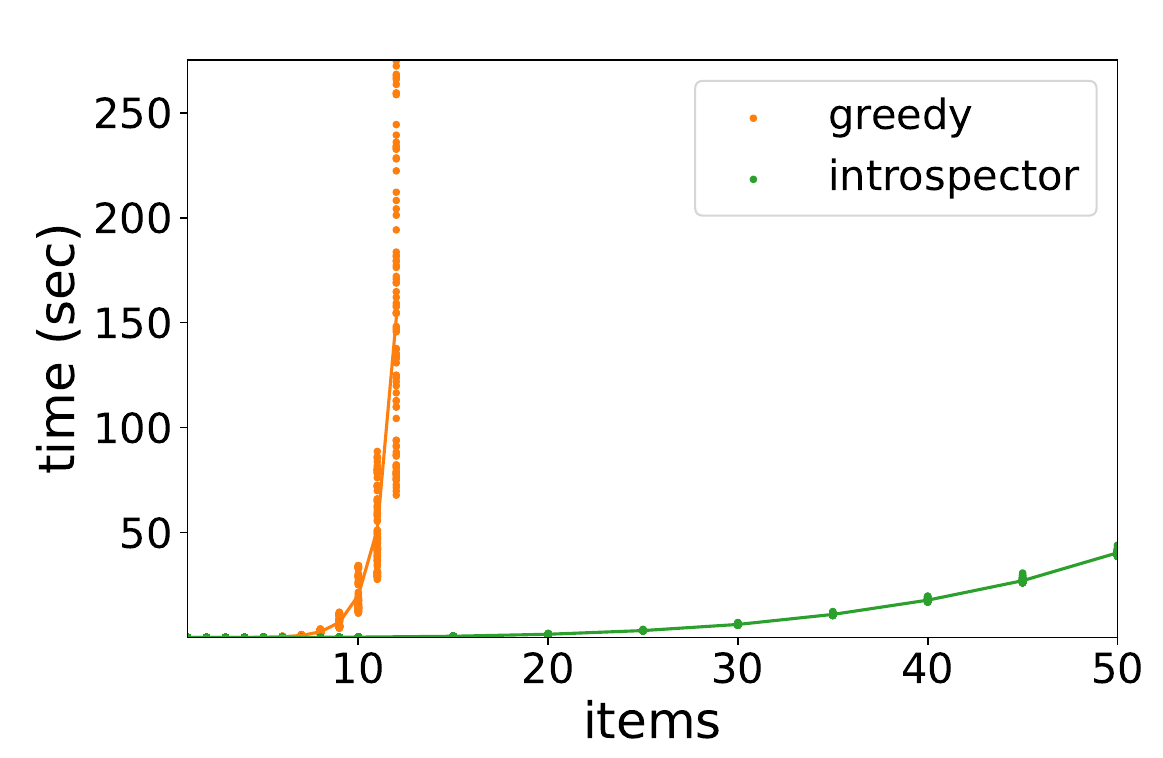}
         \caption{planner runtime (linear scale)}
     \end{subfigure}
     \hfill
     \begin{subfigure}[t]{0.33\textwidth}
     \centering
         \includegraphics[width=\textwidth]{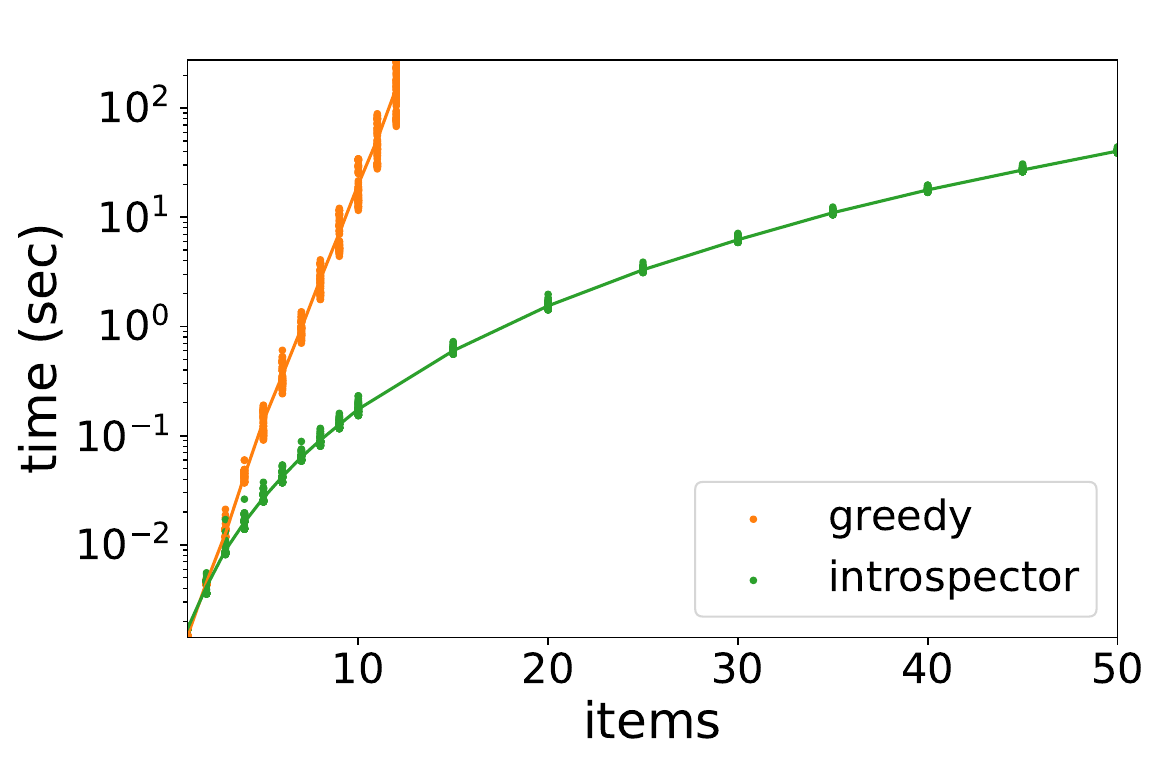}
         \caption{planner runtime (log-$y$)}
     \end{subfigure}
     \hfill
     \begin{subfigure}[t]{0.33\textwidth}
     \centering
         \includegraphics[width=\textwidth]{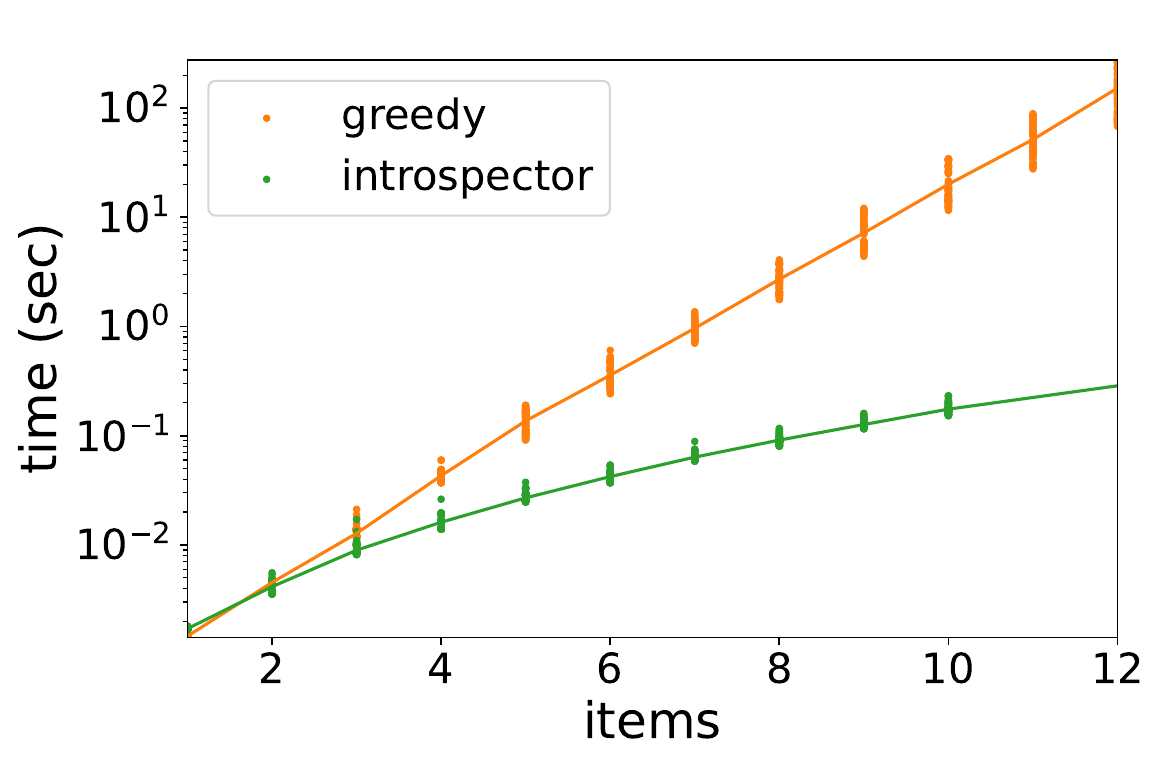}
         \caption{planner runtime, zoomed in (log-$y$)}
     \end{subfigure}
     \begin{subfigure}[t]{0.33\textwidth}
     \centering
         \includegraphics[width=\textwidth]{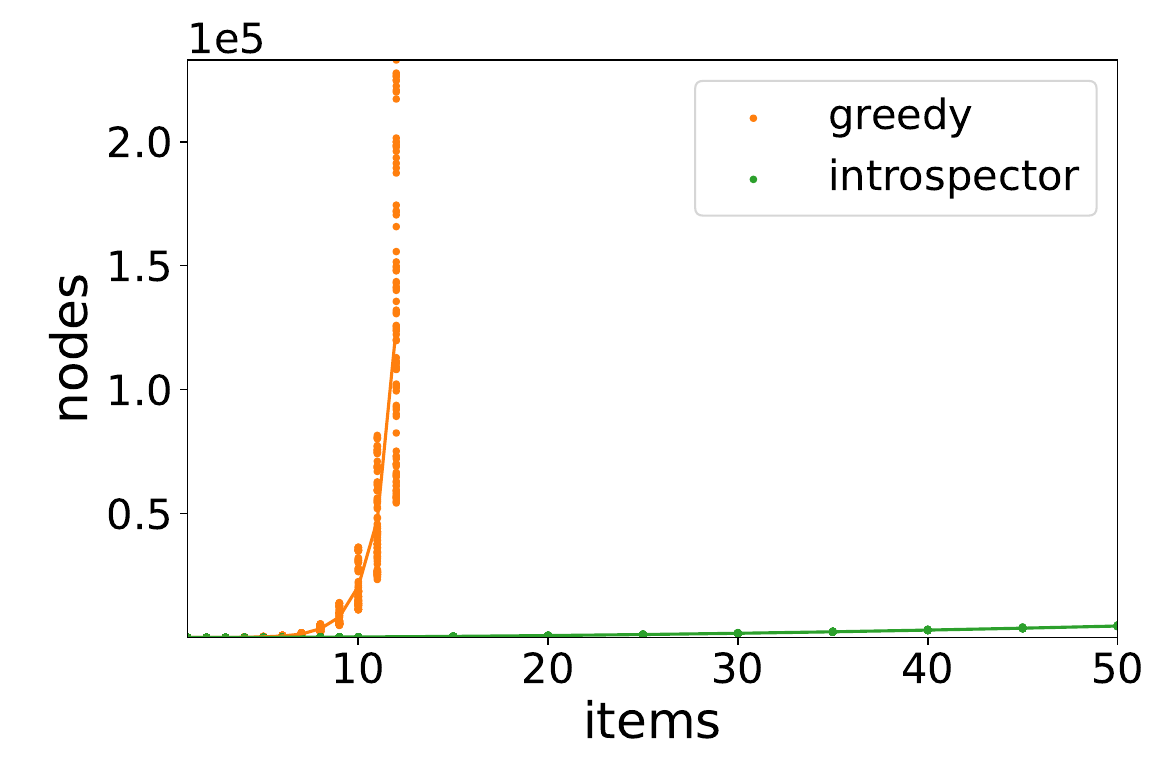}
         \caption{nodes expanded (linear scale)}
     \end{subfigure}
     \hfill
     \begin{subfigure}[t]{0.33\textwidth}
     \centering
         \includegraphics[width=\textwidth]{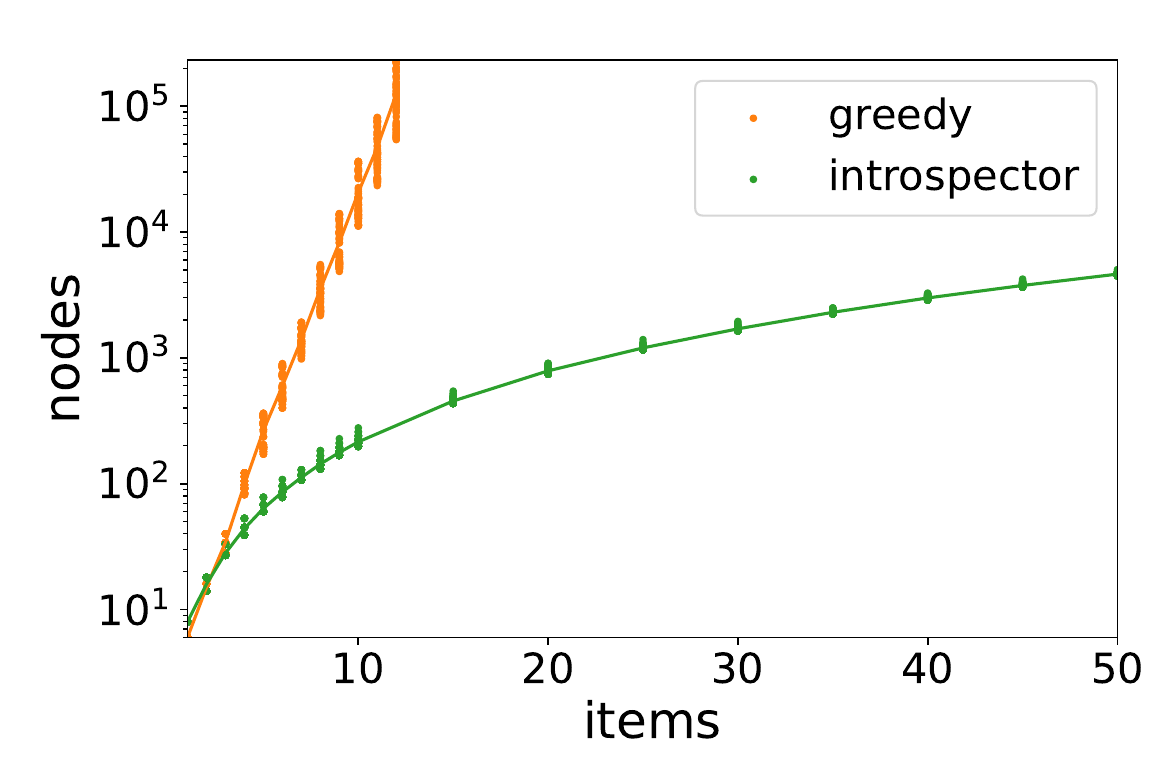}
         \caption{nodes expanded (log-$y$)}
     \end{subfigure}
     \hfill
     \begin{subfigure}[t]{0.33\textwidth}
     \centering
         \includegraphics[width=\textwidth]{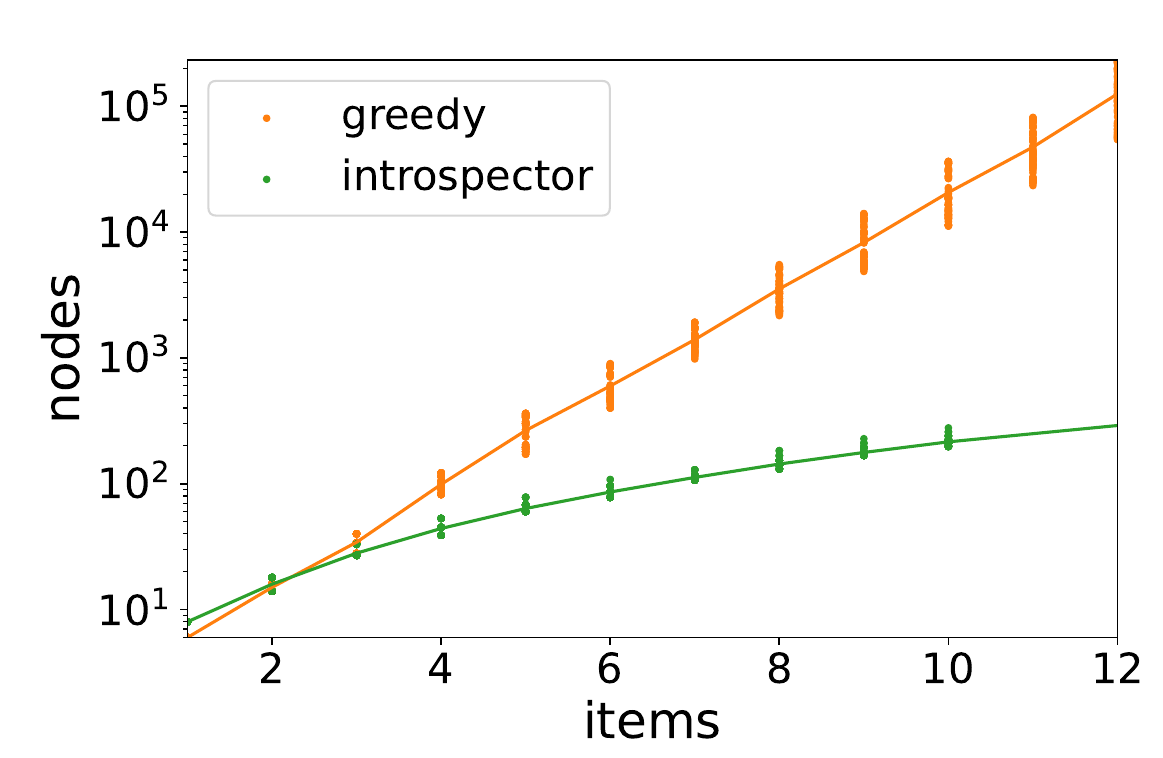}
         \caption{nodes expanded, zoomed in (log-$y$)}
     \end{subfigure}
     
\end{center}
\caption{Planning results in the bins domain with $n = 2$}
\Description{Planning results in the bins relational domain, showing that our introspection-based planner performs significantly better than greedy search.}
\label{fig:rrl-complete-4}
\end{figure}

\begin{figure}[h!] 
\begin{center}

     \begin{subfigure}[t]{0.33\textwidth}
     \centering
         \includegraphics[width=\textwidth]{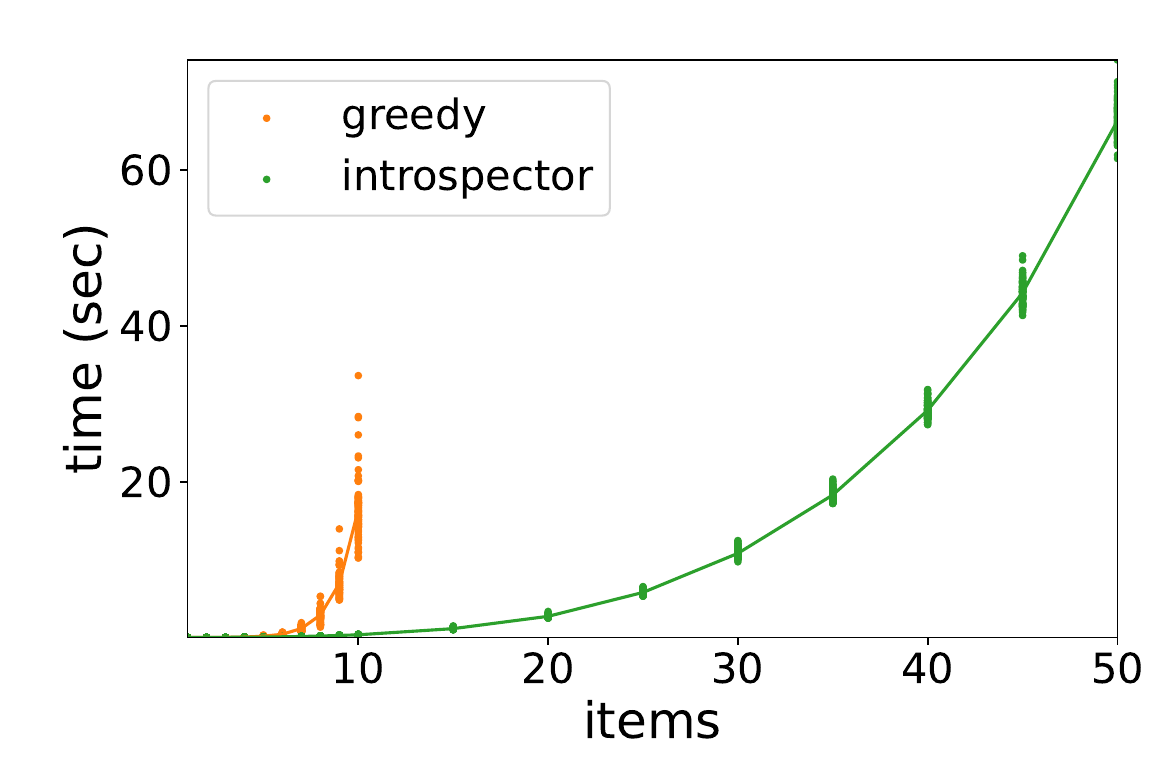}
         \caption{planner runtime (linear scale)}
     \end{subfigure}
     \hfill
     \begin{subfigure}[t]{0.33\textwidth}
     \centering
         \includegraphics[width=\textwidth]{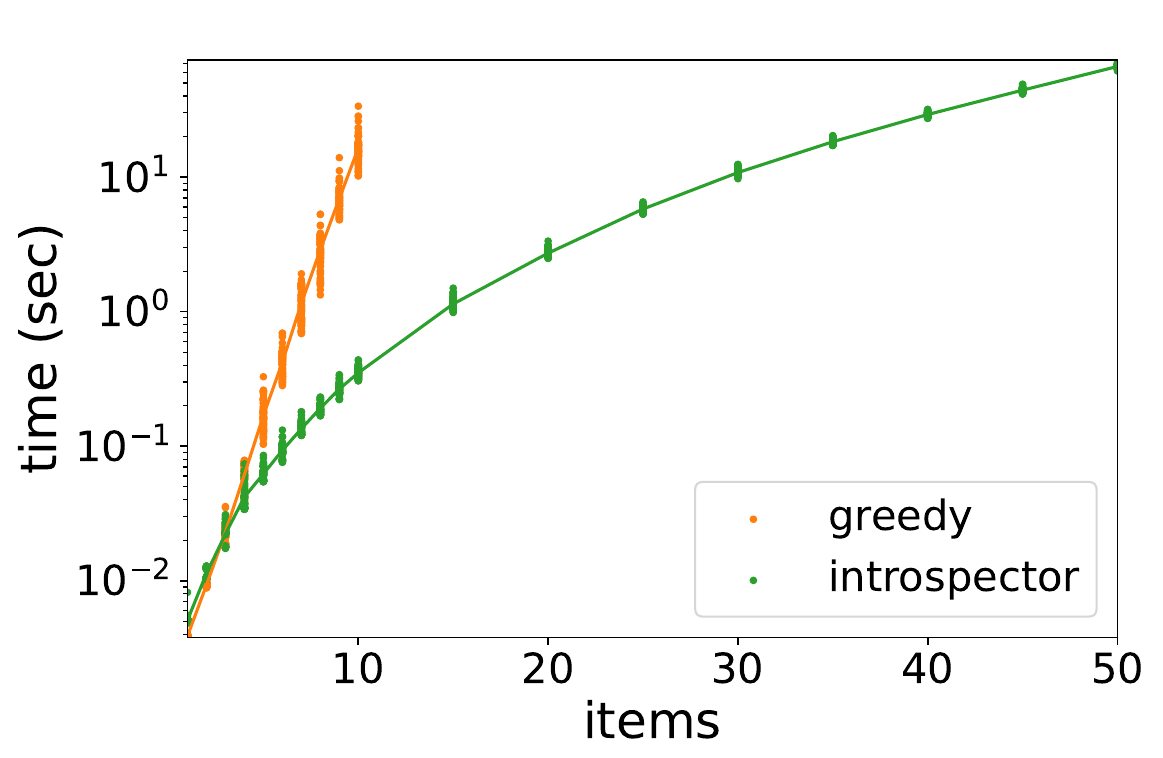}
         \caption{planner runtime (log-$y$)}
     \end{subfigure}
     \hfill
     \begin{subfigure}[t]{0.33\textwidth}
     \centering
         \includegraphics[width=\textwidth]{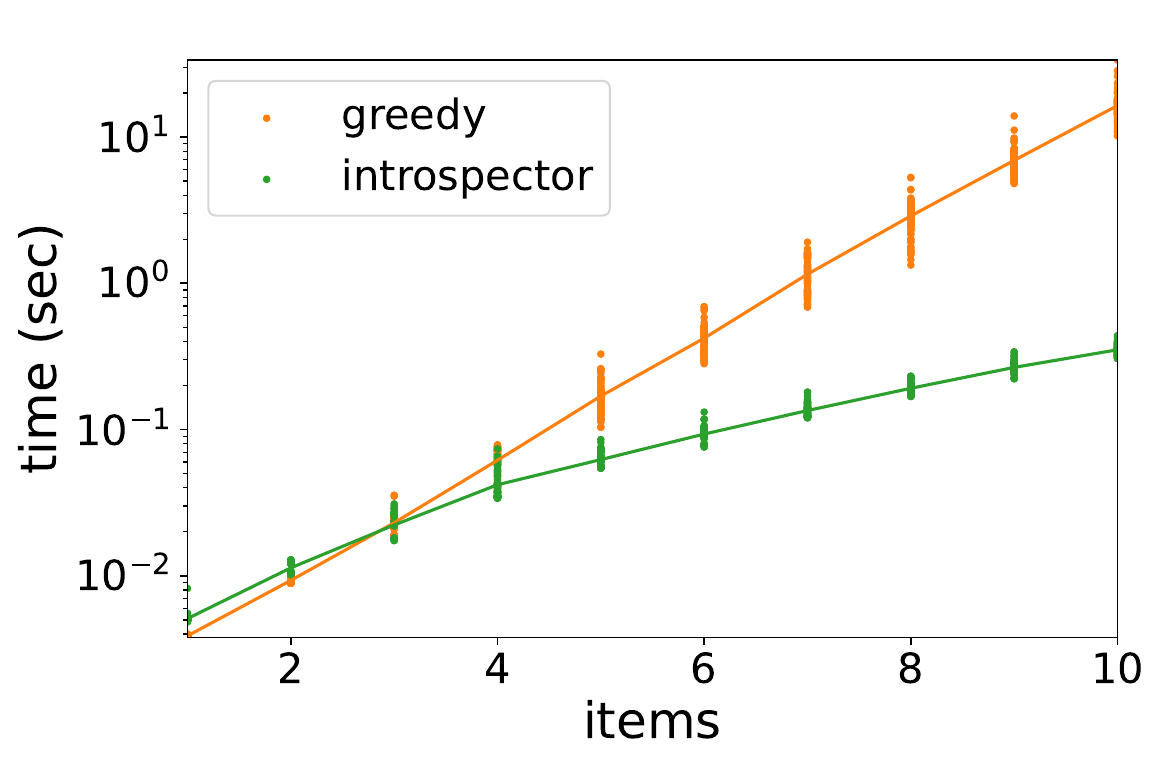}
         \caption{planner runtime, zoomed in (log-$y$)}
     \end{subfigure}
     \begin{subfigure}[t]{0.33\textwidth}
     \centering
         \includegraphics[width=\textwidth]{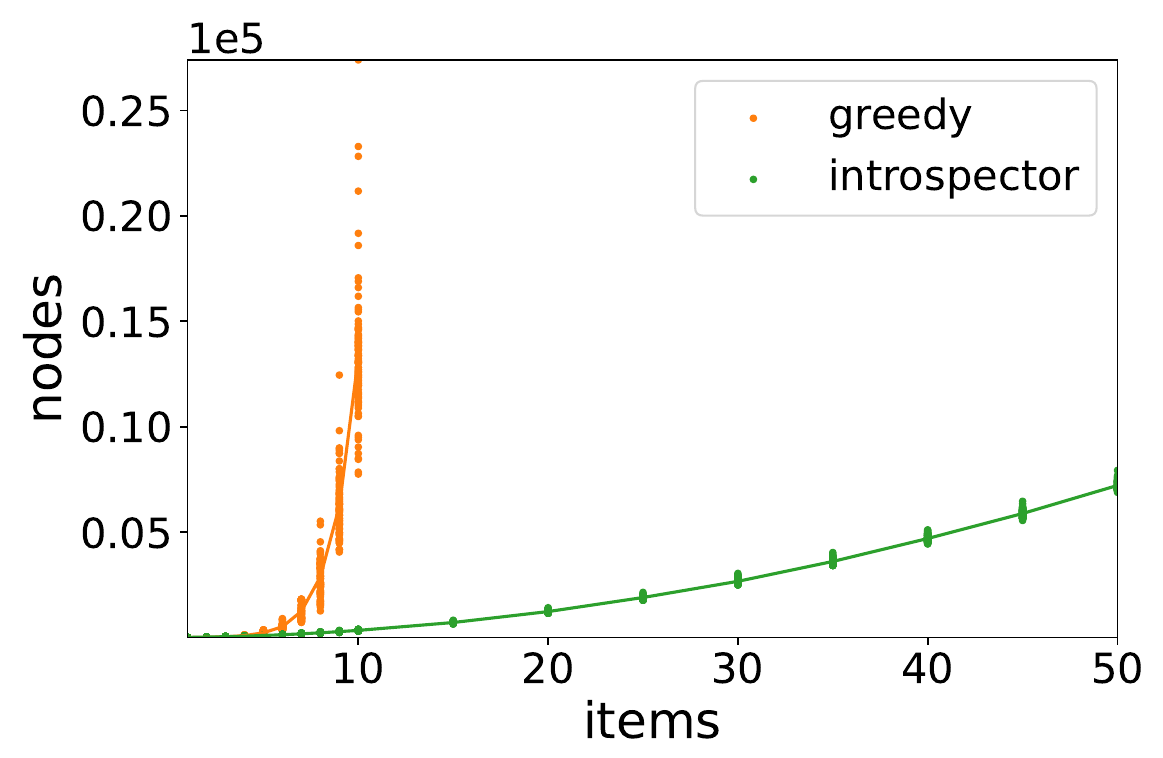}
         \caption{nodes expanded (linear scale)}
     \end{subfigure}
     \hfill
     \begin{subfigure}[t]{0.33\textwidth}
     \centering
         \includegraphics[width=\textwidth]{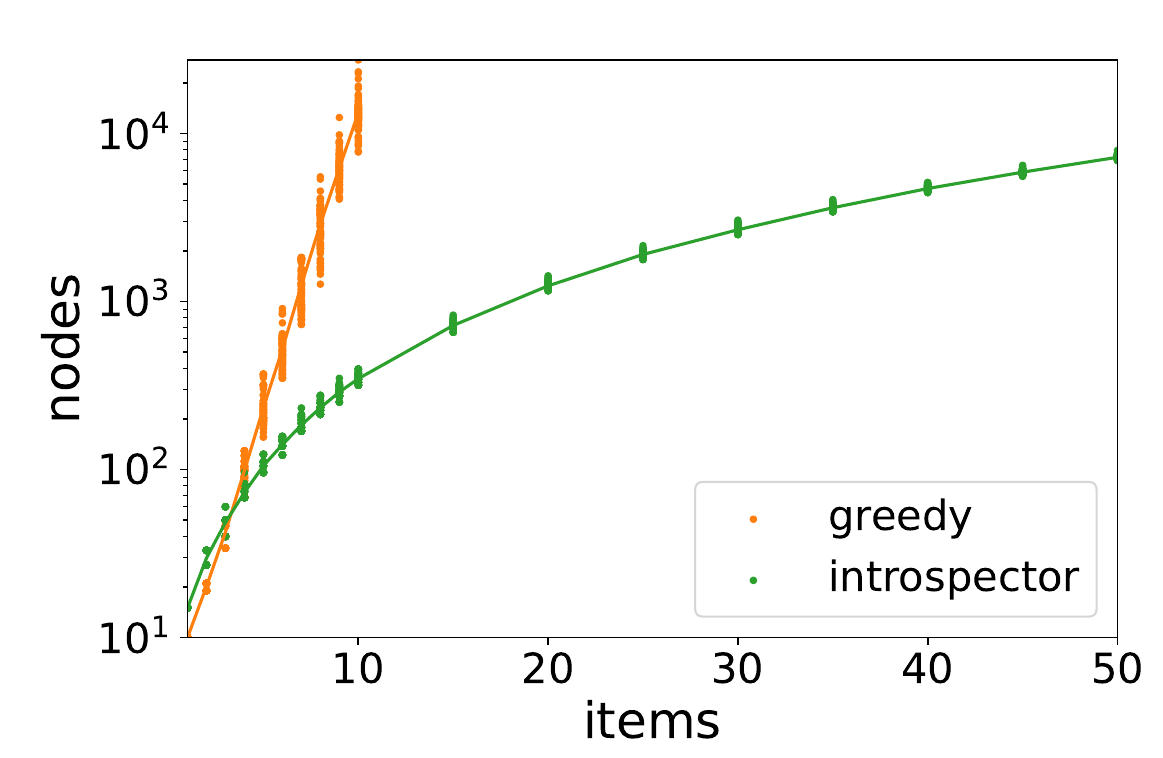}
         \caption{nodes expanded (log-$y$)}
     \end{subfigure}
     \hfill
     \begin{subfigure}[t]{0.33\textwidth}
     \centering
         \includegraphics[width=\textwidth]{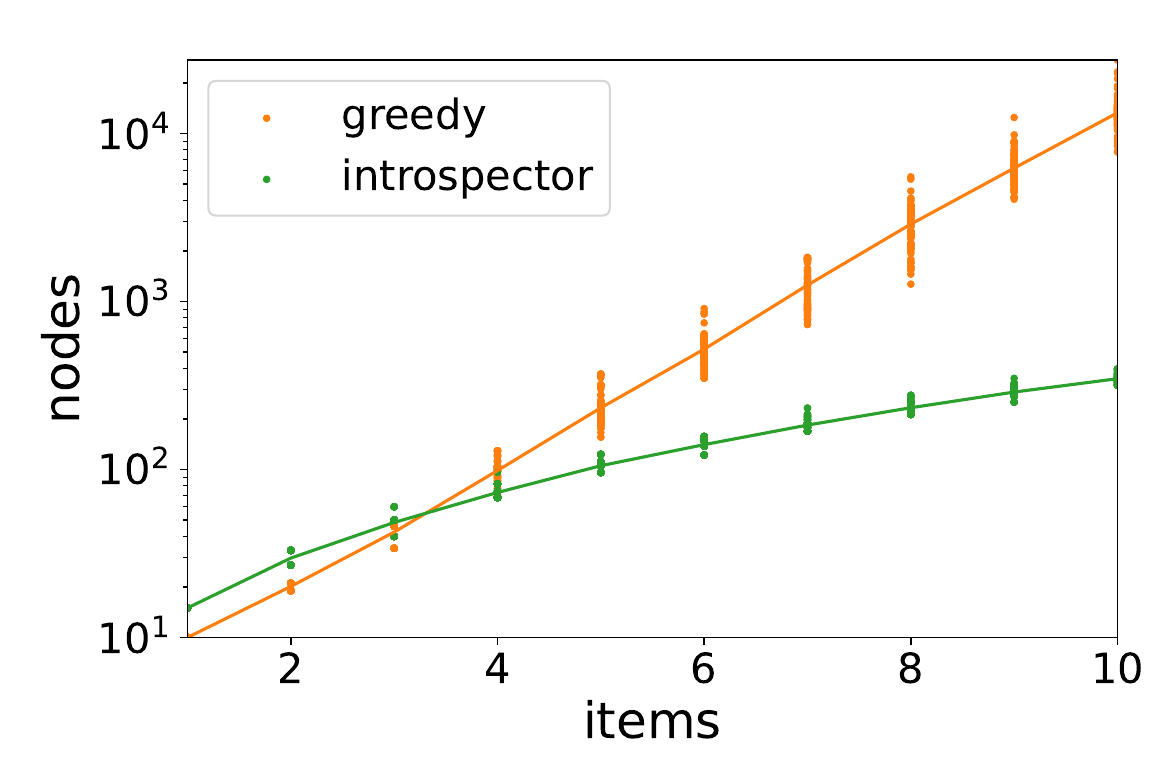}
         \caption{nodes expanded, zoomed in (log-$y$)}
     \end{subfigure}
     
\end{center}
\caption{Planning results in the bins domain with $n = 3$}
\Description{Planning results in the bins relational domain, showing that our introspection-based planner performs significantly better than greedy search.}
\label{fig:rrl-complete-5}
\end{figure}

Additional plots with results from the experiments in Section~\ref{subsec:rrl-experiments} are included below. Each figure displays planner runtime (top row) and number of nodes expanded (bottom row). The three columns are: linear-$y$, log-$y$, and zoomed-in log-$y$. The latter renders the exponential (or super-exponential) growth of the greedy planner more clearly. Results are organized as follows:
\begin{itemize}
\item Figure~\ref{fig:rrl-complete-1} contains results for the Blocks-World environment.
\item Figure~\ref{fig:rrl-complete-2} contains results for the drawers environment with $n=3$.
\item Figure~\ref{fig:rrl-complete-3} contains results for the drawers environment with $n=4$.
\item Figure~\ref{fig:rrl-complete-4} contains results for the bins environment with $n=2$.
\item Figure~\ref{fig:rrl-complete-5} contains results for the bins environment with $n=3$.
\end{itemize}

\end{document}